\documentclass[letterpaper,journal]{IEEEtran}

\usepackage{amsmath,bm,bbm,amsfonts,amssymb}
\usepackage{pifont}

\usepackage{soul,colortbl}
\usepackage[table,dvipsnames]{xcolor} 
\usepackage{color}

\usepackage{multirow}
\usepackage{tabularx}
\usepackage{booktabs}
\usepackage{pict2e}
\usepackage{url}
\usepackage{etoolbox}
\usepackage{orcidlink}
\usepackage{fontawesome5}
\usepackage{overpic}
\definecolor{cvprblue}{rgb}{0.21,0.49,0.74}
\definecolor{LightCyan}{rgb}{0.88,1,1}
\definecolor{gcolor}{RGB}{112,173,71}
\definecolor{gcolor}{RGB}{40,160,70}
\definecolor{ycolor}{RGB}{255,196,0}
\definecolor{ycolor}{RGB}{222,158,20}
\usepackage{hyperref}
\hypersetup{
     colorlinks=true,
     linkcolor=blue,
     citecolor=blue,
     filecolor=blue,
     urlcolor=magenta,
}
\usepackage{cite}

\usepackage{makecell}
\usepackage{siunitx} \usepackage[linesnumbered,ruled,vlined]{algorithm2e}
\usepackage[capitalise]{cleveref}
\Crefname{section}{Sec.}{Sects.}
\crefname{figure}{Fig.}{Figs.}
\Crefname{figure}{Fig.}{Figs.}
\crefname{table}{Tab.}{Tabs.}
\Crefname{table}{Tab.}{Tabs.}
\crefname{appendix}{}{Appendixes}
\Crefname{appendix}{}{Appendixes}

\crefformat{equation}{Eq.~{#2(#1)#3}}

\newtheorem{theorem}{Theorem}
\newtheorem{remark}{Remark}
\newtheorem{proof}{Proof}

\usepackage{subcaption}
\usepackage{arydshln}
\usepackage{tikz}

\NewDocumentCommand{\emojiBR}{ O{blue} O{\normalsize} m }{\put(1,4){\color{#1}#2 #3}}

\usepackage{contour}
\contourlength{0.1em} 

\newcommand{\myPara}[1]{\par\textbf{#1}}

\definecolor{currentOrange}{RGB}{255, 165, 0} 
\definecolor{deviatedBlue}{RGB}{0, 100, 200} 
\definecolor{sessionGreen}{RGB}{0, 150, 0} 
\definecolor{oldYellow}{RGB}{255, 204, 0} 
\definecolor{scoreRed}{RGB}{255, 0, 0}

\usepackage{xstring}   

\colorlet{highlightcolor}{red!70!black}
\newif\ifMShighlight
\MShighlightfalse

\newif\ifMSinputmode
\MSinputmodefalse

\usepackage[switch,pagewise]{lineno}
\linenumbers

\ifMShighlight
    \else
    
\fi
\newcommand{\MSfigcaption}[3]{\ifMSinputmode
    \caption{\textcolor{highlightcolor}{#2}}\else
    \caption{#2}\fi
  \label{#1}#3}

\newcommand{\MStabcaption}[2]{\ifMSinputmode
    \caption{\textcolor{highlightcolor}{#2}}\else
    \caption{#2}\fi
  \label{#1}}

\makeatletter
\let\MS@originput\input
\renewcommand{\input}[1]{\IfBeginWith{#1}{r3/}{
    \ifMShighlight
      \begingroup
        \MSinputmodetrue        \linelabel{start:#1}\color{highlightcolor}\arrayrulecolor{highlightcolor}\MS@originput{#1}\arrayrulecolor{black}\linelabel{end:#1}\endgroup
      \ignorespaces
    \else 
      \MS@originput{#1}\ignorespaces
    \fi
  }{\MS@originput{#1}\ignorespaces
  }}
\makeatother

\definecolor{rowgrayblue}{RGB}{245,248,250}
\definecolor{rowgraygreen}{RGB}{246,249,246}
\definecolor{rowgrayyellow}{RGB}{250,249,243}
\definecolor{rowgraypink}{RGB}{250,246,247}

\begin{document}

\title{Continual Action Quality Assessment via Adaptive Manifold-Aligned Graph Regularization}

\author{
    Kanglei Zhou$^{\orcidlink{0000-0002-4660-581X}}$, Qingyi Pan$^{\orcidlink{0000-0002-3823-4254}}$,  Xingxing Zhang$^{\orcidlink{0000-0003-4012-3796}}$, Hubert P. H. Shum$^{\orcidlink{0000-0001-5651-6039}}$,~\IEEEmembership{Senior Member,~IEEE},\\ Frederick W. B. Li$^{\orcidlink{0000-0002-4283-4228}}$, Xiaohui Liang$^{\orcidlink{0000-0001-6351-2538}}$,~\IEEEmembership{Member,~IEEE}, and Liyuan Wang$^{\orcidlink{0009-0002-7797-325X}}$,~\IEEEmembership{Member,~IEEE}
    \thanks{
        Manuscript received \today.
        This work was supported by the NSFC Project Nos.~62406160 and 62595773, the Beijing Natural Science Foundation L247011, the Beijing Nova Program No. 202604841279, and the Beijing Major Science and Technology Project No. Z251100008425003. \textit{(Corresponding author: Liyuan Wang).}
    }\thanks{
        Kanglei Zhou and Liyuan Wang are with the Department of Psychological and Cognitive Sciences, Tsinghua University, Beijing 100084, China. (e-mail: zhoukanglei@tsinghua.edu.cn; liyuanwang@tsinghua.edu.cn).
    }
\thanks{
        Qingyi Pan and Xingxing Zhang are with the Department of Computer Science and Technology, Tsinghua University, Beijing, China. (e-mail: pqy\_edu@163.com; xxzhang1993@gmail.com).
    }\thanks{
        Hubert P. H. Shum and Frederick W. B. Li are with the Department of Computer Science, Durham University, DH1 3LE Durham, U.K. (e-mail: hubert.shum@durham.ac.uk; frederick.li@durham.ac.uk).
    }\thanks{
        Xiaohui Liang is with the State Key Laboratory of Virtual Reality Technology and Systems, Beihang University, and the Zhongguancun Laboratory, Beijing, China (e-mail: liang\_xiaohui@buaa.edu.cn).
    }}

\markboth{IEEE Transactions on Pattern Analysis and Machine Intelligence}{Zhou \MakeLowercase{\textit{et al.}}: Continual Action Quality Assessment via Adaptive Manifold-Aligned Graph Regularization}

\maketitle

\begin{abstract}
Action Quality Assessment (AQA) quantifies human actions in videos, supporting applications in sports scoring, rehabilitation, and skill evaluation. A major challenge lies in the non-stationary nature of quality distributions in real-world scenarios, which limits the generalization of conventional methods. We introduce Continual AQA (CAQA), which equips AQA with Continual Learning (CL) capabilities to handle evolving distributions while mitigating catastrophic forgetting. Although parameter-efficient fine-tuning of pretrained models has shown promise in continual learning, our empirical study shows that the evaluated adapter-based PEFT setting provides less effective downstream adaptation than FPFT for fine-grained AQA. Our empirical and theoretical analyses reveal two insights: (i) sufficiently expressive backbone adaptation is important for bridging the upstream--downstream representation gap; yet (ii) uncontrolled FPFT may induce overfitting and feature manifold shift, thereby aggravating forgetting. To address this, we propose Adaptive Manifold-Aligned Graph Regularization (MAGR++), which couples backbone fine-tuning that stabilizes shallow layers while adapting deeper ones with a two-step feature rectification pipeline: a manifold projector to translate deviated historical features into the current representation space, and a graph regularizer to align local and global distributions. We construct four CAQA benchmarks from three datasets with tailored evaluation protocols and strong baselines, enabling systematic cross-dataset comparison. Extensive experiments show that MAGR++ achieves state-of-the-art performance, with average correlation gains of 3.6\% offline and 12.2\% online over the strongest baseline, confirming its robustness and effectiveness.
Our code is available at \url{https://github.com/ZhouKanglei/MAGRPP}.
\end{abstract}


\section{Introduction}
\label{sec:intro}
\begin{figure}
    \centering
    \begin{tikzpicture}[baseline=(current bounding box.south)]
        \node[inner sep=0] (variation) {\includegraphics[height=0.310\linewidth,clip,trim=75 85 480 85]{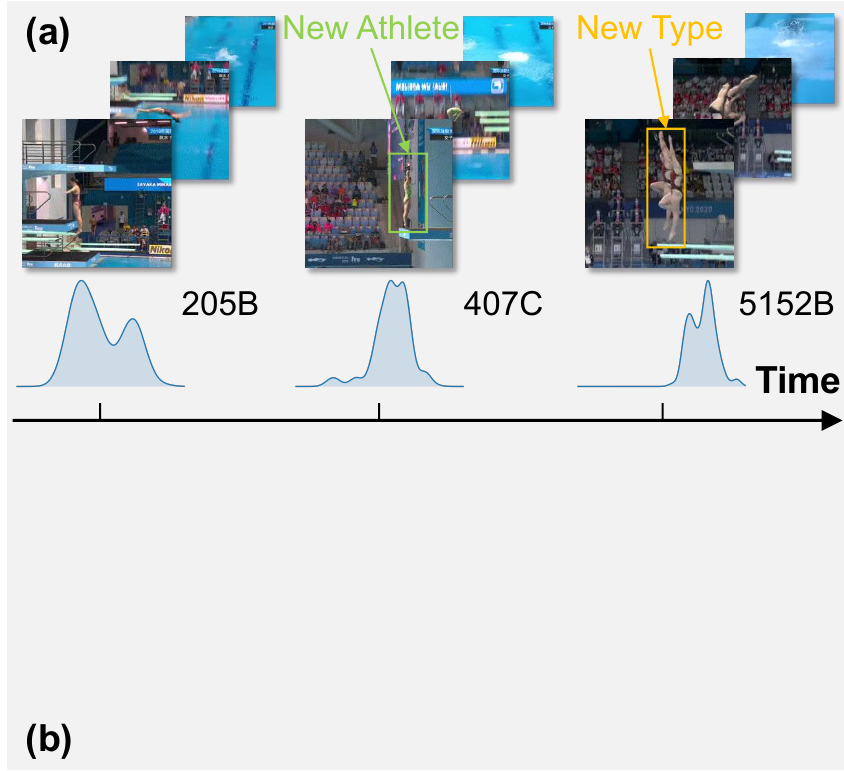}};
        \node[anchor=south west,inner sep=0] at (variation.south west) {\includegraphics[width=0.335\linewidth]{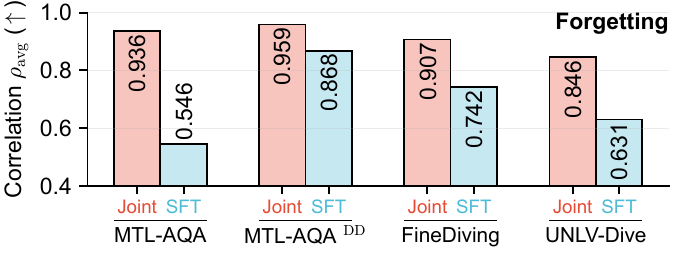}};
    \end{tikzpicture}
    \includegraphics[height=0.310\linewidth,]{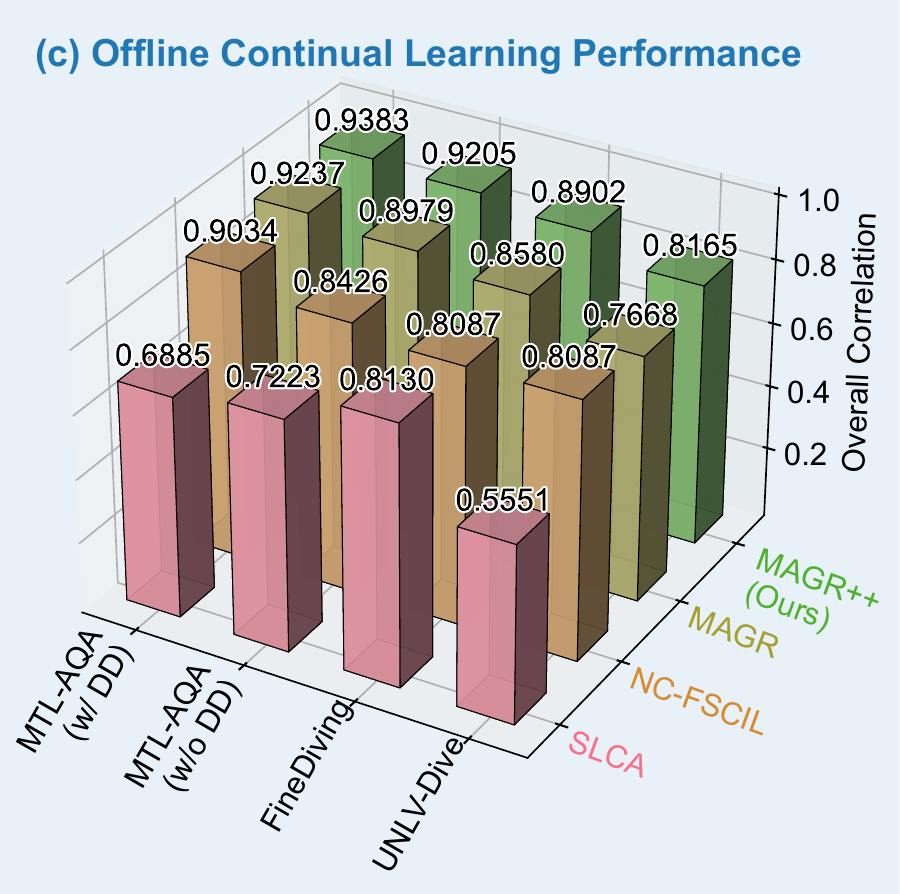}
    \includegraphics[height=0.310\linewidth,]{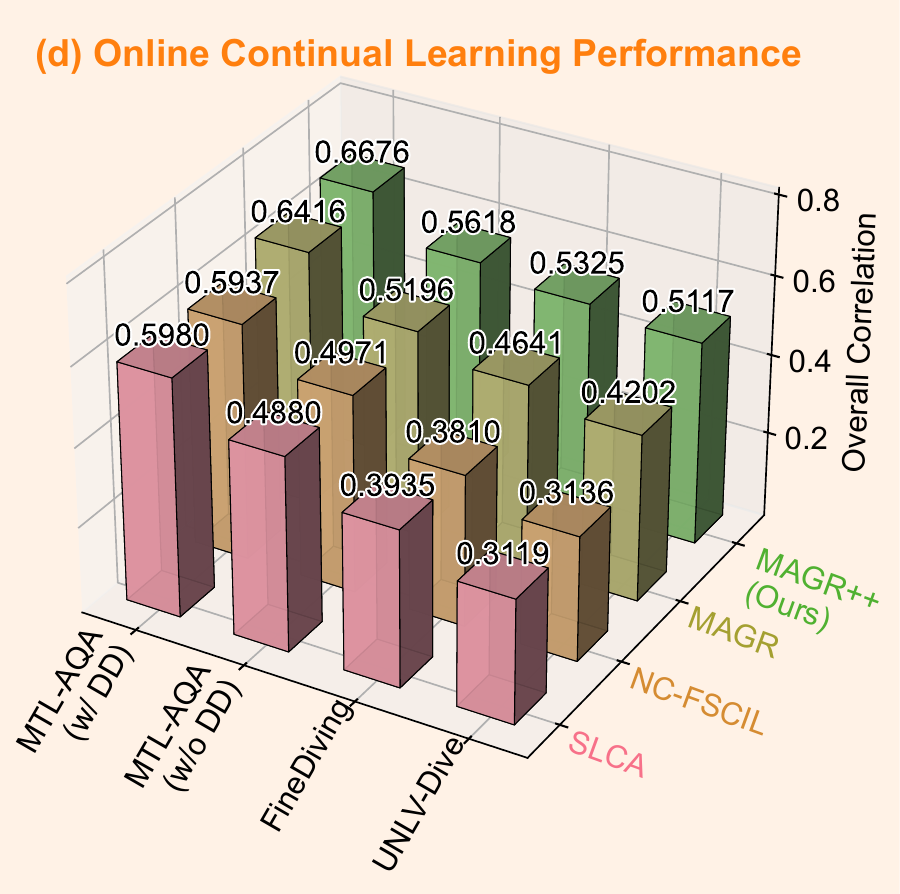}
    \MSfigcaption{fig:teaser}{
Empirical motivation and challenges of CAQA. 
\subref{fig:teaser-a}: Real-world AQA exhibits non-stationary quality distributions across action types and athlete groups. 
\subref{fig:teaser-b}: Such distribution shifts lead to severe forgetting when moving from joint training to sequential fine-tuning. 
\subref{fig:teaser-c} and \subref{fig:teaser-d}: Recent strong CL baselines still show clear performance gaps under offline and online CAQA settings, respectively.
}{}
{
    \phantomsubcaption\label{fig:teaser-a}
    \phantomsubcaption\label{fig:teaser-b}
    \phantomsubcaption\label{fig:teaser-c}
    \phantomsubcaption\label{fig:teaser-d}
    }
    \vspace{-0.1cm}
\end{figure}

\IEEEPARstart{A}{ction} Quality Assessment (AQA) \cite{gedamu2024self,majeedi2024rica,dong2024interpretable,liu2025adaptive,xu2025dancefix} evaluates how well human actions are performed in videos, offering an objective alternative to subjective judgment. It has diverse applications in sports scoring \cite{zeng2024multimodal,ji2023localization,zhang2023logo}, rehabilitation \cite{zhou2023video}, skill assessment \cite{parmar2021piano,li2022surgical}, etc. As reliable annotations require domain expertise, data collection becomes costly, limiting dataset scale. To address this, many studies employ Pretrained Models (PTMs) \cite{carreira2017quo} on large-scale action recognition datasets \cite{kay2017kinetics}. 
Although these models provide strong representations, a gap still exists between coarse-grained pretraining and fine-grained AQA tasks, requiring additional downstream adaptation~\cite{dadashzadeh2024pecop,zhou2024cofinal,zhou2025phi}. 
Beyond this upstream-downstream discrepancy, real-world AQA further involves non-stationary score distributions, where scoring patterns evolve across action types, difficulty levels, and athlete groups (see \cref{fig:teaser-a}). Importantly, previously observed quality distributions do not necessarily become obsolete, since deployed AQA systems may still need to assess performers from earlier action types, skill levels, or athlete groups. Simply adapting to the latest distribution can therefore degrade previously acquired scoring capability (see \cref{fig:teaser-b}), thereby limiting real-world applicability. 

To this end, Continual Learning (CL)~\cite{wang2024comprehensive} provides a principled framework for adapting models to evolving distributions while retaining previously acquired knowledge. 
Unlike standard regression or distribution adaptation that primarily targets the current distribution, CAQA requires models to adapt sequentially while preserving reliable assessment on previously learned quality distributions. However, most CL research has focused on classification tasks~\cite{chi2022metafscil,wang2023hierarchical,wang2021afec}, whereas its extension to AQA requires continual score regression and remains largely unexplored. Bridging this gap demands clear task formulation, benchmark construction, and customized evaluation protocols. 
In this work, we formalize and extend \textbf{Continual AQA (CAQA)} as a continual score-regression setting that confronts the dilemma between capturing fine-grained motion cues through continual adaptation and maintaining stability under non-stationary score distributions. 
The complexity of CAQA poses unique challenges that render even strong CL baselines ineffective when applied directly (see \cref{fig:teaser-c,fig:teaser-d}). Existing PTM-based CL methods typically follow two paradigms: (i) extensive base-session adaptation followed by feature freezing, or (ii) Parameter-Efficient Fine-Tuning (PEFT) for continual updates \cite{zhou2025adaptive,wang2023hierarchical,xin2024parameter}. While both strategies have shown success in classification, their \textbf{limitations} become evident in CAQA. First, the high cost of expert annotations \cite{zhou2026comprehensive} makes large-scale base-session adaptation impractical. Second, unlike coarse-grained classification, AQA relies on subtle motion cues, meaning frozen features without adequate adaptation fail to generalize across evolving distributions \cite{zhou2024cofinal,zhou2025phi}. 
Finally, although PEFT-style adaptation has shown effectiveness for reducing the upstream-downstream gap in AQA~\cite{dadashzadeh2024pecop}, PECoP further relies on subsequent Full-Parameter Fine-Tuning (FPFT), suggesting that purely parameter-constrained continual adaptation may still struggle under evolving CAQA distributions. Consistent with this observation, our \textbf{empirical study} (see \cref{sec:empirical_study}) further shows that FPFT provides stronger downstream adaptation than the evaluated adapter-based PEFT baseline in AQA.

To investigate this observation, we conduct an in-depth \textbf{theoretical analysis} under a representative, storage-efficient feature replay strategy, and obtain two key insights. 
First, while FPFT is effective for realigning representations, repeated use in long-term continual adaptation risks severe overfitting. Second, continual distribution shifts cause replayed features to drift from the evolving data manifold, undermining rehearsal effectiveness. Inspired by these, we propose \textbf{Adaptive Manifold-Aligned Graph Regularization (MAGR++)}, an innovative framework that addresses both overfitting and distribution shift in CAQA. MAGR++ employs a layer-adaptive fine-tuning strategy that constrains shallow layers from drifting while fully tuning deeper ones to embrace session-specific variations, with the boundary determined adaptively. It further incorporates a manifold projector that maps historical features into the current representation space and a graph regularizer that enforces both local and global consistency between feature and quality spaces, enabling reliable replay and regression. Together, these modules allow MAGR++ to achieve effective continual adaptation while maintaining stability.

We establish four CAQA benchmarks from three AQA datasets, with tailored evaluation metrics and strong baselines for systematic cross-dataset comparison. Experiments across both offline and online CAQA settings demonstrate that MAGR++ achieves \textbf{state-of-the-art performance}, surpassing the strongest baseline by 1.6\%-6.5\% offline and 4.0\%-21.8\% online, with average gains of 3.6\% and 12.2\%, respectively.

This work substantially extends our preliminary version MAGR~\cite{zhou2024magr}. Beyond a complete rewriting of the manuscript, MAGR++ advances the prior work in three key aspects. First, we establish a theoretical foundation that characterizes a central adaptation-stability tension in CAQA: expressive backbone adaptation is needed to bridge the coarse-to-fine representation gap, yet it simultaneously amplifies forgetting and replay-feature drift. This analysis moves beyond empirical observation and provides a principled understanding of why continual score regression is challenging. Second, motivated by these findings, we develop a coordinated solution that addresses complementary aspects of this tension through layer-adaptive fine-tuning and an asynchronous two-step feature rectification pipeline, extending MAGR from a feature-replay regularizer into a more complete adaptation-rectification framework. Third, we substantially broaden the empirical scope by incorporating both offline and online CAQA across diverse protocols and datasets, enabling a more systematic evaluation of robustness, generality, and practical applicability. 

Overall, our contributions can be summarized as follows:
\begin{itemize}
\item We formalize and extend the CAQA setting by constructing continual score-regression protocols and benchmarks for non-stationary action quality distributions. \item We develop a theoretical framework that reveals two fundamental challenges of CAQA, i.e., overfitting from repeated FPFT and feature drift under feature replay. 
\item We propose a theoretically grounded CAQA method with adaptive FPFT for robust representation learning and two-step rectification for coherent feature alignment.
\item We conduct extensive experiments on four CAQA benchmarks. Our method achieves consistently state-of-the-art performance in both offline and online settings. 
\end{itemize}

\section{Related Work} \label{sec:related_work}

\myPara{Action Quality Assessment.} AQA evaluates the objective execution quality of human actions, spanning numerous applications in sports scoring \cite{bai2022action,pan2025basket,xu2025mcmoe}, rehabilitation \cite{zhou2023video}, and skill assessment \cite{zia2016automated,pan2025basket}. 
A major challenge is the scarcity of annotated labels \cite{zhou2026comprehensive}, since reliable quality scores demand domain expertise. To address this, most approaches \cite{pan2021adaptive,yu2021group,parmar2019and} leverage PTMs (e.g., I3D \cite{carreira2017quo}) to extract strong visual features and then regress quality scores either via direct regression \cite{zhou2023hierarchical} or contrastive regression \cite{ke2024two,yu2021group}. Ranking-based skill assessment methods \cite{doughty2019pros,doughty2018s,yu2021group} further alleviate annotation costs by comparing relative performance instead of relying on absolute scores. Another core difficulty lies in fine-grained temporal parsing, as PTMs are optimized for coarse action recognition while AQA demands temporal sensitivity. To this end, strategies such as Transformer variants \cite{bai2022action,dong2024interpretable,dong2026uila}, continual pretraining \cite{dadashzadeh2024pecop}, and regularization \cite{zhou2025phi,zhou2024cofinal} have been proposed to enhance feature representations.  Meanwhile, recent studies enhance fine-grained action understanding by introducing additional structural priors and multi-modal cues beyond standard RGB representations. Specifically, these approaches leverage human-centric masks \cite{xu2024fineparser,xu2025human,wang2021tsa}, skeletal data \cite{qi2025action,pan2019action}, textual modalities \cite{majeedi2024rica,xu2025quality,xu2024vision,xu2025language}, sub-action parsing \cite{gedamu2023fine}, and gaze cues \cite{huang2024egoexolearn} to better capture detailed motion semantics and temporal structure. Furthermore, non-stationary variations across tasks pose additional challenges for AQA that conventional static training paradigms cannot address effectively. These evolving score distributions motivate the need for continual adaptation mechanisms in AQA systems.

\myPara{Continual Action Quality Assessment.}
CAQA~\cite{dadashzadeh2024pecop,li2024continual,zhou2024magr} extends CL to AQA under non-stationary score distributions.
Unlike conventional CL, which is mainly studied in classification settings, CAQA is inherently a regression problem, where quality scores must remain comparable across sessions and domains.
This makes stability–plasticity trade-offs more challenging, as the model must update fine-grained motion representations while preserving previously learned score relationships.
Existing approaches address this problem through domain-aware adaptation~\cite{dadashzadeh2024pecop}, incremental category expansion~\cite{li2024continual}, or manifold-based alignment under feature replay~\cite{zhou2024magr}.
However, limited backbone adaptation may restrict the ability to capture fine-grained quality cues under evolving distributions.
Our empirical results demonstrate that effective CAQA requires constrained full-parameter adaptation, which motivates the design of our MAGR++ method.

\myPara{Continual Learning.}
CL \cite{wang2023incorporating,wang2024comprehensive} enables models to acquire new knowledge from a stream of tasks without forgetting previous ones. This capability is particularly valuable in real-world applications such as robotics, surveillance, and other dynamic vision domains \cite{zhou2025adaptive}.
The main challenge is to avoid catastrophic forgetting of previously learned knowledge. Current efforts can be broadly divided into constraint-based and replay-based methods. Constraint-based methods such as SI~\cite{zenke2017continual}, EWC~\cite{james2017ewc}, and LwF~\cite{li2017learning} impose regularization to preserve old knowledge without storing past data, but often suffer from limited scalability. Replay-based methods, by contrast, achieve stronger retention via exemplar storage (e.g., MER~\cite{riemer2019learning}, DER++~\cite{buzzega2020dark}, TOPIC~\cite{tao2020few},  GEM~\cite{kukleva2021generalized}, and STAR \cite{eskandar2025star}), which raises memory and privacy concerns. More recently, feature replay has emerged as a lightweight and privacy-preserving alternative (e.g., SLCA~\cite{zhang2023slca}, NC-FSCIL~\cite{yang2023neural}, FS-Aug~\cite{li2024continual}, MAGR~\cite{zhou2024magr},  and ProNC~\cite{wang2026rethinking}). However, applying feature replay in domains like AQA is challenging due to significant domain gaps, and continual adaptation of the backbone often induces manifold shifts, misaligning old and new feature distributions. Hence, our work adopts feature replay as the primary technical approach for CAQA, while mitigating feature misalignment.

\section{Preliminaries for AQA and CAQA} \label{sec:foundation}
In this section, we formulate AQA and CAQA, empirically investigate FPFT and PEFT in AQA, and provide an in-depth theoretical analysis with storage-efficient feature replay, which yields key implications for the design of MAGR++.

\begin{figure}
    \centering
    \vspace{-0.2cm}
    \includegraphics[width=\linewidth]{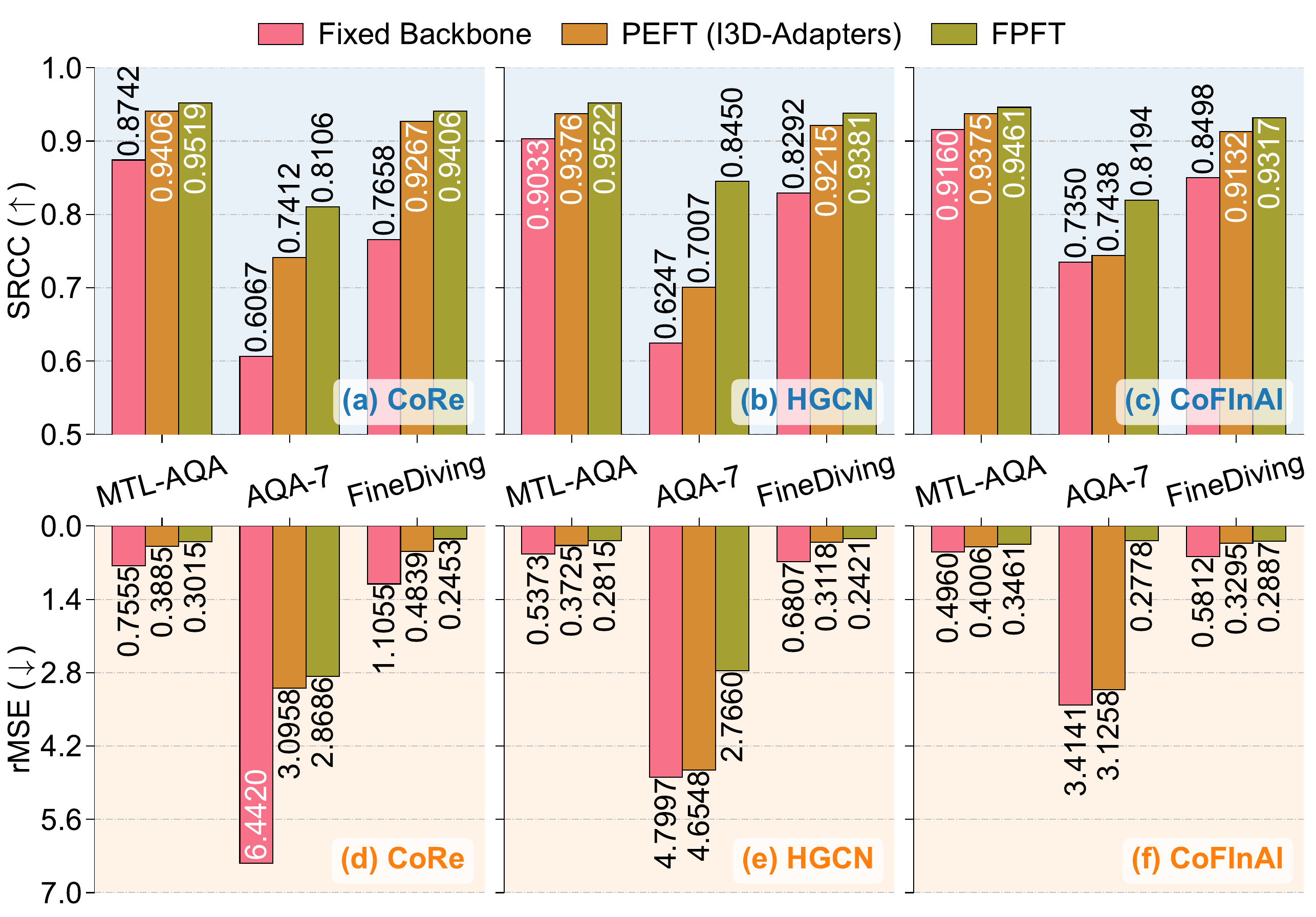}
    \caption{
        SRCC and rMSE comparison of fixed backbone, PEFT (I3D-Adapters), and FPFT in representative AQA tasks.
    }
    \label{fig:fft_vs_peft}
    \phantomsubcaption\label{fig:fft_vs_peft-a}
    \phantomsubcaption\label{fig:fft_vs_peft-b}
    \phantomsubcaption\label{fig:fft_vs_peft-c}
    \phantomsubcaption\label{fig:fft_vs_peft-d}
    \phantomsubcaption\label{fig:fft_vs_peft-e}
    \phantomsubcaption\label{fig:fft_vs_peft-f}
    \vspace{-0.2cm}
\end{figure}

\subsection{Task Formulation of AQA and CAQA}
In AQA, the goal is to assign a quantitative score $\hat{y} \in \mathbb{R}$ to a video $\mathbf{x} \in \mathbb{R}^{F \times H \times W \times 3}$, where $F$, $H$, and $W$ denote the number of frames, height, and width. A backbone $f$ extracts features $\bm{h}=f(\mathbf{x})$, and a regressor $g$ predicts scores $\hat{y}=g(\bm{h})$, trained on a labeled dataset $\mathcal{D}_{\text{train}}=\{(\mathbf{x}_n,y_n)\}_{n=1}^N$. Traditional AQA assumes that both $\mathcal{D}_{\text{train}}$ and the parameters $\bm{\theta}_f,\bm{\theta}_g$ remain fixed once trained. In practice, however, new actions, user populations, and individual variations continually emerge, leading to shifts in the underlying distribution. We therefore define \textbf{CAQA}: given a sequence of datasets $\{\mathcal{D}_{\text{train}}^t\}_{t=1}^T$ across sessions, the backbone $f^t$ and regressor $g^t$ are updated in each session to adapt to new data while retaining past knowledge. For clarity, we omit parameter dependence (e.g., $\bm{\theta}_f^t, \bm{\theta}_g^t$) in the notation when referring to session $t$.

A critical challenge in CAQA is \textbf{catastrophic forgetting}, where learning from new data degrades performance on previous sessions. Rehearsal \cite{wang2024comprehensive} is a simple yet effective remedy that stores and replays past samples. To this end, CAQA employs the storage-efficient feature replay \cite{yang2023neural}, which maintains a memory bank $\mathcal{M}$ containing only a small set of representative latent embeddings $\bm{h}$ from previous sessions.
Building on this, we define the CAQA objective as:
\begin{equation}
    \min_{\bm{\theta}_f^t,\,\bm{\theta}_g^t}~ 
    \mathcal{L}_{\text{D}} + \mathcal{L}_{\text{M}},
    \label{eq:caqa}
\end{equation}
where $\mathcal{L}_{\text{D}}$ denotes the regression loss on the current session data $\mathcal{D}_{\text{train}}^t$, and $\mathcal{L}_{\text{M}}$ represents the replay loss on the memory bank $\mathcal{M}$. This joint objective enables the model to incrementally refine its assessment ability across sessions while effectively retaining previously acquired knowledge.

\subsection{Empirical Study of PEFT and FPFT for AQA} \label{sec:empirical_study}
In this work, FPFT refers to full-parameter downstream adaptation, where all backbone and regression-head parameters are jointly optimized during training. In contrast, unless otherwise specified, PEFT~\cite{xin2024parameter} refers to parameter-efficient downstream adaptation, where the backbone remains frozen and only lightweight adaptation modules (e.g., prompts, LoRA, or adapters) are updated. PEFT methods are generally effective when upstream representations are sufficiently strong and the downstream task requires only limited adaptation, as they constrain optimization to a lightweight parameter subspace. However, AQA fundamentally differs from conventional recognition tasks: existing upstream models are typically pretrained on coarse-grained action recognition datasets, which are often poorly aligned with the fine-grained temporal and motion cues required for AQA~\cite{zhou2024cofinal,zhou2025phi}. Consequently, the effectiveness of parameter-constrained adaptation for AQA, particularly relative to FPFT, remains largely underexplored.

To close this gap, we adopt representative baselines for a fair empirical comparison. 
Existing parameter-efficient adaptation for AQA has predominantly been explored through adapter-based designs with convolutional video backbones such as I3D. Following prior adapter-based PEFT designs for AQA~\cite{dadashzadeh2024pecop}, we adopt a downstream adapter-based PEFT setting, where lightweight adapters and the regression head are optimized while the backbone remains frozen.  Concretely, we compare fixed backbone (no adaptation), PEFT with I3D-Adapters, and FPFT across three benchmarks (MTL-AQA~\cite{parmar2019and}, AQA-7~\cite{parmar2019action}, FineDiving~\cite{xu2022finediving}) with three prediction heads: CoRe~\cite{yu2021group}, HGCN~\cite{zhou2023hierarchical}, and CoFInAl~\cite{zhou2024cofinal}. As shown in \cref{fig:fft_vs_peft}, FPFT consistently achieves superior performance in both SRCC (defined in \cref{eq:rho}) and rMSE \cite{zhou2026comprehensive} across all datasets and models. Compared with the fixed backbone, FPFT yields average gains of $+$3.33\% SRCC and $-$31.23\% rMSE, confirming that upstream features are poorly aligned for AQA. Compared with PEFT, FPFT achieves average gains of $+$2.11\% SRCC and $-$14.25\% rMSE, 
with the most significant improvement on AQA-7 ($+$5.89\% SRCC, $-$44.21\% rMSE), while the gains are smaller on FineDiving ($+$0.22\% SRCC, $-$2.85\% rMSE). On MTL-AQA, FPFT improves SRCC by $+$0.23\% but incurs a modest $+$4.31\% increase in rMSE.
These results show that, under the evaluated adapter-based AQA setting, FPFT provides more effective downstream adaptation, particularly when the upstream-downstream discrepancy is large. We emphasize that this empirical observation is specific to the evaluated AQA-adapted PEFT baseline and does not imply that FPFT universally outperforms all PEFT methods.

\begin{figure}
    \centering
    \vspace{-0.1cm}
    \includegraphics[width=0.85\linewidth,clip,trim=110 120 180 140]{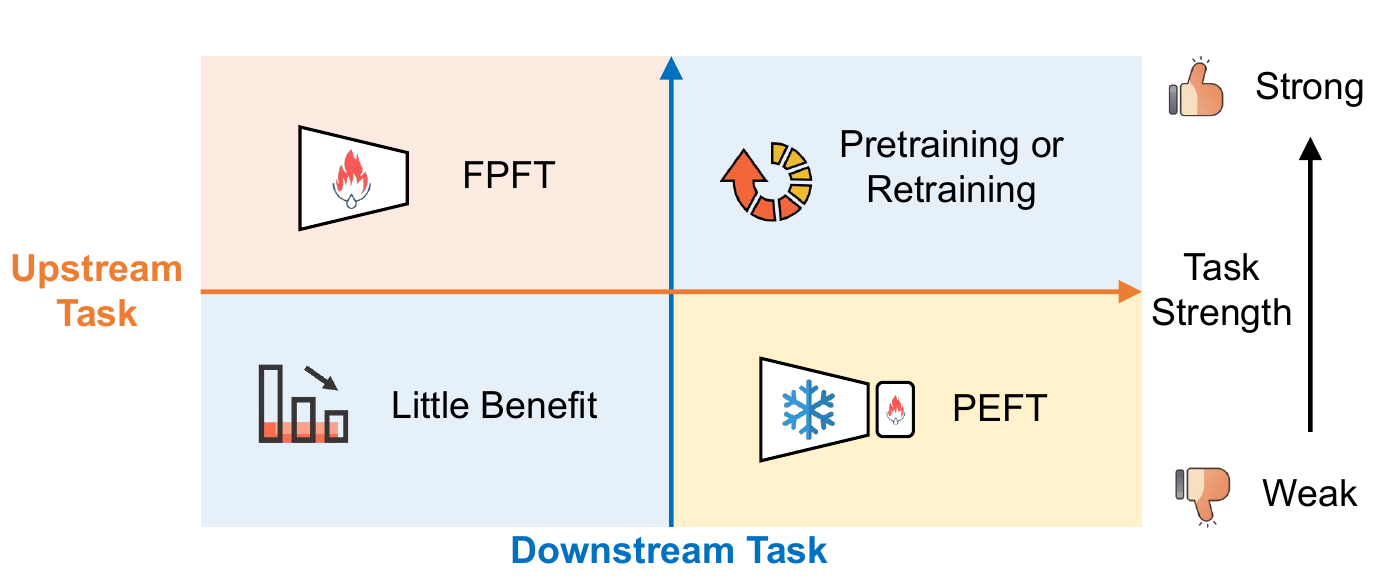}
    \caption{Illustration of fine-tuning choices under different upstream-downstream discrepancies. Parameter-constrained adaptation can be effective when limited downstream adjustment is sufficient, whereas larger discrepancies may benefit from more expressive full-parameter adaptation, as observed in our AQA experiments.}
    \label{fig:strategy_decision_map}
    \vspace{-0.1cm}
\end{figure}

To place these findings in a broader context, we organize fine-tuning strategies according to the relative upstream-downstream discrepancy, as illustrated in \cref{fig:strategy_decision_map}. 
The empirical comparison above is limited to the representative adapter-based PEFT setting used in AQA and is not intended to establish a universal ranking between FPFT and all PEFT methods. We therefore complement this empirical observation with a broader theoretical analysis of subspace-constrained adaptation under explicit assumptions.

\subsection{Theoretical Analysis of FPFT with Feature Replay} \label{sec:theoretical_foundation}
\myPara{Why FPFT Matters for AQA.}
As shown in \cref{sec:empirical_study}, FPFT consistently outperforms the evaluated adapter-based PEFT baseline in AQA. This empirical observation motivates a broader theoretical question: under what conditions can parameter-constrained adaptation limit downstream transfer? 
Rather than extrapolating the empirical result to all practical PEFT methods, \cref{thm:peft-vs-fpft} analyzes a broader class of subspace-constrained adaptation.  It shows that a projection gap can arise when the downstream optimum contains components outside the admissible update subspace, leading to positive excess risk under the stated assumptions, whereas full-space adaptation is not subject to this particular restriction.

\begin{theorem}
\label{thm:peft-vs-fpft}
Let the upstream model $\phi_{\bm{\theta}_{\text{up}}}$ denote a pre-trained AQA scorer on a source domain $\mathcal{D}_{\text{up}}$ (typically large-scale action recognition), and define the downstream task on a target AQA domain $\mathcal{D}_{\text{down}}$ with a distinct data distribution. The goal is to adapt $\phi_{\bm{\theta}_{\text{up}}}$ by minimizing the downstream risk $R_{\text{down}}(\bm{\theta})=\mathbb{E}_{(x,y)\sim\mathcal{D}_{\text{down}}}\ell(\phi_{\bm{\theta}}(x),y)$, where $\bm{\theta}_{\text{down}}^\star=\arg\min_{\bm{\theta}}R_{\text{down}}(\bm{\theta})$. Following the unified view of PEFT methods~\cite{he2021towards}, prompt tuning, adapter tuning, and LoRA can be interpreted within a common parameter-efficient adaptation framework. For theoretical analysis, we model such constrained adaptation through a low-dimensional update subspace $\mathcal{S}$. Taking LoRA as an example, updates are restricted to a low-rank subspace $\mathcal{S}:=\mathrm{range}(\mathbf{U})\subset\mathbb{R}^d$ via $\bm{\theta}=\bm{\theta}_{\text{up}}+\mathbf{U}\bm{\alpha}$, with $\mathbf{U}\in\mathbb{R}^{d\times r}$ and $\mathrm{rank}(\mathbf{U})=r\!\ll\!d$, while FPFT allows full-space updates. The upstream Jacobian $\mathbf{J}_{\text{up}}(x):=\nabla_{\bm{\theta}}\phi_{\bm{\theta}_{\text{up}}}(x)$ measures the local sensitivity of outputs to parameters, and its Neural Tangent Kernel (NTK)-style Gram matrix $\mathbf{\Sigma}_0:=\mathbb{E}_{x\sim\mathcal{D}_{\text{down}}}[\mathbf{J}_{\text{up}}(x)\mathbf{J}_{\text{up}}(x)^\top]\succeq0$ characterizes the curvature and geometry around $\bm{\theta}_{\text{up}}$.
Assume:
\begin{itemize}
\item[(A1)] (\emph{Curvature}) The loss $\ell(\cdot,y)$ is twice differentiable and locally strongly convex,  
i.e., $\ell''(z,y)\ge\mu>0$, ensuring a quadratic lower bound on the risk landscape.
\item[(A2)] (\emph{Nondegeneracy}) 
The curvature of $\mathbf{\Sigma}_0$ on the orthogonal complement $\mathcal{S}^\perp$  
is positive, i.e., $\lambda_{\min}(\mathbf{\Sigma}_0\!\mid_{\mathcal{S}^\perp})>0$, ensuring that useful and stable descent directions exist consistently outside the PEFT subspace.
\item[(A3)] (\emph{Linearization}) 
There exist constants $\rho>0$ and $L_\varepsilon\!\ge\!0$ such that for all small perturbations $\bm{v}$ with $\|\bm{v}\|\!\le\!\rho$,  
$\big|R_{\text{down}}(\bm{\theta}_{\text{up}}+\bm{v})
-\widetilde R_{\text{down}}(\bm{v})\big|
\le\tfrac{L_\varepsilon}{2}\|\bm{v}\|^2$,  
where the linearized risk is  
$\widetilde R_{\text{down}}(\bm{v})
=\mathbb{E}_{(x,y)\sim\mathcal{D}_{\text{down}}}\!
\ell\big(\phi_{\bm{\theta}_{\text{up}}}(x)
+\mathbf{J}_{\text{up}}(x)^\top\bm{v},\,y\big)$,  
and both the restricted and full minimizers,  
$\bm{v}_{\mathcal{S}}=\arg\min_{\bm{v}\in\mathcal{S}}\widetilde R_{\text{down}}(\bm{v})$  
and $\bm{v}^\star=\arg\min_{\bm{v}}\widetilde R_{\text{down}}(\bm{v})$,  
lie within the local region $\|\bm{v}\|\le\rho$.
\end{itemize}
Under these assumptions, for any PEFT parameter $\bm{\alpha}$  
(such that $\bm{v}=\mathbf{U}\bm{\alpha}\in\mathcal{S}$), the following holds:
\begin{equation}
\label{eq:excess-risk-peft-sigma}
R_{\text{down}}(\bm{\theta}_{\text{up}}+\mathbf{U}\bm{\alpha}) - R_{\text{down}}(\bm{\theta}_{\text{down}}^\star)
\ge\tfrac{\mu}{2}\,\|\Pi_{\mathcal{S}^\perp}^{(\mathbf{\Sigma}_0)}\bm{\Delta}\|_{\mathbf{\Sigma}_0}^2 - C_\varepsilon,
\end{equation}
where $\bm{\Delta} := \bm{\theta}_{\text{down}}^\star-\bm{\theta}_{\text{up}}$, 
$\|\bm{w}\|_{\mathbf{\Sigma}_0}^2 := \bm{w}^\top \mathbf{\Sigma}_0 \bm{w}$, 
$\Pi_{\mathcal{S}^\perp}^{(\mathbf{\Sigma}_0)}$ is the $\mathbf{\Sigma}_0$-orthogonal projection onto $\mathcal{S}^\perp$ (well-defined under assumption \textbf{(A2)}), 
and $C_\varepsilon := \tfrac{L_\varepsilon}{2}(\|\bm{v}_{\mathcal{S}}\|^2+\|\bm{v}^\star\|^2)$.
\end{theorem}

\cref{eq:excess-risk-peft-sigma} shows that, under the subspace-constrained adaptation setting,  excess risk arises whenever the downstream optimum $\bm{\theta}_{\text{down}}^\star$ has a nonzero projection onto $\mathcal{S}^\perp$, the complement of its update subspace. The projection term $\|\Pi_{\mathcal{S}^\perp}^{(\mathbf{\Sigma}_0)}\bm{\Delta}\|_{\mathbf{\Sigma}_0}^2$ measures the portion of adaptation inaccessible to PEFT under the curvature geometry of $\mathbf{\Sigma}_0$. In contrast, full-space adaptation is not subject to this specific projection constraint.
The proof of \cref{thm:peft-vs-fpft} is provided in \cref{sec:peft-vs-fpft_proof}.

\myPara{When FPFT Meets Feature Replay.}  
While \cref{thm:peft-vs-fpft} shows the advantage of FPFT over PEFT in static transfer, the CAQA setting requires continual adaptation to evolving distributions.
In practice, feature replay~\cite{yang2023neural} is widely adopted to mitigate forgetting with limited extra memory. However, combining FPFT with feature replay introduces new risks: FPFT may induce overfitting due to large parameter shifts, and historical features stored in memory can drift away from the representations produced by the continually updated backbone. To better understand these challenges, \cref{thm:fpft-replay} formalizes FPFT under replay and establishes stability conditions.  

\begin{theorem}
\label{thm:fpft-replay}
Let $f_{\bm{\theta}_f}:\mathcal{X}\!\to\!\mathbb{R}^m$ be the encoder (backbone), 
$g_{\bm{\theta}_g}:\mathbb{R}^m\!\to\!\mathbb{R}$ be the output head, 
and the full model be $\phi_{\bm{\theta}}:=g_{\bm{\theta}_g}\circ f_{\bm{\theta}_f}$, where $\bm{\theta}=\{\bm{\theta}_f, \bm{\theta}_g\}$.  
At session $t$, parameters update from $\bm{\theta}^{t-1}$ to $\bm{\theta}^{t}$ 
by FPFT on new data $\mathcal{D}_t$ while replaying a memory buffer $\mathcal{M}_{t-1}$.  
We denote the update vector as $\bm{\Delta}_t := \bm{\theta}^t - \bm{\theta}^{t-1}$ 
and its associated step size as $\Delta_t := \|\bm{\Delta}_t\|$.  
Assume:
\begin{itemize}
\item[(B1)] (\emph{Loss Smoothness})  
The loss $\ell(\cdot,y)$ (e.g., MSE or KL divergence) is $1$-Lipschitz continuous with respect to its scalar input, ensuring that small prediction changes lead to bounded loss variation.

\item[(B2)] (\emph{Head Regularity})  
For any parameter $\bm{\theta}_g$, the prediction head $g_{\bm{\theta}_g}$ is $L_g$-Lipschitz continuous with respect to its feature input, limiting sensitivity in score prediction caused by feature perturbations.

\item[(B3)] (\emph{Backbone Stability})  
The backbone mapping $f_{\bm{\theta}_f}$ satisfies  
$\|f_{\bm{\theta}'_f}(x)-f_{\bm{\theta}_f}(x)\|\le L_f\,\|\bm{\theta}'_f-\bm{\theta}_f\|$ 
uniformly for all $x\in\mathcal{X}$ and parameters $\bm{\theta}_f,\bm{\theta}'_f$, 
ensuring representation smoothness during encoder updates.

\item[(B4)] (\emph{Model Continuity})  
The overall mapping $\phi_{\bm{\theta}}=g_{\bm{\theta}_g}\!\circ\! f_{\bm{\theta}_f}$ is uniformly continuous, satisfying  
$\|\phi_{\bm{\theta}'}(x)-\phi_{\bm{\theta}}(x)\|\le L_\phi\,\|\bm{\theta}'-\bm{\theta}\|$, 
which guarantees bounded prediction drift under finite-step updates.
\end{itemize}
Under these regularity and stability conditions, for any past task $k<t$, the expected forgetting satisfies
\begin{equation}
\label{eq:forgetting-bound}
\mathbb{E}\!\left[\psi_t(k)\right]
\;\;\lesssim\;\;
L_g L_f\,\Delta_t \;+\; C\,L_\phi\,\Delta_t \;+\; E_\text{opt},
\end{equation}
where $\psi_t(k)$ denotes the forgetting on task $k$ after session $t$, 
the first term quantifies the replay-drift contribution from feature mismatch, 
the second term reflects the growth of the hypothesis class (for some constant $C>0$), 
and $E_\text{opt}\ge 0$ accounts for the residual optimization error at session $t$.
\end{theorem}

\cref{eq:forgetting-bound} shows that FPFT’s flexibility comes at the cost of stability: large updates $\Delta_t$ amplify both replay drift ($L_gL_f\Delta_t$) and generalization growth ($C L_\phi\Delta_t$), leading to potential overfitting and forgetting. Moreover, stale features from the old encoder deviate from the updated one, weakening replay supervision. 
These findings establish the general principle that continual FPFT should control parameter-induced representation drift while preserving sufficient adaptation. MAGR++ instantiates this principle through layer-adaptive tuning and feature rectification, which respectively regulate representation updates and reduce replay-feature drift. A detailed proof of \cref{thm:fpft-replay} is deferred to \cref{sec:proof-fpft-replay}.

\begin{figure}
    \centering
    \includegraphics[width=\linewidth,clip,trim=140 80 140 76]{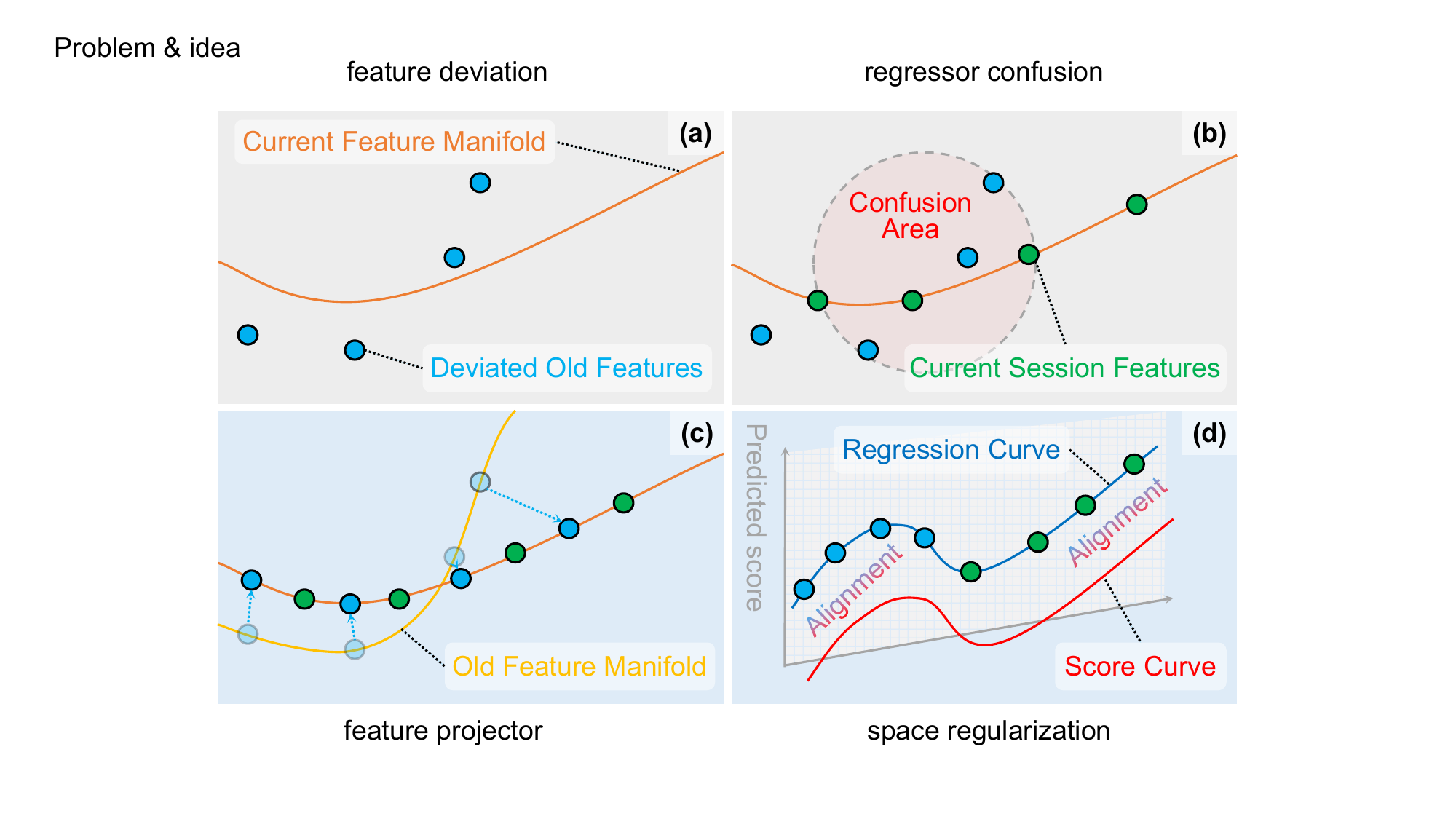}
    \caption{
        Core idea of MAGR++: 
        (a) Old features (\textcolor{deviatedBlue}{blue circles}) deviate from the current manifold (\textcolor{currentOrange}{orange curve}) due to manifold shift; 
        (b) Mixing old and new features (\textcolor{sessionGreen}{green circles}) leads to confusion in score regression; 
        (c) The manifold projector translates old features from the previous manifold (\textcolor{oldYellow}{yellow curve}) to the current one; 
        (d) The feature space is further aligned with the quality score space.
    }
    \label{fig:idea}{
    \phantomsubcaption\label{fig:idea-a}\phantomsubcaption\label{fig:idea-b}\phantomsubcaption\label{fig:idea-c}\phantomsubcaption\label{fig:idea-d}}\end{figure}

\begin{remark}[Extension under Weaker Assumptions]
\label{rem:weaker-assumption}
We extend the analysis of \cref{thm:peft-vs-fpft,thm:fpft-replay} and show that analogous qualitative conclusions hold under weaker, trajectory-local regularity.
For the projection-gap result in \cref{thm:peft-vs-fpft}, the excess-risk lower bound relies on two essential components: (i) a curvature-controlled growth condition and (ii) a nonzero component of the downstream displacement outside the PEFT subspace $\mathcal{S}$. 
Global strong convexity \textbf{(A1)} can therefore be relaxed to a local Polyak-\L{}ojasiewicz (PL) condition near $\bm{\theta}_{\text{up}}$, which still guarantees gradient-dominance in a neighborhood of interest. 
Similarly, the linearization assumption \textbf{(A3)} can be weakened to a second-order remainder control
$
R_{\text{down}}(\bm{\theta}_{\text{up}}+\bm v)
=
\widetilde R_{\text{down}}(\bm v)
+
\mathcal{O}(\|\bm v\|^2),
$
so that the leading quadratic term governing the projection gap remains unchanged up to higher-order residuals. 
The non-degeneracy requirement \textbf{(A2)} only needs to hold directionally: it suffices that 
$\bm{\theta}_{\text{down}}^\star \notin \bm{\theta}_{\text{up}}+\mathcal{S}$
under the $\mathbf{\Sigma}_0$-induced geometry.
Under these relaxed conditions, the projection-gap structure of \cref{eq:excess-risk-peft-sigma} is preserved up to constants and higher-order terms.
For the forgetting bound in \cref{thm:fpft-replay}, the derivation depends on controlling prediction drift via the update magnitude 
$\Delta_t=\|\bm{\theta}^t-\bm{\theta}^{t-1}\|$. 
Replacing the global continuity assumption \textbf{(B4)} with trajectory-local Lipschitzness of $\phi_{\bm{\theta}}$ around $\bm{\theta}^{t-1}$ still ensures that prediction drift scales proportionally with $\Delta_t$, yielding
$
\mathbb{E}[\psi_t(k)] = \mathcal{O}(\Delta_t) + E_{\text{opt}}.
$
If additional perturbations are injected into $\bm{\theta}_f^{t-1}$ as in \cref{eq:mutation}, the effective parameter shift increases beyond the optimization-induced $\Delta_t$, and the bound accordingly predicts amplified forgetting. 
Conversely, when perturbations remain small, the induced shift is limited and forgetting growth is correspondingly controlled. 
This theoretical dependence on the effective parameter shift is empirically validated in \cref{fig:mutation}.
\end{remark}

\begin{figure*}
    \centering
    \includegraphics[width=\linewidth,clip,trim=0 75 0 90]{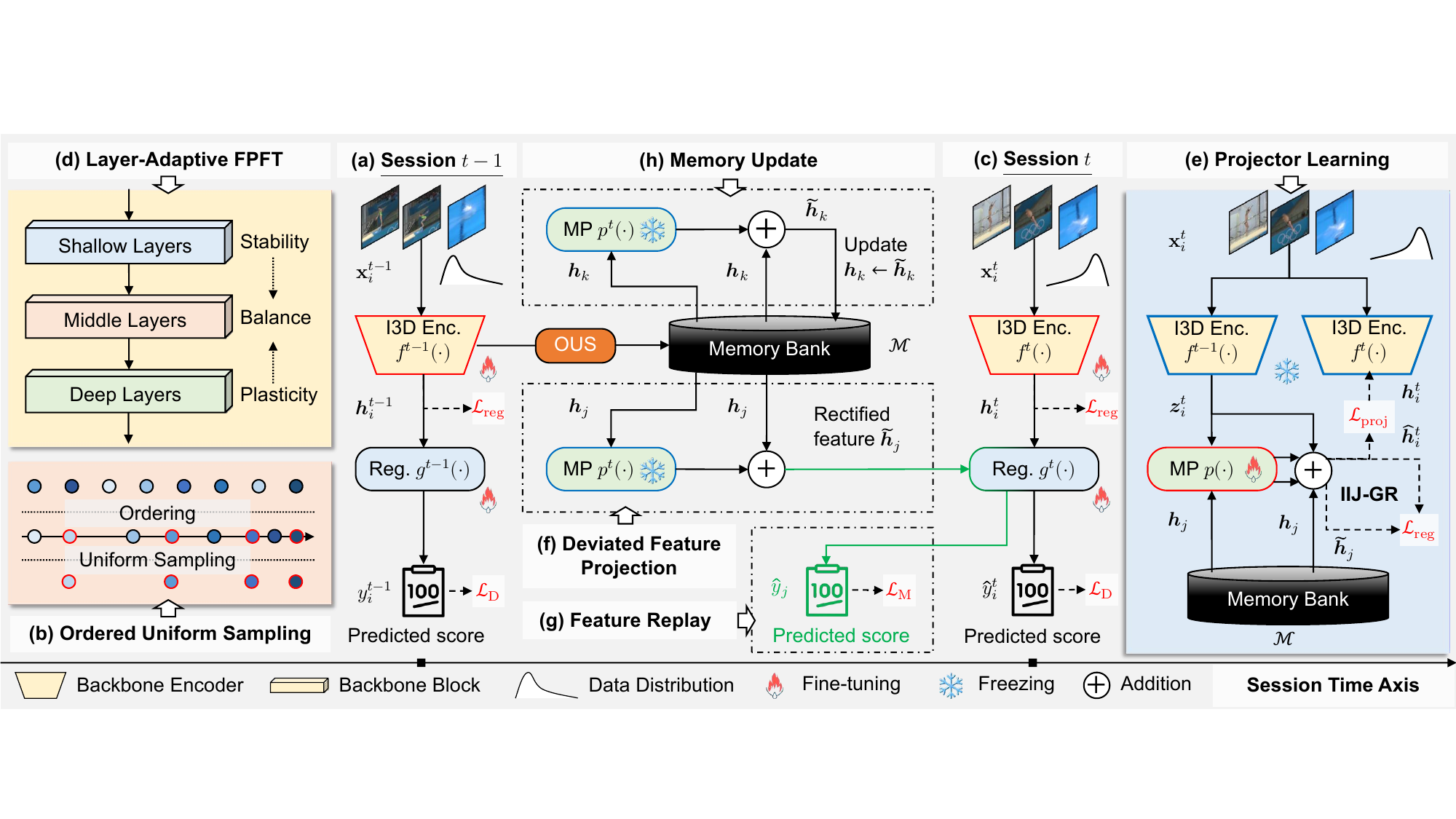}
    \MSfigcaption{fig:framework}{Overview of MAGR++. At the end of the session $t-1$ \subref{fig:framework-a}, representative features are selected via Ordered Uniform Sampling (OUS, \subref{fig:framework-b}) and stored in the memory bank $\mathcal{M}$. At the start of the session $t$ \subref{fig:framework-c}, the backbone is adapted with layer-adaptive FPFT \subref{fig:framework-d} to balance stability and plasticity. A Manifold Projector (MP) is then trained \subref{fig:framework-e} to align old features with the evolving feature space \subref{fig:framework-f}, enabling effective replay and adaptation \subref{fig:framework-g}. Finally, the memory bank is refreshed with rectified old features and newly sampled prototypes \subref{fig:framework-h}.}{
    \phantomsubcaption\label{fig:framework-a}\phantomsubcaption\label{fig:framework-b}\phantomsubcaption\label{fig:framework-c}\phantomsubcaption\label{fig:framework-d}\phantomsubcaption\label{fig:framework-e}\phantomsubcaption\label{fig:framework-f}\phantomsubcaption\label{fig:framework-g}\phantomsubcaption\label{fig:framework-h}}
\vspace{-0.5cm}
\end{figure*}

\myPara{Practical Implications for MAGR++.}
\cref{thm:peft-vs-fpft} characterizes a potential limitation of subspace-constrained adaptation when the downstream-optimal displacement contains components outside the admissible update subspace, while \cref{thm:fpft-replay}further reveals two crucial risks of expressive full-parameter adaptation in CAQA: (i) FPFT may suffer from overfitting as the hypothesis class grows across sessions, and (ii) feature replay is vulnerable to representation drift when old features become misaligned with the updated backbone. 
In the original MAGR~\cite{zhou2024magr}, synchronous projector training exacerbates this misalignment, leading to unstable replay correction. 

Notably, the implications of \cref{thm:peft-vs-fpft} extend beyond simply reducing update magnitudes. Existing globally constrained stabilization strategies, such as EWC~\cite{james2017ewc} and slow learning-rate adaptation in SLCA~\cite{zhang2023slca}, still exhibit limited performance in \cref{tab:online,tab:offline}, suggesting that uniformly restricting updates may fail to simultaneously preserve fine-grained quality-sensitive cues and continual adaptability in AQA. As shown in \cref{fig:cluster}, constraining shallow layers preserves robust performance, whereas deeper layers remain critical for adapting quality-aware representations, revealing heterogeneous adaptation sensitivity across network stages. 
Motivated by these observations, MAGR++ introduces \textbf{two key improvements}: 
(i) layer-adaptive fine-tuning, which constrains updates on low-level layers while fully tuning high-level ones, thereby balancing stability and adaptability; and 
(ii) asynchronous feature rectification, which allows the backbone and projector to converge before replay correction, preventing noisy updates and ensuring robust continual adaptation.

\section{Adaptive Manifold-Aligned Graph Regularization (MAGR++)} \label{sec:method}

In this section, we first introduce the task definition, motivation, and overall architecture, and then describe the core modules. The training strategy is presented in \cref{sec:training}.

\subsection{Motivation and Framework Overview}

\myPara{Addressing CAQA Challenges with MAGR++.}
As discussed in \cref{sec:foundation}, CAQA poses a dilemma for CL: fine-tuning the backbone is essential for adapting to new sessions, yet it inevitably causes feature manifold shift and catastrophic forgetting. As illustrated in \cref{fig:idea}, this shift causes stored feature prototypes (blue circles) to deviate from the new feature manifold (orange curve), making them ineffective for replay (see \cref{fig:idea-a}) and likely to confuse the score regressor when mixed with current features (see \cref{fig:idea-b}). These challenges undermine replay-based methods and highlight the need for shift-aware adaptation. To address this, MAGR++ introduces a two-step solution: (i) mapping old features from their previous manifold (yellow curve) onto the current session’s manifold (see \cref{fig:idea-c}); and (ii) readjusting the translated distribution for optimal alignment with target scores (see \cref{fig:idea-d}).

\myPara{Framework Overview.}
\cref{fig:framework} depicts the overall pipeline of MAGR++. 
At the end of session $t-1$ (see \cref{fig:framework-a}), Ordered Uniform Sampling (OUS, see \cref{fig:framework-b}) stores representative features in the memory bank $\mathcal{M}$ by sorting samples by quality scores and uniformly selecting them across the score range, ensuring diverse coverage. 
At the beginning of each new session $t$ (see \cref{fig:framework-c}), the backbone is adapted to the new data distribution via layer-adaptive FPFT (see \cref{fig:framework-d}).
Next, the Manifold Projector (MP) is trained to align session-wise feature spaces (see \cref{fig:framework-e}) by using paired features from the frozen backbone $f^{t-1}$ and the updated backbone $f^t$, learning a mapping from the old space to the new one. Additionally, an Intra-Inter-Joint Graph Regularization (IIJ-GR) promotes feature alignment across sessions. The regressor is jointly trained on rectified old features and new features (see \cref{fig:framework-g}), enabling adaptation to new data while retaining previously acquired knowledge.
Finally, after regressor training, old features are first updated via MP to align with the current feature space (see \cref{fig:framework-h}), and new prototypes from session $t$ are chosen and added to $\mathcal{M}$. 

Formally, at session $t$, MAGR++ optimizes the backbone ($\bm{\theta}_f^t$), regressor ($\bm{\theta}_g^t$), and manifold projector ($\bm{\theta}_p^t$) by minimizing the following composite training objective:
\begin{equation}
\begin{aligned}
    \min_{\bm{\theta}_f^t,\,\bm{\theta}_g^t,\,\bm{\theta}_p^t}\ 
    &\ \mathcal{L}_{\text{D}} + \mathcal{L}_{\text{M}} + \lambda_{\text{tune}}\,\mathcal{L}_{\text{tune}} + \lambda_{\text{proj}}\,\mathcal{L}_{\text{proj}} + \lambda_{\text{reg}}\,\mathcal{L}_{\text{reg}} \\
    \text{s.t.} \quad
    &\ \hat{y}_i^t = g^t(f^t(\mathbf{x}_i^t)), ~~ (\mathbf{x}_i^t, y_i^t) \in \mathcal{D}^t_{\text{train}}, \\
    &\ \hat{y}_j = g^t(p^t(\bm{h}_j)), ~~ (\bm{h}_j, y_j) \in \mathcal{M}.
\end{aligned}
\label{eq:magrpp-obj}
\end{equation}
Here, $\mathcal{L}_{\text{D}}$ and $\mathcal{L}_{\text{M}}$ follow \cref{eq:caqa}, $\mathcal{L}_{\text{tune}}$ is the FPFT loss (see \cref{eq:la} in \cref{sec:layer_adaptive}), and $\mathcal{L}_{\text{proj}}$ (see \cref{eq:proj} in \cref{sec:mp}) and $\mathcal{L}_{\text{reg}}$ (see \cref{eq:iij} in \cref{sec:iij_gr}) regularize the projector and regressor, respectively. The coefficients $\lambda_{\text{tune}}$, $\lambda_{\text{proj}}$, and $\lambda_{\text{reg}}$ balance the corresponding loss terms.

\subsection{Stability-Plasticity Balance via Layer-Adaptive FPFT}
\label{sec:layer_adaptive}
A central challenge in CAQA is the stability-plasticity dilemma: FPFT provides strong adaptation to new sessions but may also amplify overfitting and catastrophic forgetting.
 As indicated in \cref{thm:fpft-replay}, stable continual adaptation therefore requires controlling representation updates without excessively suppressing plasticity. A uniform constraint over the entire backbone, however, may be suboptimal because different layers exhibit heterogeneous adaptation behavior. Motivated by prior findings in representation learning~\cite{datta2025deep}, we adopt the common hierarchical interpretation of pretrained backbones, where shallower layers tend to encode lower-level spatial cues, intermediate layers capture increasingly semantic information, and deeper layers become more task-specific. We use this hierarchy as an interpretative motivation for layer-wise adaptation rather than as a strict functional decomposition of the backbone. Based on this observation, we propose \textbf{layer-adaptive FPFT} (see \cref{fig:la_fpft}), which imposes stronger stability constraints on shallower layers while allowing deeper layers to adapt more freely. The adaptive layer-selection criterion determines the transition boundary from base-session statistics, while constrained full tuning regulates updates below this boundary.
This design provides a layer-wise realization of  \cref{thm:fpft-replay}, rather than uniformly restricting adaptation across the backbone. Existing results further support this choice: static alternatives such as a fixed backbone and standard FPFT yield inferior stability-adaptation trade-offs (see \cref{tab:ablation_mtl-a}), while conservative adaptation strategies such as SLCA remain substantially below MAGR++ across both offline and online settings. Together with the layer-wise analysis in \cref{fig:cluster}, these observations motivate adaptive control across backbone depth instead of a uniform adaptation policy. 
\begin{figure}
    \centering
    \includegraphics[width=\linewidth,clip,trim=165 135 165 160]{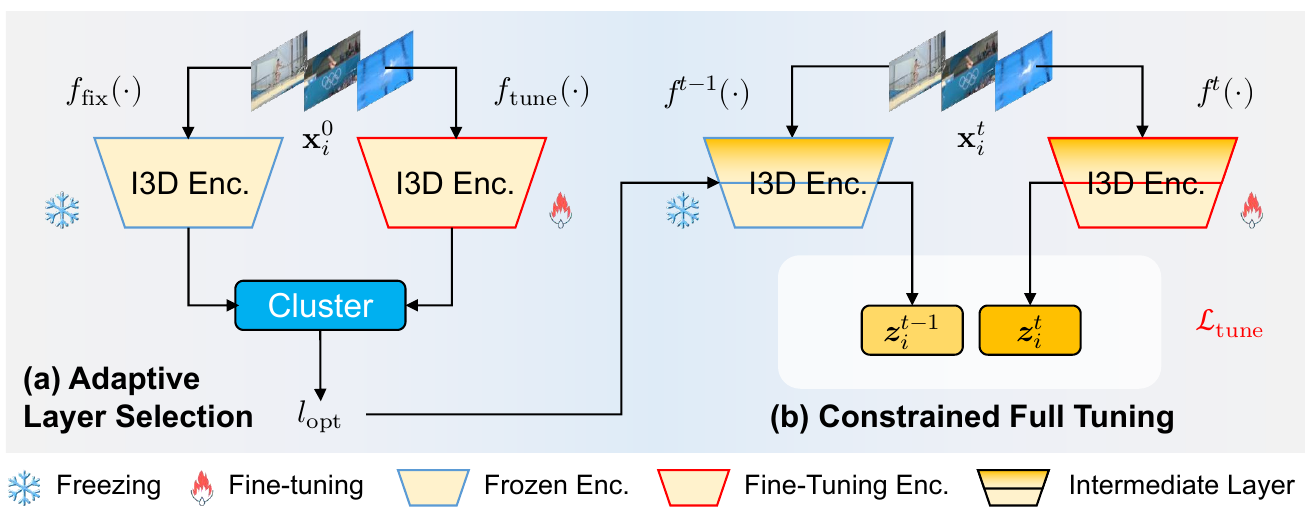}
    \MSfigcaption{fig:la_fpft}{Illustration of the proposed layer-adaptive FPFT strategy. (a) Adaptive layer selection determines the optimal layer boundary $l_{\mathrm{opt}}$. (b) Constrained full tuning stabilizes shallow layers while preserving the adaptability of deeper quality-aware representations.}{
\phantomsubcaption\label{fig:la_fpft-a}
    \phantomsubcaption\label{fig:la_fpft-b}
    }
\end{figure} 

\myPara{Adaptive Layer Selection.}  
The key to layer-adaptive FPFT lies in identifying the optimal boundary between stable and adaptive layers.  
Exhaustive grid search is infeasible since it requires future session data and incurs prohibitive cost. 
Instead, we exploit only base-session data to guide boundary selection, preventing information leakage.  
As shown in \cref{fig:la_fpft-a}, we evaluate the abstraction degree encoded at each layer by treating clustering quality as a proxy for feature abstraction, denoted as \(\mathcal{C}^l\) for the \(l\)-th layer. Shallower layers dominated by low-level appearance cues generally yield noisy and less separable structures. In contrast, deeper layers encode more abstract and quality-aware features that are easier to cluster by action scores. 
For each layer \(l\), we compute the Davies-Bouldin index from both the frozen backbone \(f_{\text{fix}}\) and the fine-tuned backbone \(f_{\text{tune}}\), and define the abstraction ratio as  
$
r^l = \mathcal{C}^l_{\text{tune}} / \mathcal{C}^l_{\text{fix}}.
$
The Davies-Bouldin index is adopted here as a stable measure that jointly captures intra-cluster compactness and inter-cluster separability, while alternative clustering metrics are used in the experimental analysis to validate the robustness of this choice (see \cref{fig:cluster}).  
The key intuition is that \(r^l \leq 1\) indicates no gain, or even degradation, from fine-tuning, suggesting that the layer is still dominated by low-level cues, whereas \(r^l > 1\) signifies improved separability and abstraction through fine-tuning. Accordingly, the optimal boundary is determined as:
\begin{equation}
l_{\text{opt}} = \min \big\{ l \in \{1,\dots,L\} \;|\; r^l > 1 + \epsilon \big\},
\label{eq:Lopt}
\end{equation}
where \(\epsilon\) is a small margin (e.g., 0.05) introduced to enhance robustness against noise. Layers below \(l_{\text{opt}}\) are constrained to remain stable, while deeper layers are fully fine-tuned to adapt to evolving quality distributions. This principled criterion eliminates future-data dependence, avoids costly grid search, and provides a robust mechanism for balancing stability and adaptability in CAQA.
As shown in \cref{fig:cluster}, the layer boundaries selected by our method are consistent with those obtained via exhaustive grid search, verifying its effectiveness.  

\myPara{Constrained Full Tuning.}  
During the current session $t$, we constrain the backbone's updates for layers below $l_{\text{opt}}$ when adapting the backbone $f^t$ (see \cref{fig:la_fpft-b}).  
Given current session data $\mathbf{x}_i^t$, we obtain paired features $\bm{z}_i^{t,l}$ from $f^t$ and $\bm{z}_i^{t-1,l}$ from the previous backbone $f^{t-1}$ at layer $l<l_{\text{opt}}$.  
We then enforce consistency between features across sessions using a feature-matching loss, defined as follows: 
\begin{equation} \label{eq:la}
\mathcal{L}_{\text{tune}} = \frac{1}{N_t}\sum_{i=1}^{N_{t}}\sum_{l < l_{\text{opt}}} 
\big\| \bm{z}_i^{t,l} - \bm{z}_i^{t-1,l} \big\|_2^2,
\end{equation}
where $N_t$ is the number of samples in the current session and $\bm{z}_i^{t,l}$ denotes the feature from layer $l$ at session $t$.  
Unlike freezing features entirely, this soft constraint still permits shallow layers to undergo limited adaptation, as suggested by \cref{fig:cluster}, while effectively preventing uncontrolled drift.  
This design stabilizes generic low-level representations without suppressing their necessary refinement, thereby balancing stability and adaptability during FPFT. Interestingly, our experiments further show that applying the constraint only at layer $l_{\text{opt}}-1$ is sufficient to achieve stable performance, offering a lightweight yet effective regularization strategy.

\subsection{Deviated Feature Translation via Manifold Projector} \label{sec:mp}

As identified in \cref{sec:foundation}, CAQA suffers from manifold shift under FPFT, as features extracted from earlier sessions become inconsistent with the updated backbone, causing replayed samples to misalign with the current representation and degrade regressor performance. 
Existing strategies address representation drift from different perspectives. Experience replay~\cite{buzzega2020dark} retains raw inputs and may raise storage and privacy concerns, while backbone freezing~\cite{yang2023neural} preserves representation stability at the cost of limited adaptation. 
More closely related to our setting, learnable drift-compensation methods estimate mappings between consecutive feature spaces and use them to update historical representations~\cite{gomez2024exemplar}. Our method builds on this general principle rather than claiming feature-space drift compensation itself as novel. Specifically, we model only the residual representation shift induced by continual adaptation and perform rectification within an asynchronous replay adaptation framework. Compared with direct full-space transformation, this residual formulation avoids relearning unchanged feature components and achieves better stability and accuracy in our CAQA experiments (see \cref{tab:ablation_mtl-b}). We therefore introduce the \textbf{Manifold Projector (MP)}, which estimates the residual manifold shift between adjacent sessions and translates deviated historical features into the current representation space without requiring raw inputs.

\myPara{Projector Learning.}
The key to MP lies in estimating the manifold shift without accessing raw data from previous sessions. We cast this as a self-supervised prediction problem using only current-session inputs. As shown in \cref{fig:framework-e}, at the start of session $t$ we freeze the previous backbone $f^{t-1}$ and compute initial features $\bm{z}_j^t=f^{t-1}(\mathbf{x}_j^t)$ for each current sample $\mathbf{x}_j^t$. We then train a projector $p(\cdot)$ to predict the updated features produced by the adapting backbone $f^t$ as follows:
\begin{equation}\label{eq_mp_learning}
\bm{\hat{h}}_j^t=\bm{z}_j^t+p(\bm{z}_j^t),
\end{equation}
where the residual connection stabilizes optimization. 
The projector is optimized by minimizing the discrepancy between the predicted features $\bm{\hat{h}}_j^t$ and the actual updated features $\bm{h}_j^t = f^t(\mathbf{x}_j^t)$ as follows:
\begin{equation} \label{eq:proj}
\mathcal{L}_{\text{proj}} = \frac{1}{N_t} \sum_j \| \bm{h}_j^t - \bm{\hat{h}}_j^t \|_2^2,
\end{equation}
where $N_t$ is the number of samples in session $t$, and $\|\cdot\|_2^2$ denotes the mean squared error. 
In parallel, old features fetched from the memory bank are updated via the projector and used to compute a regularization loss with respect to $\bm{\hat{h}}_j^t$, further constraining projector learning (detailed in \cref{sec:iij_gr}).  
This design enables the projector to capture representation shifts effectively without requiring access to old data.

\myPara{Deviated Feature Projection.}  
Once trained with sufficient coverage of the current session data, the projector is applied to translate old features from previous sessions into the current representation space.  
For an old feature $\bm{h}_i^s$ stored in memory ($s < t$), the corrected feature is computed as:
\begin{equation} \label{eq_fp}
\bm{h}_i^s \leftarrow \bm{h}_i^s + p(\bm{h}_i^s).
\end{equation}
This translation aligns old features with the updated manifold, ensuring stable replay that alleviates catastrophic forgetting.

\subsection{Feature Alignment via Intra-Inter-Joint Graph Regularizer} \label{sec:iij_gr}

\begin{figure}
    \centering
    \includegraphics[width=\linewidth,clip,trim=95 165 120 140]{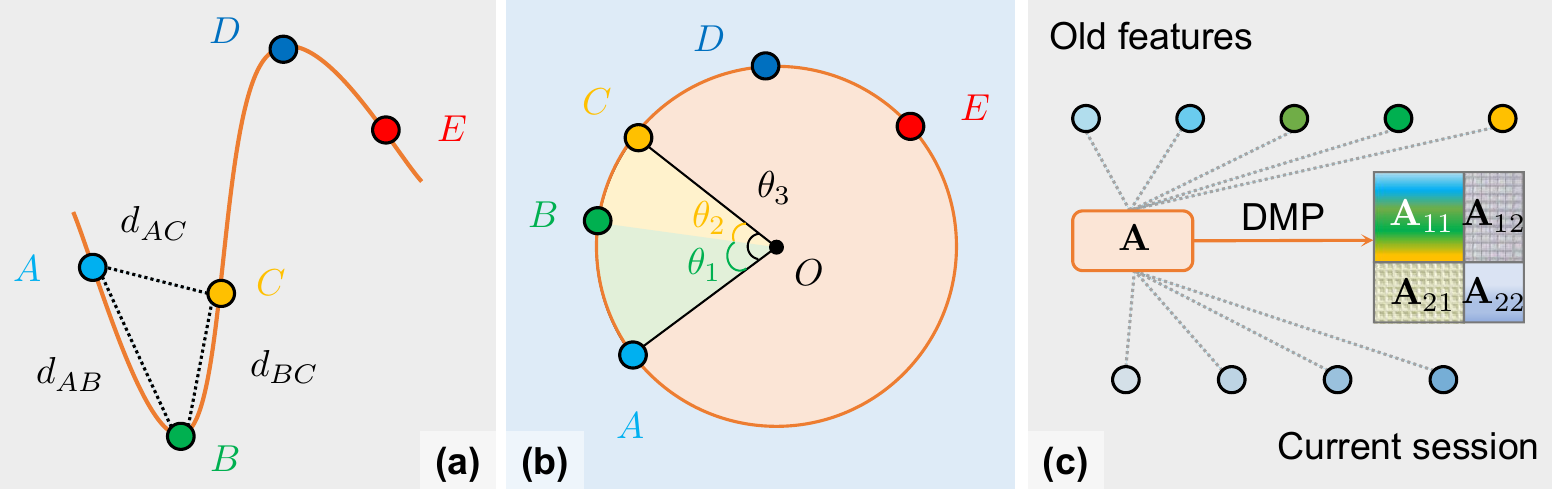}
    \caption{Illustrations of IIJ-GR: \subref{fig:iij_gr-a} Euclidean distance, \subref{fig:iij_gr-b} Angular distance, and \subref{fig:iij_gr-c} Distance Matrix Partitioning (DMP).}
    {
        \phantomsubcaption\label{fig:iij_gr-a}\phantomsubcaption\label{fig:iij_gr-b}\phantomsubcaption\label{fig:iij_gr-c}}
    \label{fig:iij_gr}
    \vspace{-0.2cm}
\end{figure}

While MP translates old features into the current representation space, it does not ensure that the overall feature distribution remains aligned with the quality score distribution (see \cref{fig:idea-c}). Existing graph-based CL methods \cite{tao2020few} often rely on Euclidean distances, which fail to capture the geodesic structure of quality relationships and thus distort feature–score alignment (see \cref{fig:iij_gr-a}). Furthermore, features from different sessions may suffer from inconsistent scaling, which disrupts the relative ordering of quality scores and confuses regression. To address these issues, we propose the \textbf{Intra-Inter-Joint Graph Regularizer (IIJ-GR)}, which explicitly enforces both local (intra-session) and global (inter-session) consistency between the feature and score spaces, as shown in \cref{fig:iij_gr}. The key innovation lies in leveraging angular distances on a unit hypersphere (see \cref{fig:iij_gr-b}), together with a distance matrix partitioning strategy (see \cref{fig:iij_gr-c}) that provides a principled mechanism to preserve the geometric structure of quality relationships across sessions, thereby enhancing both stability and accuracy in CAQA.

\myPara{Distance Matrix Partitioning.}  
Given a mini-batch of old features $\bm{h}_i^s$ from the memory bank and $\bm{h}_j^t$ from the current session, we form a joint feature matrix 
\(\mathbf{H} = [\bm{h}_1^s, \dots, \bm{h}_{b_1}^s, \bm{h}_1^t, \dots, \bm{h}_{b_2}^t]\).  
All features are normalized onto a unit hypersphere, where each row of \(\tilde{\mathbf{H}}\) has unit length. Their pairwise angular distances are:
\begin{equation}
\mathbf{A} = \arccos\!\left(\tilde{\mathbf{H}} \tilde{\mathbf{H}}^\top \right).
\end{equation}
Compared to Euclidean distance, angular distance better preserves the geodesic structure of semantic similarity \cite{bao2023pcfgaze}, which is crucial for reflecting fine-grained quality cues.  
The resulting matrix $\mathbf{A}$ is then partitioned into four sub-blocks corresponding to intra-session relations in past data, intra-session relations in current data, and inter-session relations across the two.  
This partitioning disentangles local (within-session) and global (cross-session) dependencies, enabling structured supervision that balances stability and adaptability in distribution alignment.

\begin{table*}[h!]
    \centering
    \setlength{\tabcolsep}{3pt}
    \MStabcaption{tab:offline}{\textbf{Offline} performance comparison. The primary metric is $\rho_{\text{avg}}$, with Joint Training (JT) as the Upper Bound (UB) and Sequential Fine-Tuning (SFT) as the Lower Bound (LB). Best results are highlighted in bold. $\uparrow$/$\downarrow$ indicate higher/lower is better.}{}
    \vspace{0.2cm}
\begin{minipage}{0.50\linewidth}
        \centering
        \subcaption{MTL-AQA (default, w/o difficulty label)}
        \vspace{-1.5mm}
        \resizebox{\linewidth}{!}{
        \begin{tabular}{l r c cc S[table-format=+1.4, input-symbols={\textbf}] c}
        \toprule
        \textbf{Method} & \textbf{Publisher} & \textbf{Memory} & $\rho_{\text{avg}}$ ($\uparrow$) & $\rho_{\text{aft}}$ ($\downarrow$) & \multicolumn{1}{c}{$\rho_{\text{fwt}}$ ($\uparrow$)} & RL2 ($\downarrow$) \\
        \midrule
        \rowcolor{rowgrayblue}JT (UB) & -- & None & 0.9360 & -- & {--} & 0.0038 \\
        \rowcolor{rowgrayblue}SFT (LB) & -- & None & 0.5458 & 0.1524 & 0.0538 & 0.0226 \\
        \rowcolor{rowgraygreen}SI \cite{zenke2017continual} & ICML'17 & None & 0.5526 & 0.2677 & 0.0350 & 0.0235  \\
        \rowcolor{rowgraygreen}EWC \cite{james2017ewc} & PNAS'17 & None & 0.2312 & 0.1553 & 0.0343 & 0.0608 \\
        \rowcolor{rowgraygreen}LwF \cite{li2017learning} & TPAMI'17 & None & 0.4581 & 0.1894 & 0.0490 & 0.0260 \\
        \rowcolor{rowgrayyellow}MER \cite{riemer2019learning} & ICLR'19 & Raw Data & 0.8720 & 0.1303 & 0.0625 & 0.0075 \\
        \rowcolor{rowgrayyellow}DER++ \cite{buzzega2020dark} & NeurIPS'20 & Raw Data & 0.8334 & 0.1775 & 0.0433 & 0.0082 \\
        \rowcolor{rowgrayyellow}TOPIC \cite{tao2020few} & CVPR'20 & Raw Data & 0.7693 & 0.1427 & 0.1391 & 0.0121 \\
        \rowcolor{rowgrayyellow}GEM \cite{kukleva2021generalized} & ICCV'21 & Raw Data & 0.8583 & 0.0950 & 0.1429 & 0.0075 \\
        \rowcolor{rowgrayyellow}STAR \cite{eskandar2025star} & ICLR'25 & Raw Data & 0.8935 & 0.0891 & 0.1832 & 0.0070 \\
        \rowcolor{rowgraypink}Feature MER & -- & Feature & 0.7283 & 0.2255 & 0.0535 & 0.0133 \\
        \rowcolor{rowgraypink}SLCA \cite{zhang2023slca} & ICCV'23 & Feature & 0.7223 & 0.1852 & 0.1665 & 0.0127 \\
        \rowcolor{rowgraypink}NC-FSCIL \cite{yang2023neural} & ICLR'23 & Feature & 0.8426 & 0.1146 & 0.0718 & 0.0084 \\
        \rowcolor{rowgraypink}FS-Aug \cite{li2024continual} & TCSVT'24 & Feature & 0.8060 & 0.1456 & 0.0790 & 0.0116 \\
        \rowcolor{rowgraypink}MAGR \cite{zhou2024magr} & ECCV'24 & Feature & 0.8979 & 0.0223 & \bf 0.1914 & 0.0068 \\
        \rowcolor{rowgraypink}ProNC \cite{wang2026rethinking} & ICLR'26 & Feature & 0.8976 & 0.1722 & 0.1784 & 0.0071 \\
        \rowcolor{rowgraypink}MAGR++ (Ours) & -- & Feature & \bf 0.9205 & \bf 0.0103 & 0.1274 & \bf 0.0064 \\
        \bottomrule
        \end{tabular}
        }
    \end{minipage}\hfill
\begin{minipage}{0.50\linewidth}
        \centering
        \subcaption{MTL-AQA (w/ difficulty)}
        \vspace{-1.5mm}
        \resizebox{\linewidth}{!}{
        \begin{tabular}{l r c cc S[table-format=+1.4, input-symbols={\textbf}] c}
        \toprule
        \textbf{Method} & \textbf{Publisher} & \textbf{Memory} & $\rho_{\text{avg}}$ ($\uparrow$) & $\rho_{\text{aft}}$ ($\downarrow$) & \multicolumn{1}{c}{$\rho_{\text{fwt}}$ ($\uparrow$)} & RL2 ($\downarrow$) \\
        \midrule
        \rowcolor{rowgrayblue}JT (UB) & -- & None & 0.9587 & -- & {--} & 0.0023 \\
        \rowcolor{rowgrayblue}SFT (LB) & -- & None & 0.8684 & 0.1418 & 0.2282 & 0.0078 \\
        \rowcolor{rowgraygreen}SI \cite{zenke2017continual} & ICML'17 & None & 0.8678 & 0.2050 & 0.2491  & 0.0073 \\
        \rowcolor{rowgraygreen}EWC \cite{james2017ewc} & PNAS'17 & None & 0.8625 & 0.1267 & 0.1776 & 0.0074 \\
        \rowcolor{rowgraygreen}LwF \cite{li2017learning} & TPAMI'17 & None & 0.7852 & 0.1501 & 0.0912 & 0.0043 \\
        \rowcolor{rowgrayyellow}MER \cite{riemer2019learning} & ICLR'19 & Raw Data & 0.9234 & 0.0832  & 0.3089 & 0.0043 \\
        \rowcolor{rowgrayyellow}DER++ \cite{buzzega2020dark} & NeurIPS'20 & Raw Data & 0.9037 & 0.1230 & 0.3122 & 0.0078 \\
        \rowcolor{rowgrayyellow}TOPIC \cite{tao2020few} & CVPR'20 & Raw Data & 0.8782 & 0.1394 & 0.2304 & 0.0072 \\
        \rowcolor{rowgrayyellow}GEM \cite{kukleva2021generalized} & ICCV'21 & Raw Data & 0.8873 & 0.1707 & 0.3127 & 0.0072 \\
        \rowcolor{rowgrayyellow}STAR \cite{eskandar2025star} & ICLR'25 & Raw Data & 0.9080 & 0.0791 & 0.3210 & 0.0058 \\
        \rowcolor{rowgraypink}Feature MER & -- & Feature & 0.8785 & 0.2130 & 0.2436 & 0.0077  \\
        \rowcolor{rowgraypink}SLCA \cite{zhang2023slca} & ICCV'23 & Feature & 0.6885 & 0.2029 & 0.0958 & 0.0182  \\
        \rowcolor{rowgraypink}NC-FSCIL \cite{yang2023neural} & ICLR'23 & Feature & 0.9034 & 0.0878 & 0.1456 & 0.0055 \\
        \rowcolor{rowgraypink}FS-Aug \cite{li2024continual} & TCSVT'24 & Feature & 0.9136 & 0.1280 & 0.3145 & 0.0050  \\
        \rowcolor{rowgraypink}MAGR \cite{zhou2024magr} & ECCV'24 & Feature &  0.9237 & 0.0615 & 0.1944 & 0.0046  \\
        \rowcolor{rowgraypink}ProNC \cite{wang2026rethinking} & ICLR'26 & Feature & 0.9134 & 0.0506 & 0.3063 & 0.0048 \\
        \rowcolor{rowgraypink}MAGR++ (Ours) & -- & Feature & \bf 0.9383 & \bf 0.0181 & \bf 0.3222 & 0.0043 \\
        \bottomrule
        \end{tabular}
        }
    \end{minipage}\hfill

\begin{minipage}{0.50\linewidth}
        \centering
        \vspace{2mm}
        \subcaption{FineDiving (w/ dive number)}
        \vspace{-1.5mm}
        \resizebox{\linewidth}{!}{
        \begin{tabular}{l r c cc S[table-format=+1.4, input-symbols={\textbf}] c}
        \toprule
        \textbf{Method} & \textbf{Publisher} & \textbf{Memory} & $\rho_{\text{avg}}$ ($\uparrow$) & $\rho_{\text{aft}}$ ($\downarrow$) & \multicolumn{1}{c}{$\rho_{\text{fwt}}$ ($\uparrow$)} & RL2 ($\downarrow$) \\
        \midrule
        \rowcolor{rowgrayblue}JT (UB) & -- & None & 0.9075 & -- & {--} & 0.0041 \\
        \rowcolor{rowgrayblue}SFT (LB) & -- & None & 0.7420 & 0.1322 & 0.2135 & 0.0115  \\
        \rowcolor{rowgraygreen}SI \cite{zenke2017continual} & ICML'17 & None & 0.6863 & 0.2330 & 0.1938 & 0.0145  \\
        \rowcolor{rowgraygreen}EWC \cite{james2017ewc} & PNAS'17 & None & 0.5311 & 0.3177 & 0.1776 & 0.0201 \\
        \rowcolor{rowgraygreen}LwF \cite{li2017learning} & TPAMI'17 & None & 0.7648 & 0.0807 & 0.2894 & 0.0110 \\
        \rowcolor{rowgrayyellow}MER \cite{riemer2019learning} & ICLR'19 & Raw Data & 0.8276 & 0.1446 & 0.2806 & 0.0070 \\
        \rowcolor{rowgrayyellow}DER++ \cite{buzzega2020dark} & NeurIPS'20 & Raw Data & 0.8285 & 0.1523 & 0.2851 & 0.0068 \\
        \rowcolor{rowgrayyellow}TOPIC \cite{tao2020few} & CVPR'20 & Raw Data & 0.8006 & 0.1344 & 0.2744 & 0.0077 \\
        \rowcolor{rowgrayyellow}GEM \cite{kukleva2021generalized} & ICCV'21 & Raw Data & 0.8309 & 0.0721 & 0.2883 & 0.0068 \\
        \rowcolor{rowgrayyellow}STAR \cite{eskandar2025star} & ICLR'25 & Raw Data & 0.8417 & 0.0778 & 0.2170  & 0.0063 \\
        \rowcolor{rowgraypink}Feature MER & -- & Feature & 0.4914 & 0.2354 & 0.2344 & 0.0221 \\
        \rowcolor{rowgraypink}SLCA \cite{zhang2023slca} & ICCV'23 & Feature & 0.8130 & 0.0920 & 0.2453 & 0.0075\\
        \rowcolor{rowgraypink}NC-FSCIL \cite{yang2023neural} & ICLR'23 & Feature & 0.8087 & 0.0203 & 0.3404 & 0.0077\\
        \rowcolor{rowgraypink}FS-Aug \cite{li2024continual} & TCSVT'24 & Feature & 0.8123 & 0.1412 &  0.2928 & 0.0072 \\
        \rowcolor{rowgraypink}MAGR \cite{zhou2024magr} & ECCV'24 & Feature & 0.8580 & 0.0167 & 0.2952 & 0.0061\\
        \rowcolor{rowgraypink}ProNC \cite{wang2026rethinking} & ICLR'26 & Feature & 0.8487 & 0.0244 & 0.1007 & 0.0063 \\
        \rowcolor{rowgraypink}MAGR++ (Ours) & -- & Feature & \bf 0.8902 & \bf 0.0090 & \bf 0.3915  & \bf 0.0050 \\
        \bottomrule
        \end{tabular}
        }
    \end{minipage}\hfill
\begin{minipage}{0.50\linewidth}
        \centering
        \vspace{2mm}
        \subcaption{UNLV-Dive}
        \vspace{-1.5mm}
        \resizebox{\linewidth}{!}{
        \begin{tabular}{l r c cc S[table-format=+1.4, input-symbols={\textbf}] c}
        \toprule
        \textbf{Method} & \textbf{Publisher} & \textbf{Memory} & $\rho_{\text{avg}}$ ($\uparrow$) & $\rho_{\text{aft}}$ ($\downarrow$) & \multicolumn{1}{c}{$\rho_{\text{fwt}}$ ($\uparrow$)} & RL2 ($\downarrow$) \\
        \midrule
        \rowcolor{rowgrayblue}JT (UB) & -- & None & 0.8460 & -- & {--} & 0.0041 \\
        \rowcolor{rowgrayblue}SFT (LB) & -- & None & 0.6307 & 0.2135 & 0.3595 & 0.0163  \\
        \rowcolor{rowgraygreen}SI \cite{zenke2017continual} & ICML'17 & None & 0.1519 & 0.3822 & 0.0270 & 0.0395 \\
        \rowcolor{rowgraygreen}EWC \cite{james2017ewc} & PNAS'17 & None & 0.4096 & 0.2576 & 0.3039 & 0.0351 \\
        \rowcolor{rowgraygreen}LwF \cite{li2017learning} & TPAMI'17 & None & 0.6081 & 0.1578 & 0.3230 & 0.0208 \\
        \rowcolor{rowgrayyellow}MER \cite{riemer2019learning} & ICLR'19 & Raw Data & 0.7397 & 0.1321 & 0.0465 & 0.0110 \\
        \rowcolor{rowgrayyellow}DER++ \cite{buzzega2020dark} & NeurIPS'20 & Raw Data & 0.7206 & 0.1382 & -0.1773 & 0.0117 \\
        \rowcolor{rowgrayyellow}TOPIC \cite{tao2020few} & CVPR'20 & Raw Data & 0.4085 & 0.2647 & 0.1132 & 0.0358 \\
        \rowcolor{rowgrayyellow}GEM \cite{kukleva2021generalized} & ICCV'21 & Raw Data & 0.6538 & 0.2322 & 0.0270 & 0.0153 \\
        \rowcolor{rowgrayyellow}STAR \cite{eskandar2025star} & ICLR'25 & Raw Data &  0.7597 & 0.1238 & -0.0745 & 0.0249 \\
        \rowcolor{rowgraypink}Feature MER & -- & Feature & 0.5675 & 0.1322 & 0.1558  & 0.0314 \\
        \rowcolor{rowgraypink}SLCA \cite{zhang2023slca} & ICCV'23 & Feature & 0.5551 & 0.1085 & 0.3200 & 0.0317 \\
        \rowcolor{rowgraypink}NC-FSCIL \cite{yang2023neural} & ICLR'23 & Feature & 0.6458 & 0.0637 & -0.1677  & 0.0163 \\
        \rowcolor{rowgraypink}FS-Aug \cite{li2024continual} & TCSVT'24 & Feature & 0.7374 & \bf 0.0263 & -0.0742 & 0.0275 \\
        \rowcolor{rowgraypink}MAGR \cite{zhou2024magr} & ECCV'24 & Feature & 0.7668 & 0.0827 & 0.1227  & 0.0120 \\
        \rowcolor{rowgraypink}ProNC \cite{wang2026rethinking} & ICLR'26 & Feature & 0.7812 & 0.0697 & 0.1047 & 0.0125 \\
        \rowcolor{rowgraypink}MAGR++ (Ours) & -- & Feature & \bf 0.8165 & 0.0910  & \bf  0.3502  & \bf 0.0118 \\
        \bottomrule
        \end{tabular}
        }
    \end{minipage}
    \vspace{-0.5cm}
\end{table*}  
\myPara{Graph Regularization.}  
To align the feature geometry with quality relationships, we construct a score distance matrix $\mathbf{S} = \bm{y} - \bm{y}^\top$, where $\bm{y}$ denotes the quality labels of the joint batch.  
We then define the regularization loss as:
\begin{equation} \label{eq:iij}
\mathcal{L}_{\text{reg}} = 
\|\mathbf{A} - \mathbf{S}\|_2^2 
+ \sum_{i=1}^2\sum_{j=1}^2 \|\mathbf{A}_{ij} - \mathbf{S}_{ij}\|_2^2.
\end{equation} 
By jointly optimizing the global distance matrix $\mathbf{A}$ and its sub-blocks against $\mathbf{S}$, IIJ-GR enforces alignment between feature distances and score relationships at both intra- and inter-session levels.  
This design preserves local ranking consistency within each session while maintaining global comparability across sessions, ensuring that rectified features remain semantically meaningful for continual regression.

\begin{table*}[t]
\centering
\setlength{\tabcolsep}{3pt}
\MStabcaption{tab:online}{\textbf{Online} performance comparison. The primary metric is $\rho_{\text{avg}}$, with Sequential Fine-Tuning (SFT) as the Lower Bound (LB). Best results are in bold, excluding UB and LB. $\uparrow$/$\downarrow$ indicate higher/lower is better.}{}

\begin{minipage}{0.50\linewidth}
\centering
\subcaption{MTL-AQA (default, w/o difficulty label)}
\vspace{-1.5mm}
\resizebox{\linewidth}{!}{
\begin{tabular}{l r c cc S[table-format=+1.4, input-symbols={\textbf}] c }
\toprule
        \textbf{Method} & \textbf{Publisher} & \textbf{Memory} & $\rho_{\text{avg}}$ ($\uparrow$) & $\rho_{\text{aft}}$ ($\downarrow$) & \multicolumn{1}{c}{$\rho_{\text{fwt}}$ ($\uparrow$)} & RL2 ($\downarrow$) \\
\midrule
\rowcolor{rowgrayblue}SFT (LB) & -- & None & 0.4926 & 0.0649 & -0.1416 & 0.0558 \\
\rowcolor{rowgraygreen}SI \cite{zenke2017continual} & ICML'17 & None & 0.5243 & 0.0253 & 0.0669 & 0.0496 \\
\rowcolor{rowgraygreen}EWC \cite{james2017ewc} & PNAS'17 & None & 0.5401 & 0.0303 & 0.0850 & 0.0478 \\
\rowcolor{rowgraygreen}LwF \cite{li2017learning} & TPAMI'17 & None & 0.5243 & 0.0170 & 0.0797 & 0.0491 \\
\rowcolor{rowgrayyellow}MER \cite{riemer2019learning} & ICLR'19 & Raw Data & 0.5734 & 0.0394 & 0.0262 & 0.0438 \\
\rowcolor{rowgrayyellow}DER++ \cite{buzzega2020dark} & NeurIPS'20 & Raw Data & 0.5415 & 0.0146 & 0.0857 & 0.0469 \\
\rowcolor{rowgrayyellow}TOPIC \cite{tao2020few} & CVPR'20 & Raw Data & 0.5116 & 0.0981 & 0.0690 & 0.0546 \\
\rowcolor{rowgrayyellow}GEM \cite{kukleva2021generalized} & ICCV'21 & Raw Data & 0.5490 & 0.0612 & \bf 0.0972 & 0.0465 \\
\rowcolor{rowgrayyellow}STAR \cite{eskandar2025star} & ICLR'25 & Raw Data & 0.5331 & 0.0426 & 0.0376 & 0.0488  \\
\rowcolor{rowgraypink}Feature MER & -- & Feature & 0.3571 & 0.1444 & -0.0213 & 0.0638 \\
\rowcolor{rowgraypink}SLCA \cite{zhang2023slca} & ICCV'23 & Feature & 0.4880 & 0.0430 & -0.0282 & 0.0522 \\
\rowcolor{rowgraypink}NC-FSCIL \cite{yang2023neural} & ICLR'23 & Feature & 0.4971 & 0.0291 & -0.0463 & 0.0512 \\
\rowcolor{rowgraypink}FS-Aug \cite{li2024continual} & TCSVT'24 & Feature & 0.3322 & 0.0725 & -0.0581 & 0.0672 \\
\rowcolor{rowgraypink}MAGR \cite{zhou2024magr} & ECCV'24 & Feature & 0.5196 & 0.0337 & 0.0282 & 0.0512 \\
\rowcolor{rowgraypink}ProNC \cite{wang2026rethinking} & ICLR'26 & Feature & 0.5381 & 0.0167 & 0.0525 & 0.0501 \\
\rowcolor{rowgraypink}MAGR++ (Ours) & -- & Feature & \textbf{0.5618} & \textbf{0.0165} & 0.0405  & \bf 0.0413 \\
\bottomrule
\end{tabular}
}
\end{minipage}\hfill
\begin{minipage}{0.50\linewidth}
\centering
\subcaption{MTL-AQA (w/ difficulty label)}
\vspace{-1.5mm}
\resizebox{\linewidth}{!}{
\begin{tabular}{l r c cc S[table-format=+1.4, input-symbols={\textbf}] c}
\toprule
        \textbf{Method} & \textbf{Publisher} & \textbf{Memory} & $\rho_{\text{avg}}$ ($\uparrow$) & $\rho_{\text{aft}}$ ($\downarrow$) & \multicolumn{1}{c}{$\rho_{\text{fwt}}$ ($\uparrow$)} & RL2 ($\downarrow$) \\
\midrule
\rowcolor{rowgrayblue}SFT (LB) & -- & None & 0.6022 & 0.0647 & -0.0536 & 0.0368 \\
\rowcolor{rowgraygreen}SI \cite{zenke2017continual} & ICML'17 & None & 0.6581 & 0.0803 & 0.0429 & 0.0264 \\
\rowcolor{rowgraygreen}EWC \cite{james2017ewc} & PNAS'17 & None & 0.6371 & 0.1372 & 0.0895 & 0.0312 \\
\rowcolor{rowgraygreen}LwF \cite{li2017learning} & TPAMI'17 & None & 0.6416 & 0.0134 & 0.3076 & 0.0291 \\
\rowcolor{rowgrayyellow}MER \cite{riemer2019learning} & ICLR'19 & Raw Data & 0.6290 & 0.0833 & 0.0057 & 0.0328 \\
\rowcolor{rowgrayyellow}DER++ \cite{buzzega2020dark} & NeurIPS'20 & Raw Data & 0.6444 & 0.1133 & 0.0221 & 0.0290 \\
\rowcolor{rowgrayyellow}TOPIC \cite{tao2020few} & CVPR'20 & Raw Data & 0.6241 & 0.0916 & 0.0258 & 0.0341 \\
\rowcolor{rowgrayyellow}GEM \cite{kukleva2021generalized} & ICCV'21 & Raw Data & 0.6422 & 0.0894 & 0.0105 & 0.0296 \\
\rowcolor{rowgrayyellow}STAR \cite{eskandar2025star} & ICLR'25 & Raw Data & 0.6576 & 0.1082 & 0.0324 & 0.0258  \\
\rowcolor{rowgraypink}Feature MER & -- & Feature & 0.6065 & 0.0472 & -0.0294 & 0.0361 \\
\rowcolor{rowgraypink}SLCA \cite{zhang2023slca} & ICCV'23 & Feature & 0.5980 & 0.0827 & -0.0266 & 0.0372 \\
\rowcolor{rowgraypink}NC-FSCIL \cite{yang2023neural} & ICLR'23 & Feature & 0.5937 & 0.1006 & 0.0181 & 0.0384 \\
\rowcolor{rowgraypink}FS-Aug \cite{li2024continual} & TCSVT'24 & Feature & 0.5339 & 0.0723 & -0.0108 & 0.0468 \\
\rowcolor{rowgraypink}MAGR \cite{zhou2024magr} & ECCV'24 & Feature & 0.6416 & \textbf{0.0134} & \bf 0.3076 & \textbf{0.0236} \\
\rowcolor{rowgraypink}ProNC \cite{wang2026rethinking} & ICLR'26 & Feature & 0.6631 &  0.0997 & 0.0214 & 0.0299  \\
\rowcolor{rowgraypink}MAGR++ (Ours) & -- & Feature & \textbf{0.6676} & 0.0810 & -0.0126 & 0.0240 \\
\bottomrule
\end{tabular}
}
\end{minipage}\hfill

\begin{minipage}{0.50\linewidth}
\centering
\vspace{2mm}
\subcaption{FineDiving (w/o dive number)}
\vspace{-1.5mm}
\resizebox{\linewidth}{!}{
\begin{tabular}{l r c cc S[table-format=+1.4, input-symbols={\textbf}] c}
\toprule
        \textbf{Method} & \textbf{Publisher} & \textbf{Memory} & $\rho_{\text{avg}}$ ($\uparrow$) & $\rho_{\text{aft}}$ ($\downarrow$) & \multicolumn{1}{c}{$\rho_{\text{fwt}}$ ($\uparrow$)} & RL2 ($\downarrow$) \\
\midrule
\rowcolor{rowgrayblue}SFT (LB) & -- & None & 0.3970 & 0.0448 & 0.0557 & 0.0320 \\
\rowcolor{rowgraygreen}SI \cite{zenke2017continual} & ICML'17 & None & 0.4597 & 0.0149 & 0.0553 & 0.0742 \\
\rowcolor{rowgraygreen}EWC \cite{james2017ewc} & PNAS'17 & None & 0.3222 & 0.0471 & 0.0825 & 0.0416 \\
\rowcolor{rowgraygreen}LwF \cite{li2017learning} & TPAMI'17 & None & 0.4230 & 0.0217 & 0.1144 & 0.0298 \\
\rowcolor{rowgrayyellow}MER \cite{riemer2019learning} & ICLR'19 & Raw Data & 0.4116 & 0.0534 & 0.0426 & 0.0311 \\
\rowcolor{rowgrayyellow}DER++ \cite{buzzega2020dark} & NeurIPS'20 & Raw Data & 0.4358 & 0.0707 & 0.1045 & 0.0276 \\
\rowcolor{rowgrayyellow}TOPIC \cite{tao2020few} & CVPR'20 & Raw Data & 0.4654 & 0.0978 & 0.1086 & 0.0218 \\
\rowcolor{rowgrayyellow}GEM \cite{kukleva2021generalized} & ICCV'21 & Raw Data & 0.4414 & 0.0531 & 0.1109 & 0.0269 \\
\rowcolor{rowgrayyellow}STAR \cite{eskandar2025star} & ICLR'25 & Raw Data & 0.4760 & 0.0112 & 0.1011 & 0.0204 \\
\rowcolor{rowgraypink}Feature MER & -- & Feature & 0.1935 & 0.0998 & 0.1559 & 0.0388  \\
\rowcolor{rowgraypink}SLCA \cite{zhang2023slca} & ICCV'23 & Feature & 0.3935 & 0.3360 & 0.2346 & 0.0241 \\
\rowcolor{rowgraypink}NC-FSCIL \cite{yang2023neural} & ICLR'23 & Feature & 0.3810 & 0.0079 & \bf 0.2518 & 0.0231 \\
\rowcolor{rowgraypink}FS-Aug \cite{li2024continual} & TCSVT'24 & Feature & 0.4266 & 0.0732 & 0.1645 & 0.0287 \\
\rowcolor{rowgraypink}MAGR \cite{zhou2024magr} & ECCV'24 & Feature & 0.4641 & \textbf{0.0062} & 0.2020 & \textbf{0.0187} \\
\rowcolor{rowgraypink}ProNC \cite{wang2026rethinking} & ICLR'26 & Feature & 0.4608 & 0.0166 & 0.1197 & 0.0221 \\
\rowcolor{rowgraypink}MAGR++ (Ours) & -- & Feature & \textbf{0.5325} & 0.0094 & 0.1227 & 0.0215 \\
\bottomrule
\end{tabular}
}
\end{minipage}\hfill
\begin{minipage}{0.50\linewidth}
\centering
\vspace{2mm}
\subcaption{UNLV-Dive}
\vspace{-1.5mm}
\resizebox{\linewidth}{!}{
\begin{tabular}{l r c cc S[table-format=+1.4, input-symbols={\textbf}] c}
\toprule
        \textbf{Method} & \textbf{Publisher} & \textbf{Memory} & $\rho_{\text{avg}}$ ($\uparrow$) & $\rho_{\text{aft}}$ ($\downarrow$) & \multicolumn{1}{c}{$\rho_{\text{fwt}}$ ($\uparrow$)} & RL2 ($\downarrow$) \\
\midrule
\rowcolor{rowgrayblue}SFT (LB) & -- & None & 0.2251 & 0.2592 & -0.1432 & 0.1624 \\
\rowcolor{rowgraygreen}SI \cite{zenke2017continual} & ICML'17 & None & 0.3465 & 0.2211 & -0.3182 & 0.1087 \\
\rowcolor{rowgraygreen}EWC \cite{james2017ewc} & PNAS'17 & None & 0.3722 & 0.2631 & -0.3542 & 0.0896 \\
\rowcolor{rowgraygreen}LwF \cite{li2017learning} & TPAMI'17 & None & 0.3981 & 0.2132 & -0.3913 & 0.0814 \\
\rowcolor{rowgrayyellow}MER \cite{riemer2019learning} & ICLR'19 & Raw Data & 0.2890 & 0.2222 & -0.3567 & 0.1248 \\
\rowcolor{rowgrayyellow}DER++ \cite{buzzega2020dark} & NeurIPS'20 & Raw Data & 0.4291 & 0.1350 & -0.3106 & 0.0692 \\
\rowcolor{rowgrayyellow}TOPIC \cite{tao2020few} & CVPR'20 & Raw Data & 0.3874 & 0.2112 & -0.3454 & 0.0837 \\
\rowcolor{rowgrayyellow}GEM \cite{kukleva2021generalized} & ICCV'21 & Raw Data & 0.4094 & 0.2315 & -0.3773 & 0.0781 \\
\rowcolor{rowgrayyellow}STAR \cite{eskandar2025star} & ICLR'25 & Raw Data & 0.3757 &  0.2392 & -0.3815 & 0.0758 \\
\rowcolor{rowgraypink}Feature MER & -- & Feature & 0.1308 & 0.2126 & -0.4571 & 0.2836 \\
\rowcolor{rowgraypink}SLCA \cite{zhang2023slca} & ICCV'23 & Feature & 0.3119 & 0.1641 & -0.3082 & 0.0956 \\
\rowcolor{rowgraypink}NC-FSCIL \cite{yang2023neural} & ICLR'23 & Feature & 0.3136 & 0.1282 & -0.4892 & 0.0359 \\
\rowcolor{rowgraypink}FS-Aug \cite{li2024continual} & TCSVT'24 & Feature & 0.3639 & 0.1510 & -0.1555 & 0.0871 \\
\rowcolor{rowgraypink}MAGR \cite{zhou2024magr} & ECCV'24 & Feature & 0.4202 & 0.1947 & -0.0499 & 0.0327 \\
\rowcolor{rowgraypink}ProNC \cite{wang2026rethinking} & ICLR'26 & Feature & 0.3922 & 0.2117 &  -0.3769 & 0.0553 \\
\rowcolor{rowgraypink}MAGR++ (Ours) & -- & Feature & \textbf{0.5117} & \textbf{0.0959} & \bf 0.3927 & \textbf{0.0214} \\
\bottomrule
\end{tabular}
}
\end{minipage}
\vspace{-0.5cm}
\end{table*}  \section{Experiments}  \label{sec:exp}
We conduct extensive experiments to evaluate the effectiveness and robustness of our proposed MAGR++ for CAQA. 
In addition, we provide supplementary experiments in \cref{sec:add_exp} to further validate the generality of our approach.

\subsection{Experimental Setting}
\myPara{Datasets.} 
We build CAQA benchmarks on three diving AQA datasets of varying scales and feature-shift levels (validated in \cref{tab:deviation}). 
\textbf{MTL-AQA} \cite{parmar2019action} contains 1412 samples across 16 diving events (male/female, single/double, 3m springboard, 10m platform), with detailed annotations of action categories, commentary, and AQA scores; 1059 samples are used for training and 353 for testing. \textbf{FineDiving} \cite{xu2022finediving} includes 3000 dives from major competitions (Olympics, World Cup, etc.), annotated with 52 action types, 29 sub-action types, 23 difficulty levels, temporal boundaries, and both action and AQA scores; we adopt the official 75\%/25\% train/test split. \textbf{UNLV-Dive} \cite{parmar2017learning} provides 370 videos (300 train and 70 test) from the 2012 London Olympics 10 m platform event, with final scores ranging from 21.6–102.6 and execution scores in [0, 30]. We focus on diving datasets in the main paper because they offer multiple well-annotated benchmarks with consistent action types, ensuring fair cross-dataset comparison. Results on other actions, such as UNLV-Vault \cite{yu2021group}, are provided in \cref{sec:add_exp} to demonstrate generalization beyond diving while keeping the main analysis concise and focused.

To evaluate scalability under non-stationary settings, we further construct a cross-dataset continual AQA benchmark, CL-Dive, by unifying five diving-based datasets (including UNLV-Dive, MTL-AQA, FineDiving, and synchronized diving subsets from AQA-7 \cite{parmar2019action}). 
CL-Dive contains 3,733 training samples and 1,228 testing samples, and introduces substantial cross-dataset distribution shifts via a sequential task protocol. 
\myPara{CAQA Protocol.}  
To simulate real-world skill variations, we introduce a grade-incremental setting for CAQA that couples regression and classification challenges. The continuous quality space is discretized into $G$ grade intervals, with $S$ samples per session to induce challenging score variations. Unlike the uniform and independent class space in traditional class-incremental tasks \cite{zhang2023few}, our setting preserves contextual dependencies between adjacent grades. Here, grade order typically follows skill progression (with \cref{fig:parameter_change} analyzing task order effects), while the quality range also shifts across sessions. These challenges are compounded by the fine-grained regression nature of AQA, posing significant difficulties for lifelong learning in mitigating catastrophic forgetting.

\myPara{CAQA Metrics.} 
We adopt Spearman's Rank Correlation Coefficient (SRCC) $\rho$, 
the standard evaluation metric in AQA, to measure the rank consistency 
between predicted scores $\hat{\bm{y}}$ and ground-truth scores $\bm{y}$. 
Building on SRCC, we further define continual evaluation metrics to summarize 
overall performance, forgetting, and forward transfer across sessions. Given rank vectors $\bm{p}$ and $\bm{q}$ for $\bm{y}$ and $\hat{\bm{y}}$, SRCC is 
\begin{equation} \label{eq:rho}
  \rho = \frac{ \sum_i (p_i - \bar{p}) (q_i - \bar{q}) }{\sqrt{\sum_i (p_i - \bar{p})^2 \sum_i (q_i - \bar{q})^2}},
\end{equation}
where $\bar{p}$ and $\bar{q}$ are the mean ranks of $\bm{p}$ and $\bm{q}$.  
For continual evaluation, directly averaging session-wise SRCC values can
be unreliable because the number of evaluated samples may vary across
sessions. We therefore report the overall SRCC $\rho_{\text{avg}}$ as the \textbf{primary} CAQA metric, by pooling
predictions and ground-truth scores across the evaluated sessions before
computing SRCC. Thus, $\rho_{\rm avg}$ remains the standard SRCC criterion,
with only its aggregation adapted to the continual setting. To probe model stability and adaptability, we also report two auxiliary metrics: average forgetting $\rho_{\text{aft}} = \tfrac{1}{T-1} \sum_{t=1}^{T-1} \max_{i,j \in \{1,2,\cdots,T\}} (\rho_{i,t} - \rho_{j,t})$ and forward transfer $\rho_{\text{fwt}} = \tfrac{1}{T-1} \sum_{t=2}^{T} (\rho_{t-1,t} - \tilde{\rho}_{t})$, where $\rho_{i,j}$ denotes the correlation on the $j$-th test set after training up to task $i$ ($j \leq i$), and $\tilde{\rho}_{t}$ is the baseline correlation of a randomly initialized model on task $t$.

In addition to rank-based evaluation, we also report the Relative $\ell_2$ Distance (RL2) as a complementary metric to measure absolute prediction error. RL2 is defined as:
\begin{equation} \label{eq:rl2}
\text{RL2} = \frac{1}{N} \sum_{i=1}^{N} \left(\frac{|\hat{y}_i - y_i|}{y_{\max} - y_{\min}}\right)^2,
\end{equation}
where $y_{\max}$ and $y_{\min}$ denote the maximum and minimum ground-truth scores in the dataset. RL2 provides a scale-normalized measure of prediction accuracy, complementing SRCC by capturing absolute regression fidelity. 
\myPara{Implementation Details.}  
All experiments are conducted on two NVIDIA RTX 4090 GPUs using PyTorch. We adopt the I3D backbone \cite{carreira2017quo} with a score regression model \cite{zhou2023hierarchical}. Training is performed using Adam with a learning rate of $10^{-4}$ and weight decay of $10^{-5}$, run for up to 50 epochs in the offline setting, and for 1 epoch in the online setting. To emulate real-world constraints of limited data streams and label scarcity, we configure the incremental setting with $G=5$ and $S=20$, using batch size $b_1=5$ and mini-batch size $b_2=3$. The remaining data are reserved for base-session adaptation, providing a stable foundation before incremental updates. Batch normalization layers in the backbone are frozen to mitigate small-batch effects. MP is implemented as a two-layer MLP that learns residual feature shifts between old and new manifolds, which can be effectively captured without complex architectures. 
All loss terms are normalized to eliminate scale mismatch. Consequently, the loss weights
$\lambda_{\text{tune}}$, $\lambda_{\text{proj}}$, and $\lambda_{\text{reg}}$ are fixed to 1 throughout all experiments. This choice provides a robust default and ensures a fair comparison with MAGR \cite{zhou2024magr}.
Unless otherwise specified, all baselines are evaluated under the same backbone, memory budget, and training protocol for fair comparison. Hyperparameters are selected based on official implementations when available and further adapted to the CAQA setting.  

\subsection{Comparisons with Strong Baselines}   \label{sec:sota_cmp}
Following prior work~\cite{yu2021group,zhou2023hierarchical}, we evaluate all methods on the MTL-AQA dataset with and without Difficulty Degree (DD). In total, our evaluation covers three benchmark datasets and four experimental configurations.
For comprehensive comparison, we include representative CL baselines spanning rehearsal-free methods (SI~\cite{zenke2017continual}, EWC~\cite{james2017ewc}, and LwF~\cite{li2017learning}), raw-data replay methods (MER~\cite{riemer2019learning}, DER++~\cite{buzzega2020dark}, GEM~\cite{kukleva2021generalized}, TOPIC~\cite{tao2020few}, and STAR~\cite{eskandar2025star}), and feature-replay-based methods (SLCA~\cite{zhang2023slca}, NC-FSCIL~\cite{yang2023neural}, and ProNC~\cite{wang2026rethinking}). We further incorporate recent CAQA-specific baselines, including FS-Aug~\cite{li2024continual} and MAGR~\cite{zhou2024magr}. Among these methods, EWC and SLCA can also be viewed as representative parameter-constrained stabilization strategies. In particular, NC-FSCIL and ProNC are originally designed for few-shot CL and are included due to the small-scale and sparse supervision characteristics of AQA datasets. We then compare MAGR++ against these strong baselines in terms of offline performance (see \cref{tab:offline,tab:cl_dive}), online performance (see \cref{tab:online}), and computational efficiency (see \cref{tab:computation}).  

\myPara{Offline Performance.}
In \cref{tab:offline}, we report Upper-Bound (UB) results by jointly training all samples with the baseline \cite{zhou2023hierarchical}. These results serve only as references for fair comparison, as our focus is on enhancing CAQA performance rather than pursuing upper-bound improvements, which have been explored in recent AQA studies \cite{xu2025human,xu2025comprehensive}.  In contrast, sequentially training each task serves as the Lower Bound (LB). The gap between UB and LB reflects catastrophic forgetting, as evidenced by a 41.69\% correlation drop on MTL-AQA (w/o DD).
Among these CL baselines, rehearsal-free methods generally perform worse than replay-based methods, indicating that the complexity of video and the fine-grained nature of AQA make replay-based strategies simple yet effective. Furthermore, raw data replay methods slightly outperform recent feature replay methods, as the continually adapting feature space induces severe catastrophic forgetting.

In contrast, MAGR++ explicitly addresses manifold shift and consistently surpasses MAGR and other feature-replay baselines across datasets with varying shift severity. On MTL-AQA (w/o DD), MAGR++ improves $\rho_{\text{avg}}$ from 0.8979 to 0.9205, a 2.52\% gain over MAGR and 9.25\% over NC-FSCIL \cite{yang2023neural}. With DD, it achieves 1.58\% and 2.70\% improvements over MAGR and FS-Aug \cite{li2024continual}, respectively. On FineDiving, MAGR++ reaches 0.8902, outperforming MAGR by 3.75\% and SLCA \cite{zhang2023slca} by 9.50\%, while on UNLV-Dive it improves from 0.7668 to 0.8165, a 6.48\% increase over MAGR and 10.73\% over FS-Aug. Beyond correlation, MAGR++ also achieves lower forgetting and stronger forward transfer than all competing baselines, highlighting its superiority.

In addition, MAGR++ outperforms raw data replay methods such as MER \cite{riemer2019learning} and DER++ \cite{buzzega2020dark} by explicitly modeling feature relations across sessions, thereby mitigating catastrophic forgetting more effectively. Furthermore, by balancing plasticity and stability via effective FPFT, MAGR++ surpasses relation-based methods such as TOPIC \cite{tao2020few}.

\begin{table}[h]
    \centering
    \setlength{\tabcolsep}{5pt}
    \caption{Feature deviations and correlation gains.}
    \begin{tabular}{llS[table-format=2.2]S[table-format=2.2]S[table-format=2.2]}
    \toprule
    \multicolumn{2}{c}{\textbf{Setting}}  & {\textbf{FineDiving}} & {\textbf{MTL-AQA}} & {\textbf{UNLV-Dive}} \\ 
    \midrule
    \rowcolor{rowgrayyellow} \multicolumn{2}{l}{Deviation Strength (MSE)}   & 26.85 & 35.28 & 51.75 \\ 
    \rowcolor{rowgraypink}\cellcolor{rowgraypink}  & FS-Aug \cite{li2024continual} & -0.09 & -4.34 & +14.18 \\
    \rowcolor{rowgraypink}\cellcolor{rowgraypink}  & MAGR \cite{zhou2024magr}     & +5.66 & +6.56 & +15.64 \\ 
    \rowcolor{rowgraypink}\cellcolor{rowgraypink} \multirow{-3}{*}{$\Delta \rho_{\text{avg}}$} & MAGR++ & +9.50 & +9.25 & +26.43 \\
    \bottomrule
    \end{tabular}
    \label{tab:deviation}
\end{table}

Finally, we perform a \textit{cross-dataset analysis} by quantifying the feature deviations between pretrained and fine-tuned features under joint training, as shown in \cref{tab:deviation}.  
This deviation, measured by mean squared error (MSE), reflects the degree of manifold shift induced during continual adaptation.  
We observe that the three datasets exhibit different deviation levels: datasets with fewer samples and classes (e.g., UNLV-Dive) show larger deviations, while larger datasets (e.g., FineDiving) exhibit smaller deviations.  
When comparing replay-based methods, FS-Aug often degrades under stronger shifts (e.g., $-4.34$ on MTL-AQA), while MAGR achieves moderate gains ($+5.66$, $+6.56$, $+15.64$).  
In contrast, MAGR++ consistently delivers the largest improvements, scaling with deviation strength: $+9.50$ on FineDiving (MSE$=$26.85), $+9.25$ on MTL-AQA (MSE$=$35.28), and $+26.43$ on UNLV-Dive (MSE$=$51.75).  
This confirms the superiority of MAGR++ to mitigate feature shifts over both prior methods. 
\begin{table}[!h]
    \centering
    \setlength{\tabcolsep}{3pt}
    \MStabcaption{tab:cl_dive}{Plug-and-play performance comparison on CD-AQA. The primary metric is $\rho_{\text{avg}}$, with Joint Training (JT) as the Upper Bound (UB) and Sequential Fine-Tuning (SFT) as the Lower Bound (LB). Best results are in bold, excluding UB and LB. $\uparrow$/$\downarrow$ indicate higher/lower is better.}{}
\begin{minipage}{\linewidth}
        \centering
\resizebox{\linewidth}{!}{
        \begin{tabular}{l r c cc S[table-format=+1.4, input-symbols={\textbf}] c}
        \toprule
        \textbf{Method} & \textbf{Publisher} & \textbf{Memory} & $\rho_{\text{avg}}$ ($\uparrow$) & $\rho_{\text{aft}}$ ($\downarrow$) & \multicolumn{1}{c}{$\rho_{\text{fwt}}$ ($\uparrow$)} & RL2 ($\downarrow$) \\
        \midrule
        \multicolumn{3}{l}{\textit{w/ MUSDL \cite{tang2020uncertainty}}}
        \\
        \midrule
        \rowcolor{rowgrayblue}JT (UB) & -- & None & 0.9211 & {--} & {--} & 0.0035 \\
        \rowcolor{rowgrayblue}SFT (LB) & -- & None & 0.7771 & 0.0821 &  0.7337 & 0.0104 \\
        \rowcolor{rowgraygreen}LwF \cite{li2017learning} & TPAMI'17 & None & 0.7697 & 0.0675 & 0.7200 & 0.0152 \\
        \rowcolor{rowgrayyellow}DER++ \cite{buzzega2020dark} & NeurIPS'20 & Raw Data & 0.7846 & 0.0985 & 0.6618 & 0.0109 \\
        \rowcolor{rowgrayyellow}STAR \cite{eskandar2025star} & ICLR'25 & Raw Data & 0.7700 & 0.0529 & 0.6619 & 0.0108 \\
        \rowcolor{brown!8}Feature MER & -- & Feature & 0.7039 & 0.1204 & 0.6272 & 0.0142 \\
        \rowcolor{brown!8}FS-Aug \cite{li2024continual} & TCSVT'24 & Feature & 0.6264 & 0.2029 & 0.6869 & 0.0151 \\
        \rowcolor{brown!8}MAGR \cite{zhou2024magr} & ECCV'24 & Feature & 0.8137 & 0.0588 & \bf 0.7266 & 0.0113 \\
        \rowcolor{brown!8}ProNC \cite{wang2026rethinking} & ICLR'26 & Feature & 0.8318 & 0.0626 & 0.7051 & 0.0080 \\
        \rowcolor{brown!8}MAGR++ (Ours) & -- & Feature & \bf 0.8724 & \bf 0.0163 & 0.7155 & \bf 0.0059 \\
        
        \midrule
        \multicolumn{3}{l}{\textit{w/ HGCN \cite{zhou2023hierarchical}}}
        \\
        \midrule
        \rowcolor{rowgrayblue}JT (UB) & -- & None & 0.9246 & {--} & {--} & 0.0041 \\
        \rowcolor{rowgrayblue}SFT (LB) & -- & None & 0.7472 & 0.1322 & 0.8284 & 0.0121 \\
        \rowcolor{rowgraygreen}LwF \cite{li2017learning} & TPAMI'17 & None & 0.7518 & 0.0911 & 0.8524 & 0.0112 \\
        \rowcolor{rowgrayyellow}DER++ \cite{buzzega2020dark} & NeurIPS'20 & Raw Data & 0.7612 & 0.0782 & 0.8024 & 0.0127 \\
        \rowcolor{rowgrayyellow}STAR \cite{eskandar2025star} & ICLR'25 & Raw Data & 0.7578 & 0.1180 & 0.6984 & 0.0137 \\
        \rowcolor{brown!8}Feature MER & -- & Feature & 0.6939 & 0.1435 & 0.8120 & 0.0138 \\
        \rowcolor{brown!8}FS-Aug \cite{li2024continual} & TCSVT'24 & Feature & 0.6453 & 0.1742 & 0.8121 & 0.0143 \\
        \rowcolor{brown!8}MAGR \cite{zhou2024magr} & ECCV'24 & Feature & 0.7582 & 0.1188 & \bf 0.8701 & 0.0109 \\
        \rowcolor{brown!8}ProNC \cite{wang2026rethinking} & ICLR'26 & Feature & 0.8318 & 0.0626 & 0.7050 & 0.0080 \\
        \rowcolor{brown!8}MAGR++ (Ours) & -- & Feature & \bf 0.8670 & \bf 0.0116 &  0.8642 & \bf 0.0063 \\

        \midrule
        \multicolumn{3}{l}{\textit{w/ DAE \cite{zhang2024auto}}}
        \\
        \midrule
        \rowcolor{rowgrayblue}JT (UB) & -- & None & 0.9339 & {--} & {--} & 0.0033 \\
        \rowcolor{rowgrayblue}SFT (LB) & -- & None & 0.7387 & 0.1129 & 0.7771 & 0.0116 \\
        \rowcolor{rowgraygreen}LwF \cite{li2017learning} & TPAMI'17 & None & 0.7663 & 0.0811 & \bf 0.7879 & 0.0108 \\
        \rowcolor{rowgrayyellow}DER++ \cite{buzzega2020dark} & NeurIPS'20 & Raw Data & 0.7825 & 0.0527 & 0.7331 & 0.0101 \\
        \rowcolor{rowgrayyellow}STAR \cite{eskandar2025star} & ICLR'25 & Raw Data & 0.7351 & 0.0921 & 0.7366 & 0.0111 \\
        \rowcolor{brown!8}Feature MER & -- & Feature & 0.6737 & 0.1598 & 0.7448 & 0.0131 \\
        \rowcolor{brown!8}FS-Aug \cite{li2024continual} & TCSVT'24 & Feature & 0.6158 & 0.2089 & 0.7526 & 0.0168 \\
        \rowcolor{brown!8}MAGR \cite{zhou2024magr} & ECCV'24 & Feature & 0.7763 & 0.0876 & 0.7318 & 0.0100 \\
        \rowcolor{brown!8}ProNC \cite{wang2026rethinking} & ICLR'26 & Feature & 0.7901 & 0.0729 & 0.7495 & 0.0090 \\
        \rowcolor{brown!8}MAGR++ (Ours) & -- & Feature & \bf 0.8788 & \bf 0.0060 & 0.7181 & \bf 0.0061 \\

        \midrule
        \multicolumn{3}{l}{\textit{w/ PHI \cite{zhou2025phi}}}
        \\
        \midrule
        \rowcolor{rowgrayblue}JT (UB) & -- & None & 0.9322 & {--} & {--} & 0.0033 \\
        
        \rowcolor{rowgrayblue}SFT (LB) & -- & None & 0.7628 & \bf 0.0345 & 0.7189 & 0.0166 \\
        \rowcolor{rowgraygreen}LwF \cite{li2017learning} & TPAMI'17 & None & 0.7023 & 0.1050 & 0.7456 & 0.0210 \\
        \rowcolor{rowgrayyellow}DER++ \cite{buzzega2020dark} & NeurIPS'20 & Raw Data & 0.7583 & 0.1124 & 0.5539 & 0.0114 \\
        \rowcolor{rowgrayyellow}STAR \cite{eskandar2025star} & ICLR'25 & Raw Data & 0.7108 & 0.1388 & 0.5562 & 0.0128 \\
        \rowcolor{brown!8}Feature MER & -- & Feature & 0.7444 & 0.0847 & 0.5421 & 0.0213 \\
        \rowcolor{brown!8}FS-Aug \cite{li2024continual} & TCSVT'24 & Feature & 0.2232 & 0.6255 & 0.6245 & 0.0318 \\
        \rowcolor{brown!8}MAGR \cite{zhou2024magr} & ECCV'24 & Feature & 0.8110 & 0.0755 & 0.7785 & 0.0127 \\
        \rowcolor{brown!8}ProNC \cite{wang2026rethinking} & ICLR'26 & Feature & 0.8199 & 0.0630 & \bf 0.7867 & 0.0090 \\
        \rowcolor{brown!8}MAGR++ (Ours) & -- & Feature & \bf 0.8539 & 0.0683 &  0.7131 & 0.0085 \\
   
        \bottomrule
        \end{tabular}
        }
    \end{minipage}
\end{table}  
\myPara{Plug-and-Play Performance Comparison.}
In \cref{tab:cl_dive}, we further evaluate MAGR++ with four representative AQA heads on CD-AQA, including uncertainty-aware methods (MUSDL \cite{tang2020uncertainty} and DAE \cite{zhang2024auto}), a graph-based method (HGCN \cite{zhou2023hierarchical}), and a domain-shift-aware method (PHI \cite{zhou2025phi}). MAGR++ consistently achieves the best $\rho_{\text{avg}}$ across all heads, with relative improvements of 4.88\%, 4.23\%, 11.23\%, and 4.15\% over the second-best baseline on MUSDL, HGCN, DAE, and PHI, respectively. It also reduces RL2 by 26.3\%, 21.3\%, 32.2\%, and 5.6\%, showing that the improvements hold for both ranking correlation and score regression accuracy. Moreover, MAGR++ achieves the lowest forgetting on three of the four heads, with $\rho_{\text{aft}}$ reductions of 69.2\%, 81.5\%, and 88.6\% on MUSDL, HGCN, and DAE. Although PHI yields slightly lower forgetting with ProNC/SFT, MAGR++ still obtains the highest ranking correlation and the lowest scoring error, indicating a stronger overall accuracy-stability trade-off. These results demonstrate that MAGR++ can serve as a plug-and-play continual learning framework for diverse AQA heads rather than being coupled to a specific model design. 
\myPara{Online Performance.}  
In \cref{tab:online}, online results exhibit a different trend compared to offline.  
Rehearsal-free methods such as SI \cite{zenke2017continual} and LwF \cite{li2017learning} achieve competitive performance relative to replay-based methods, mainly due to the domain gap between the pretraining domain (coarse action recognition) and the fine-grained AQA domain \cite{zhou2025phi,zhou2024cofinal}.  
The single-epoch nature of online training, coupled with severe manifold shift, makes feature replay generally less effective than raw data replay.  
In contrast, MAGR++ explicitly manages feature shifts and further stabilizes adaptation through layer-adaptive fine-tuning, leading to consistent improvements over MAGR and other baselines.  
Specifically, on MTL-AQA (w/o DD), MAGR++ improves $\rho_{\text{avg}}$ from 0.5196 to 0.5618 (+8.12\% over MAGR, +6.57\% over MER).  
On MTL-AQA (w/ DD), MAGR++ attains a correlation of 0.6676, 4.05\% higher than MAGR and 3.61\% higher than DER++. 
On FineDiving, the correlation reaches 0.5325, 14.74\% higher than MAGR and 14.43\% higher than TOPIC. 
The largest gain is observed on UNLV-Dive, where MAGR++ improves correlation from 0.4202 to 0.5117, 21.77\% higher than MAGR and 19.27\% higher than DER++.
These results highlight the effectiveness of MAGR++ in mitigating catastrophic forgetting and adapting under severe distribution shifts in the online scenario.

\begin{table}[th]
    \centering
    \caption{Computational performance on MTL-AQA. All metrics are reported as improvements over the offline LB.}
    \label{tab:computation}
\resizebox{\linewidth}{!}{
    \begin{tabular}{lccc
    S[table-format=+1.4, input-symbols={\textbf}]
    S[table-format=+1.4, input-symbols={\textbf}]
    S[table-format=+1.4, input-symbols={\textbf}]
    }
    \toprule
    \bf Method & \makecell{\bf Params. \\ \bf (M)} & \makecell{\bf Training \\ \bf Time (h)} & \makecell{\bf Inference \\ \bf FPS} & {\makecell{$\Delta\rho_{\text{avg}}$}} & {\makecell{$\Delta\rho_{\text{aft}}$}} & {\makecell{$\Delta\rho_{\text{fwt}}$}} \\
    \midrule
    \rowcolor{rowgrayyellow} Feature MER & 12.62 & 2.22 & 85.83 & +0.1825 & +0.0731 & -0.0003 \\
    \rowcolor{rowgrayyellow} SLCA \cite{zhang2023slca} & 13.62 & 2.27 & 85.83 & +0.1765 & -0.0672 & +0.1127 \\
    \rowcolor{rowgrayyellow} NC-FSCIL \cite{yang2023neural} & 12.62 & 2.33 & 85.83 & +0.2968 & -0.0378 & +0.0180 \\
    \rowcolor{rowgrayyellow} MAGR \cite{zhou2024magr} & 12.63 & 2.23 & 85.83 & +0.3521 & -0.1301 & +0.1376 \\
    \rowcolor{rowgrayyellow} ProNC \cite{wang2026rethinking} & 12.62 & 2.35 & 85.83 & +0.3518 & -0.0198 & +0.1246 \\
    \rowcolor{rowgraypink} MAGR++ (Ours) & 12.63 & 2.32 & 85.83 & +0.3925 & -0.1421 & +0.0736 \\
    \bottomrule
    \end{tabular}
    }
\end{table}

\myPara{Computational Efficiency.}  
\cref{tab:computation} reports model size, training time, and inference speed.
MAGR++ introduces only negligible overhead, adding 0.01M parameters and 0.09h training time compared to MAGR.
At inference time, all methods share the same pipeline, including the backbone (I3D), input resolution, frame length, and regression head.
CL components, such as feature replay, graph regularization, and the manifold projector, are only used during training and are disabled at inference.
As a result, MAGR++ achieves the same inference speed (85.83 FPS) as all baselines, demonstrating that the proposed improvements incur no additional runtime cost.
 
\subsection{Ablation Study}
\begin{figure}
    \centering
    \includegraphics[width=\linewidth]{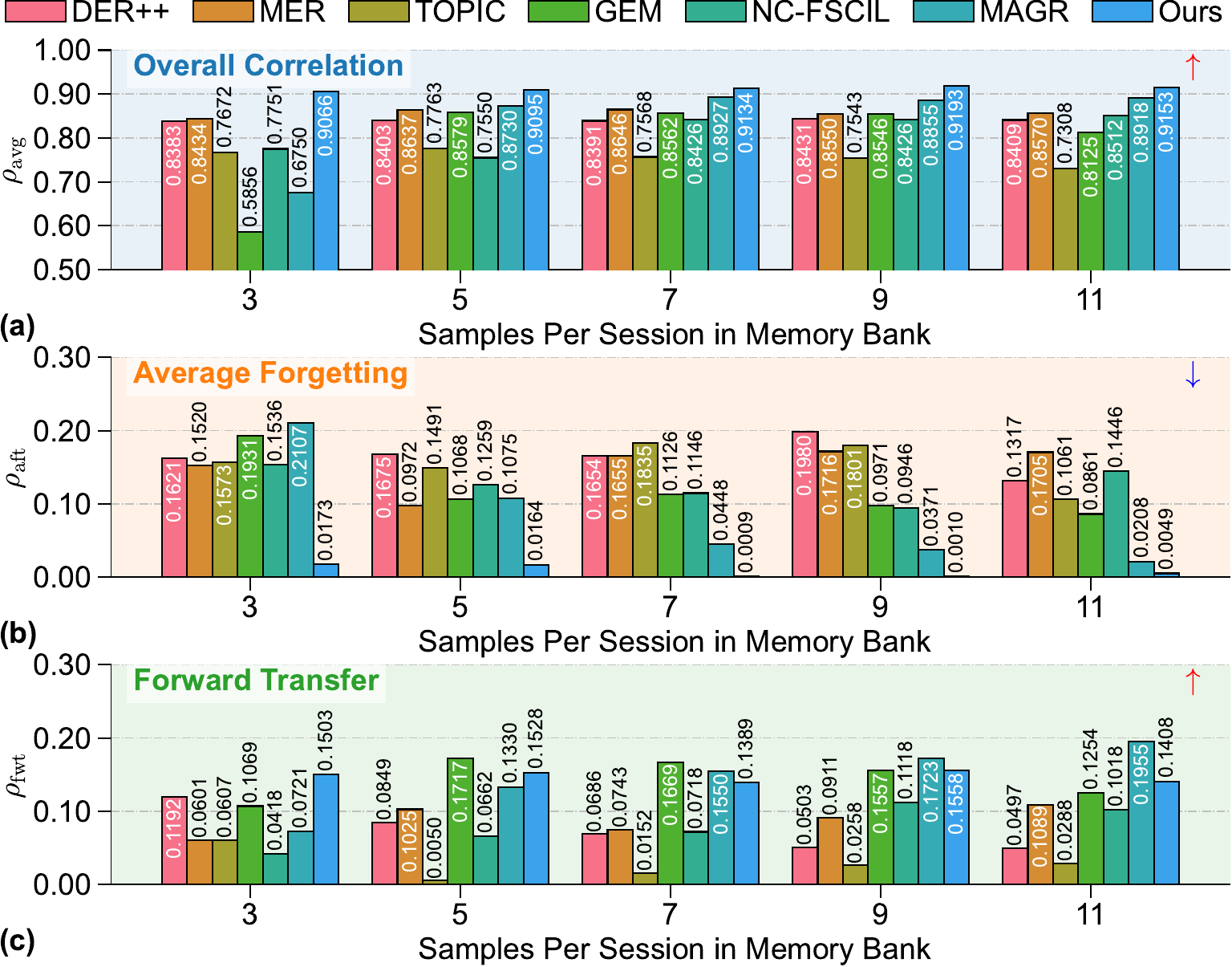}
    \caption{Comparison of different memory sizes on MTL-AQA.}
    \label{fig:memory_size}
    \phantomsubcaption\label{fig:memory_size-a}
    \phantomsubcaption\label{fig:memory_size-b}
    \phantomsubcaption\label{fig:memory_size-c}
    \vspace{-0.25cm}
\end{figure}

\begin{figure}
    \centering
    \includegraphics[width=\linewidth]{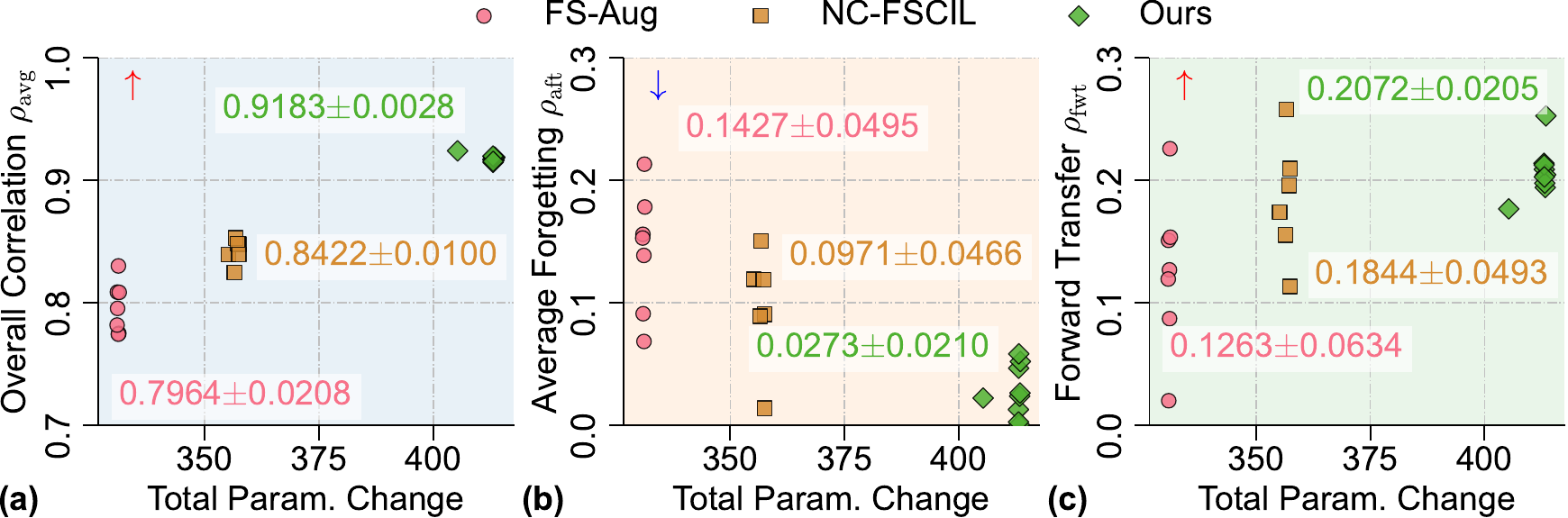}
    \caption{Performance vs. parameter changes across task orders.}
    \label{fig:parameter_change}
    \phantomsubcaption\label{fig:parameter_change-a}
    \phantomsubcaption\label{fig:parameter_change-b}
    \phantomsubcaption\label{fig:parameter_change-c}
\end{figure}

\begin{figure}
    \centering
    \includegraphics[width=\linewidth]{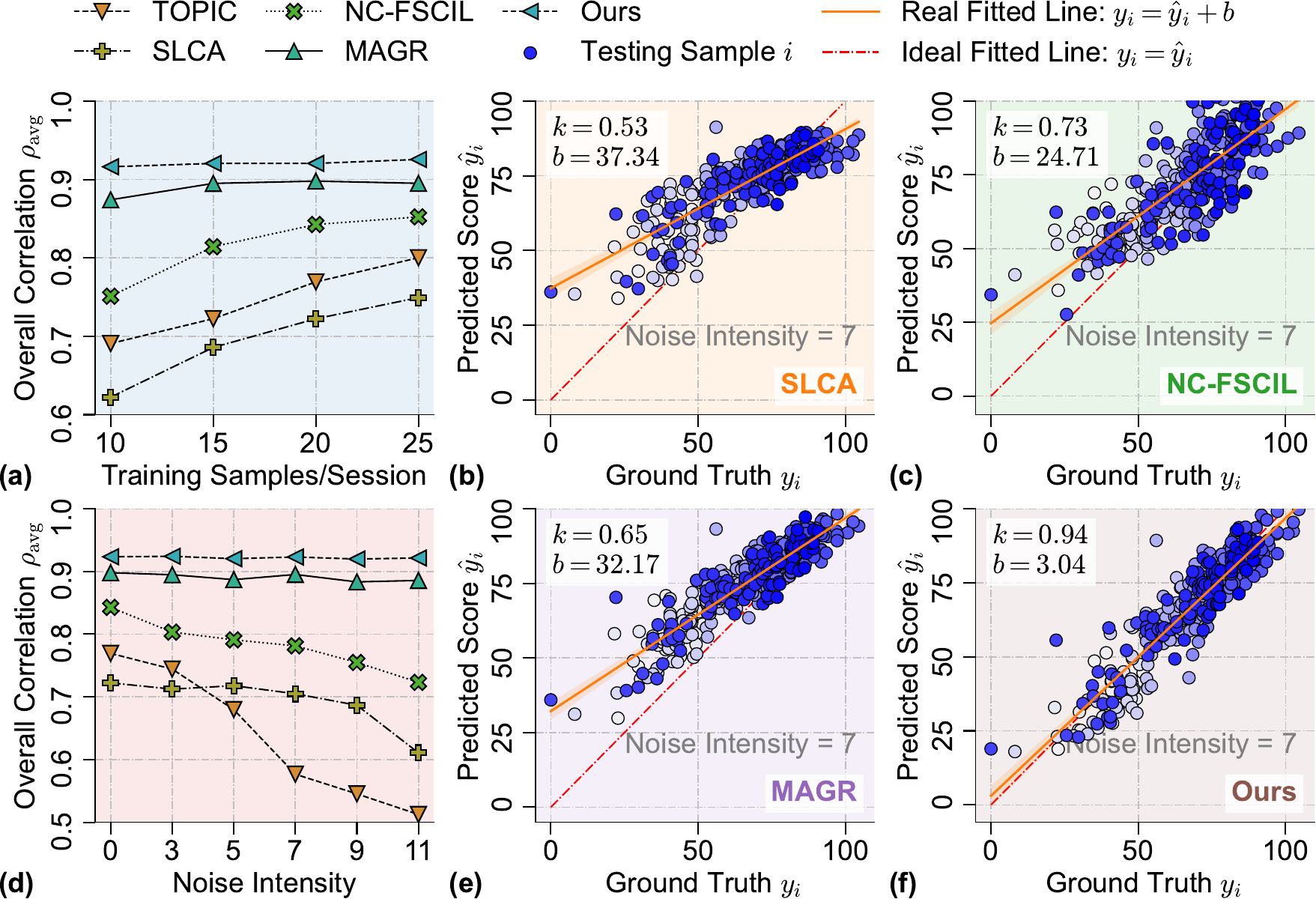}
    \caption{
        Performance comparison under label scarcity and labeling noise. 
        \subref{fig:sol-a} varies the number of training samples per session, and \subref{fig:sol-d} evaluates robustness under noisy annotations. 
        \subref{fig:sol-b}, \subref{fig:sol-c}, \subref{fig:sol-e}, and \subref{fig:sol-f} show correlation plots at noise level~7, with fitted regression lines. 
    }
    \label{fig:sol}
    \phantomsubcaption\label{fig:sol-a}
    \phantomsubcaption\label{fig:sol-b}
    \phantomsubcaption\label{fig:sol-c}
    \phantomsubcaption\label{fig:sol-d}
    \phantomsubcaption\label{fig:sol-e}
    \phantomsubcaption\label{fig:sol-f}
    \vspace{-0.2cm}
\end{figure}

Unless otherwise specified, all ablation experiments are conducted on MTL-AQA.
\cref{tab:ablation,tab:ablation_mtl} report the main results.  
Beyond these, we further investigate the impact of memory size (see \cref{fig:memory_size}), task order (see \cref{fig:parameter_change}), and robustness against sparse and noisy labeling attacks (see \cref{fig:sol}).  

\myPara{Impact of Core Modules.}
\cref{tab:ablation} reports the ablation results of the core components in MAGR++.
Compared with the full model (ID~1), removing MP (ID~3) reduces $\rho_{\text{avg}}$ from 0.9205 to 0.8418 and increases $\rho_{\text{aft}}$ from 0.0103 to 0.0995, indicating severe forgetting without feature rectification.
Similarly, removing IIJ-GR (ID~4) decreases $\rho_{\text{avg}}$ to 0.8548.
When all three modules, namely layer-adaptive FPFT (LA-FPFT), MP, and IIJ-GR, are removed (ID~8), $\rho_{\text{avg}}$ further drops to 0.7283.
Moreover, the combined removals in ID~5, ID~6, and ID~7 consistently lead to larger performance degradation than removing any single module, indicating clear interaction effects among LA-FPFT, MP, and IIJ-GR.
These results indicate that stable CL performance arises from the coordinated interaction among LA-FPFT, MP, and IIJ-GR.

\begin{table}[h!]
    \centering
    \MStabcaption{tab:ablation}{Ablation study of core components on MTL-AQA. Performance variations are reported relative to ID 1.}
\resizebox{\linewidth}{!}{
    \begin{tabular}{ccccllll}
    \toprule
    \multirow{2.5}{*}{\textbf{ID}} & \multicolumn{3}{c}{\textbf{Setting}} & 
    \multirow{2.5}{*}{$\rho_{\text{avg}}$ ($\uparrow$)} & 
    \multirow{2.5}{*}{$\rho_{\text{aft}}$ ($\downarrow$)} & 
    \multirow{2.5}{*}{$\rho_{\text{fwt}}$ ($\uparrow$)} \\
    \cmidrule(lr){2-4}
    & LA-FPFT & MP & IIJ-GR \\
    \midrule
    \rowcolor{rowgrayyellow}  1 & \ding{51} & \ding{51} & \ding{51} 
      & 0.9205 
      & 0.0103 
      & 0.1274 \\

    \rowcolor{rowgraypink} 2 & \ding{53} & \ding{51} & \ding{51} 
      & 0.9135$^{-1\%}$ 
      & 0.0233$^{+126\%}$ 
      & 0.1204$^{-5\%}$ \\

    \rowcolor{rowgraypink} 3 & \ding{51} & \ding{53} & \ding{51} 
      & 0.8418$^{-9\%}$ 
      & 0.0995$^{+866\%}$ 
      & 0.1082$^{-15\%}$ \\

    \rowcolor{rowgraypink} 4 & \ding{51} & \ding{51} & \ding{53} 
      & 0.8548$^{-7\%}$ 
      & 0.0143$^{+39\%}$ 
      & 0.1211$^{-5\%}$ \\

    \rowcolor{rowgraypink} 5 & \ding{53} & \ding{53} & \ding{51} 
      & 0.8016$^{-13\%}$ 
      & 0.1427$^{+1285\%}$ 
      & 0.0965$^{-24\%}$ \\

    \rowcolor{rowgraypink} 6 & \ding{51} & \ding{53} & \ding{53} 
      & 0.7725$^{-16\%}$ 
      & 0.0687$^{+567\%}$ 
      & 0.0984$^{-23\%}$ \\

    \rowcolor{rowgraypink} 7 & \ding{53} & \ding{51} & \ding{53} 
      & 0.8356$^{-9\%}$ 
      & 0.0286$^{+178\%}$ 
      & 0.1135$^{-11\%}$ \\

    \rowcolor{rowgraypink} 8 & \ding{53} & \ding{53} & \ding{53} 
      & 0.7283$^{-21\%}$ 
      & 0.2255$^{+2090\%}$ 
      & 0.0535$^{-58\%}$ \\
    \bottomrule
    \end{tabular}
    }
\end{table}

\myPara{Impact of Design Choices.}
\cref{tab:ablation_mtl} compares alternative designs for each component.
In \cref{tab:ablation_mtl-a}, LA-FPFT achieves the best balance between average performance and forgetting, outperforming standard FPFT, PEFT, and a fixed backbone.
For manifold projection (see \cref{tab:ablation_mtl-b}), the proposed two-stage MLP-based MP outperforms the one-stage variant used in MAGR~\cite{zhou2024magr}, with $\rho_{\text{avg}}$ decreasing by 3\% when the latter is used.
We also explicitly compare our residual MP with a direct-mapping drift-compensation variant following the paradigm of Gomez-Villa et al.~\cite{gomez2024exemplar}. This variant directly learns the transformation from the previous feature space to the current one, corresponding to MP without the residual formulation. It achieves $\rho_{\text{avg}}=0.8933$ and $\rho_{\text{aft}}=0.0389$, compared with $0.9205$ and $0.0103$ for our residual MP, respectively. Replacing the MLP projector with a Transformer-based design (see ID 3 in \cref{tab:ablation_mtl-b}) does not improve performance and slightly increases forgetting.
This is because MP is formulated as a residual learning problem, where the projector only models the incremental shift between adjacent sessions.
Such shifts are typically smooth and low-magnitude, making residual correction more effective and stable than learning a full mapping with a heavier Transformer.
For graph regularization (see \cref{tab:ablation_mtl-c}), jointly modeling local (II-GR) and global (J-GR) relations (see ID 1 in \cref{tab:ablation_mtl-c}) outperforms using either alone (see ID 2 and 3 in \cref{tab:ablation_mtl-c}), and the OUS strategy (see ID 1 in \cref{tab:ablation_mtl-d}) consistently outperforms random sampling (see ID 2 in \cref{tab:ablation_mtl-d}).

\begin{table}[h!]
    \centering
\MStabcaption{tab:ablation_mtl}{Ablation study of design choices for core components in MTL-AQA. Performance variations are reported relative to the full model (ID 1).}{}
\begin{minipage}{\linewidth}
        \subcaption{Fine-Tuning Strategies} \label{tab:ablation_mtl-a}
        \resizebox{\linewidth}{!}{
        \begin{tabular}{cp{2.8cm}lll}
        \toprule
        \textbf{ID} & \textbf{Setting} & $\rho_{\text{avg}}$ ($\uparrow$) & $\rho_{\text{aft}}$ ($\downarrow$) & \multicolumn{1}{c}{$\rho_{\text{fwt}}$ ($\uparrow$)} \\
            \midrule
            \rowcolor{rowgrayyellow} 1 & Ours w/ LA FPFT & 0.9205 & 0.0103 & 0.1274 \\
            \rowcolor{rowgraypink} 2 & ~~ w/ FPFT & 0.9135$^{-1\%}$ & 0.0233$^{+126\%}$ & 0.1204$^{-5\%}$ \\
            \rowcolor{rowgraypink} 3 & ~~ w/ Fixed Backbone & 0.4855$^{-47\%}$ & 0.0523$^{+408\%}$ & 0.0746$^{-41\%}$ \\
            \rowcolor{rowgraypink} 4 & ~~ w/ PEFT (Adapters) & 0.8703$^{-5\%}$ & 0.0180$^{+75\%}$ & 0.2991$^{+135\%}$ \\
            \bottomrule
        \end{tabular}
        }
    \end{minipage}

    \begin{minipage}{\linewidth}
        \vspace{2mm}
        \subcaption{Manifold Projection Strategies} \label{tab:ablation_mtl-b}
        \resizebox{\linewidth}{!}{
        \begin{tabular}{cp{2.8cm}lll}
        \toprule
        \textbf{ID} & \textbf{Setting} & $\rho_{\text{avg}}$ ($\uparrow$) & $\rho_{\text{aft}}$ ($\downarrow$) & \multicolumn{1}{c}{$\rho_{\text{fwt}}$ ($\uparrow$)} \\
            \midrule
            \rowcolor{rowgrayyellow} 1 & Ours w/ MP & 0.9205 & 0.0103 & 0.1274 \\
            \rowcolor{rowgraypink} 2 & ~~w/ Direct Mapping \cite{gomez2024exemplar} & 0.8933$^{-3\%}$ & 0.0389$^{+278\%}$ & 0.0693$^{-46\%}$  \\
            \rowcolor{rowgraypink} 3 & ~~w/ Transformer MP 
            & 0.9068$^{-2\%}$ 
            & 0.0196$^{+90\%}$ 
            & 0.1208$^{-5\%}$ \\
            \rowcolor{rowgraypink} 4 & ~~w/ One-Stage MP & 0.8944$^{-3\%}$ & 0.0355$^{+245\%}$ & 0.1149$^{-10\%}$ \\
            \bottomrule
        \end{tabular}
        }
    \end{minipage}

    \begin{minipage}{\linewidth}
        \vspace{2mm}
        \subcaption{Graph Regularization Strategies} \label{tab:ablation_mtl-c}
        \resizebox{\linewidth}{!}{
        \begin{tabular}{cp{2.8cm}lll}
        \toprule
        \textbf{ID} & \textbf{Setting} & $\rho_{\text{avg}}$ ($\uparrow$) & $\rho_{\text{aft}}$ ($\downarrow$) & \multicolumn{1}{c}{$\rho_{\text{fwt}}$ ($\uparrow$)} \\
            \midrule
            \rowcolor{rowgrayyellow} 1 & Ours w/ IIJ-GR & 0.9205 & 0.0103 & 0.1274 \\
            \rowcolor{rowgraypink} 2 &~~w/ J-GR (w/o II-GR)   & 0.8730$^{-5\%}$ & 0.0131$^{+27\%}$ & 0.1095$^{-14\%}$ \\
            \rowcolor{rowgraypink} 3 &~~w/ II-GR (w/o J-GR)   & 0.8991$^{-2\%}$ & 0.0314$^{+205\%}$ & 0.1586$^{+25\%}$ \\
            \bottomrule
        \end{tabular}
        }
    \end{minipage}

    \begin{minipage}{\linewidth}
        \vspace{2mm}
        \subcaption{Sampling Strategies} \label{tab:ablation_mtl-d}
        \resizebox{\linewidth}{!}{
        \begin{tabular}{cp{2.8cm}lll}
        \toprule
        \textbf{ID} & \textbf{Setting} & $\rho_{\text{avg}}$ ($\uparrow$) & $\rho_{\text{aft}}$ ($\downarrow$) & \multicolumn{1}{c}{$\rho_{\text{fwt}}$ ($\uparrow$)} \\
            \midrule
            \rowcolor{rowgrayyellow} 1 & Ours w/ OUS & 0.9205 & 0.0103 & 0.1274 \\
            \rowcolor{rowgraypink} 2 & ~~w/ Random Sampling & 0.9143$^{-1\%}$ & 0.0397$^{+285\%}$ & 0.1191$^{-7\%}$ \\
            \bottomrule
        \end{tabular}
        }
    \end{minipage}
    \vspace{-0.10cm}
\end{table}

\myPara{Impact of Memory Size.}
\cref{fig:memory_size} depicts the trade-off between accuracy and storage by varying the number of replayed samples per session. Most baselines, such as DER++ and MER, exhibit relatively flat performance across different sizes (e.g., DER++ $\rho_{\text{avg}}$ ranges narrowly from 0.8383 to 0.8434), indicating limited sensitivity. In contrast, MAGR benefits substantially from larger memory: $\rho_{\text{avg}}$ improves from 0.6750 at 3 samples to 0.8918 at 11 samples. Our MAGR++ method consistently outperforms all competitors under every setting, achieving the best balance between accuracy and forgetting. For instance, with only 3 samples per session, our method already achieves $\rho_{\text{avg}}=0.9066$ and $\rho_{\text{aft}}=0.0173$, clearly outperforming MAGR (0.6750, 0.2107) and NC-FSCIL (0.7751, 0.1536). This advantage stems from the fact that MAGR++ more effectively addresses feature shifts and better exploits limited samples to recover underlying data distributions, thereby mitigating forgetting. We choose 10 samples per session for a fair comparison in \cref{sec:sota_cmp}.

\begin{figure}
    \centering
    \includegraphics[width=\linewidth]{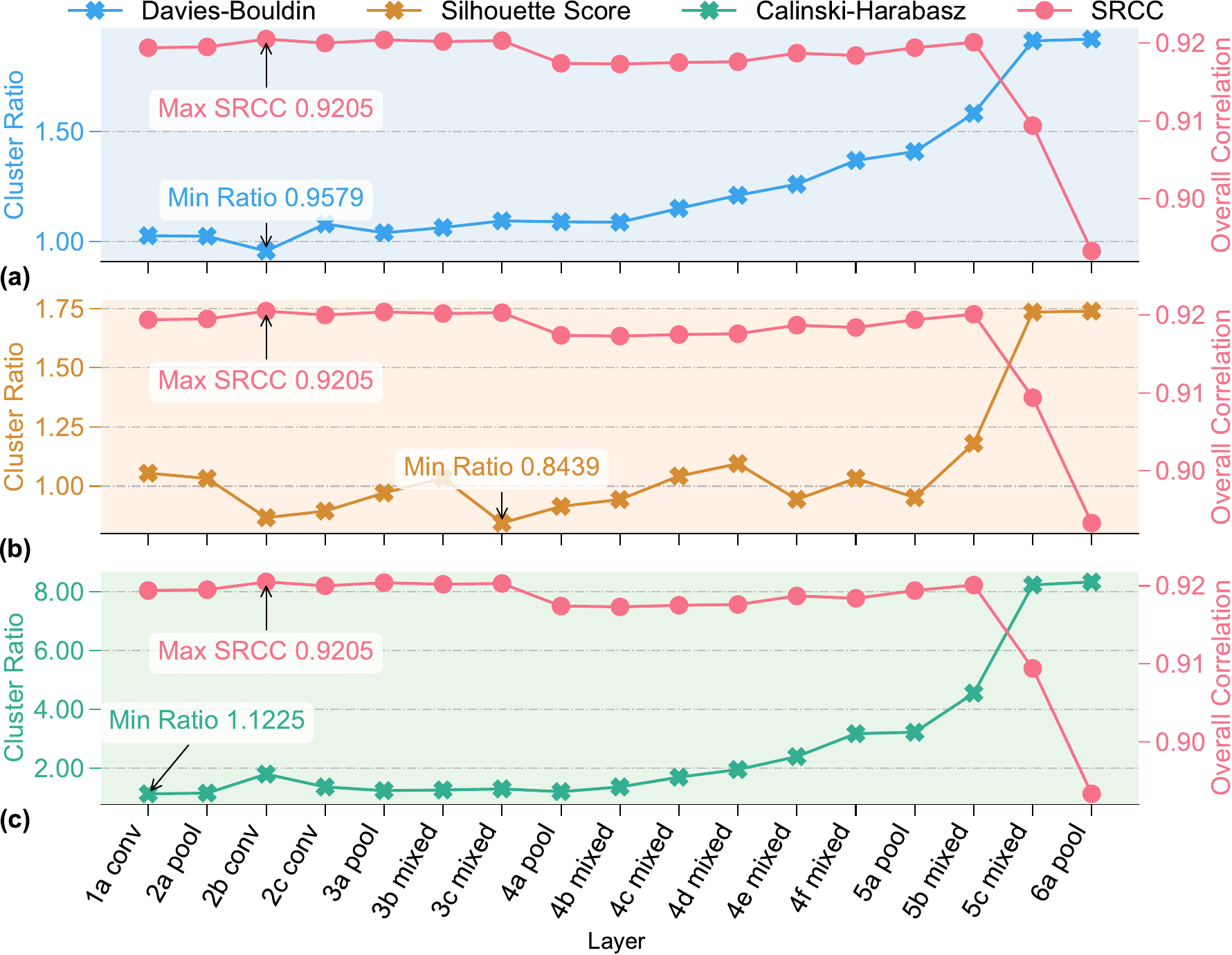}
    \MSfigcaption{fig:cluster}{Cluster quality ratio and overall correlation.}{
    \phantomsubcaption\label{fig:cluster-a}
    \phantomsubcaption\label{fig:cluster-b}
    \phantomsubcaption\label{fig:cluster-c}
    }
\end{figure}

\myPara{Impact of Feature Supervision Layer.}  
\cref{fig:cluster} compares different clustering quality metrics for selecting the feature supervision layer in layer-adaptive FPFT. Among the evaluated metrics, the Davies–Bouldin index (see \cref{fig:cluster-a}) shows the strongest and most consistent correlation with downstream performance, while alternative measures such as the Silhouette score (see \cref{fig:cluster-b}) and the Calinski–Harabasz index (see \cref{fig:cluster-c}) exhibit similar but slightly less stable trends, supporting the Davies–Bouldin index as a reliable criterion for boundary selection.  
We further examine the sensitivity of the layer selection threshold $\epsilon$ in \cref{fig:epsilon}. While the best performance is obtained when $\epsilon \in [0.025, 0.080]$, the method continues to select a competitive boundary and maintain strong performance even outside this interval, as long as $\epsilon < 0.9$. This indicates that the proposed layer-adaptive strategy is robust to threshold choice and does not require careful tuning of $\epsilon$.  
Under this criterion, empirical results in \cref{fig:cluster,fig:epsilon} identify the third layer as the optimal supervision point, which aligns with the outcome of our dynamic layer selection strategy. 
This agrees with the hierarchy of deep representations: earlier layers encode transferable low-level patterns, whereas higher layers capture task-specific semantics requiring adaptation. Supervising the third layer therefore balances generalizable representations with deeper-layer specialization.

\begin{figure}
    \centering
    \includegraphics[width=\linewidth]{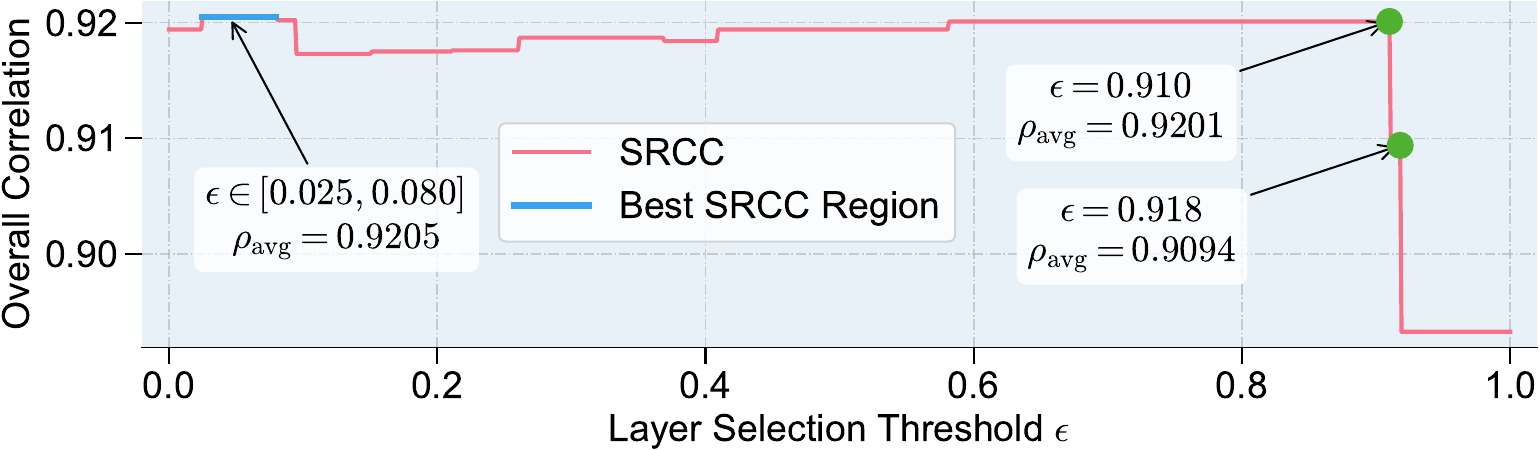}
    \MSfigcaption{fig:epsilon}{Impact of layer selection threshold $\epsilon$.}{}
\vspace{-0.25cm}
\end{figure}

\myPara{Impact of Task Order.}
In practice, human skill progression is not strictly linear. To assess the impact of varying task sequences, we shuffled the task order multiple times and measured both performance and parameter changes (see \cref{fig:parameter_change}). Our method shows strong robustness to task-order variation, achieving an average performance of $0.9183 \pm 0.0028$, which is not only more stable but also higher than FS-Aug and NC-FSCIL. Moreover, in terms of parameter dynamics, our approach exhibits larger yet more structured parameter changes while maintaining higher performance and lower forgetting (see \cref{fig:parameter_change-b}), highlighting its superior ability to manage catastrophic forgetting caused by parameter shifts.

\myPara{Impact of Semi-Supervision and Noise.}
\cref{fig:sol} evaluates MAGR++ under semi-supervised settings with limited labels (see \cref{fig:sol-a}) and in the presence of noisy annotations (see \cref{fig:sol-d}), two realistic and challenging issues in AQA that directly impact model generalization. Compared to MAGR, MAGR++ achieves consistently higher performance owing to its two-stage feature rectification and effective FPFT backbone tuning. With only 10 samples per session, MAGR++ reaches $\rho_{\text{avg}}=0.9162$ compared to MAGR’s 0.8741, and the margin remains at 25 samples (0.9256 vs.\ 0.8951). Under noise, MAGR++ is notably more resilient, sustaining 0.9234 at intensity 7 (see \cref{fig:sol-f}, $\hat{y}_i = 0.94 y_i + 3.04$) while MAGR drops sharply to 0.7813 (see \cref{fig:sol-e}, $\hat{y}_i = 0.65 y_i + 32.17$). Beyond MAGR, MAGR++ also outperforms other baselines: at 15 samples it achieves 0.9205 compared to 0.8950 for SLCA and 0.8142 for NC-FSCIL, and at noise level 7 it yields 0.9234 compared to 0.8032 for SLCA (see \cref{fig:sol-b}, $\hat{y}_i = 0.53 y_i + 37.34$), 0.7548 for NC-FSCIL (see \cref{fig:sol-c}, $\hat{y}_i = 0.73 y_i + 24.71$), while TOPIC lags further at 0.7049. These results verify the superior robustness of MAGR++ under both label scarcity and noisy supervision.

\subsection{Qualitative and Quantitative Results}
We provide additional experimental results to further validate the effectiveness of our proposed method.  

\begin{figure}
    \centering
    \includegraphics[width=\linewidth]{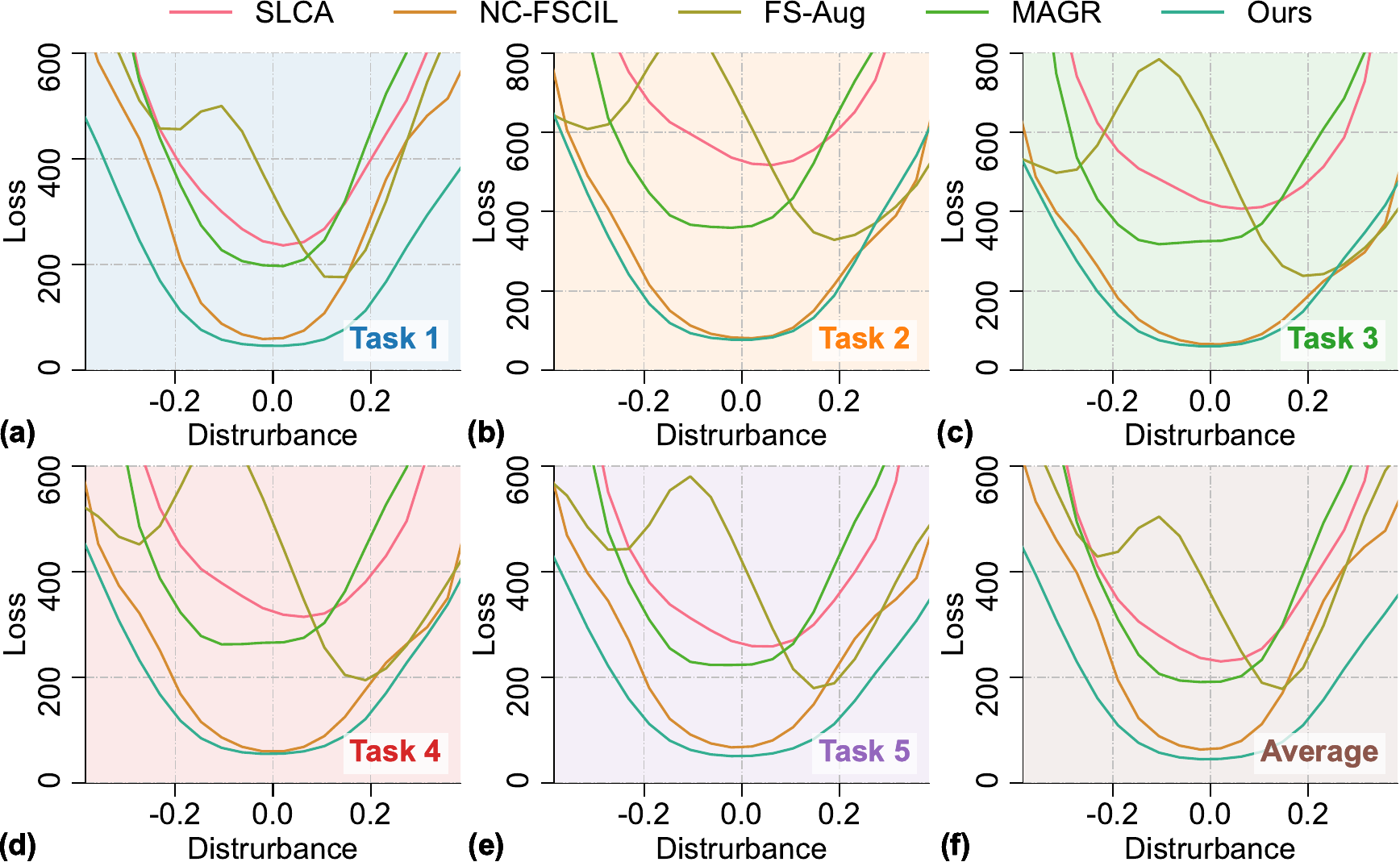}
    \caption{Loss landscapes on MTL-AQA. \subref{fig:loss_landscape-a}-\subref{fig:loss_landscape-e} show loss curves for five sessions, while \subref{fig:loss_landscape-f} presents the average curve.}
    \label{fig:loss_landscape}
    \phantomsubcaption\label{fig:loss_landscape-a}
    \phantomsubcaption\label{fig:loss_landscape-b}
    \phantomsubcaption\label{fig:loss_landscape-c}
    \phantomsubcaption\label{fig:loss_landscape-d}
    \phantomsubcaption\label{fig:loss_landscape-e}
    \phantomsubcaption\label{fig:loss_landscape-f}
\end{figure}

\myPara{Flatness of Loss Landscape.}  
To assess model generalization, we visualize the flatness of the loss landscape \cite{deng2021flattening}. After training each session, model weights are perturbed along 10 random directions, and the resulting loss curves are averaged across perturbation magnitudes. \cref{fig:loss_landscape-a,fig:loss_landscape-b,fig:loss_landscape-c,fig:loss_landscape-d,fig:loss_landscape-e} illustrate session-wise results, while \cref{fig:loss_landscape-f} shows the average over all five sessions.  
MAGR++ consistently exhibits flatter and lower loss landscapes than all other baselines, suggesting it converges to more stable minima that enhance generalization and mitigate catastrophic forgetting. In contrast, FS-Aug produces uneven and fluctuating curves, reflecting sensitivity to noise and a lack of robustness in its learned representations.  

\begin{figure}
    \centering
    \includegraphics[width=\linewidth]{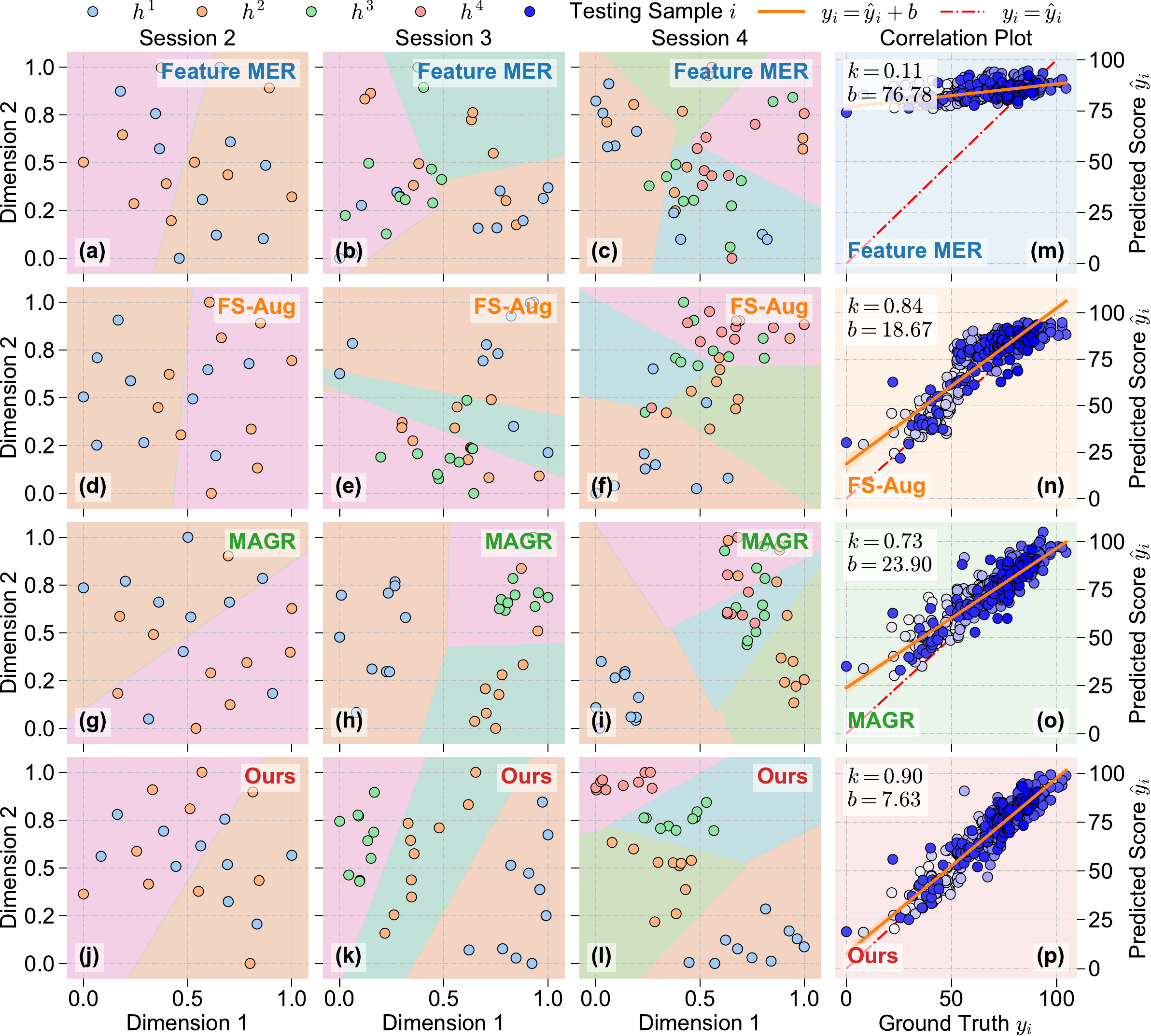}
    \caption{Visualization of t-SNE feature distributions (first three columns) and overall correlation plots (last column). The feature space is projected into two dimensions and normalized to $[0,1]$. }
    \label{fig:tsne}
    \phantomsubcaption\label{fig:tsne-a}
    \phantomsubcaption\label{fig:tsne-b}
    \phantomsubcaption\label{fig:tsne-c}
    \phantomsubcaption\label{fig:tsne-d}
    \phantomsubcaption\label{fig:tsne-e}
    \phantomsubcaption\label{fig:tsne-f}
    \phantomsubcaption\label{fig:tsne-g}
    \phantomsubcaption\label{fig:tsne-h}
    \phantomsubcaption\label{fig:tsne-i}
    \phantomsubcaption\label{fig:tsne-j}
    \phantomsubcaption\label{fig:tsne-k}
    \phantomsubcaption\label{fig:tsne-l}
    \phantomsubcaption\label{fig:tsne-m}
    \phantomsubcaption\label{fig:tsne-n}
    \phantomsubcaption\label{fig:tsne-o}
    \phantomsubcaption\label{fig:tsne-p}
    \vspace{-0.2cm}
\end{figure}

\myPara{Visualization of Addressing Feature Shifts.}  
To better understand catastrophic forgetting, we visualize feature distributions and scatter correlation in \cref{fig:tsne}. In \cref{fig:tsne-a,fig:tsne-b,fig:tsne-c,fig:tsne-m}, MER exhibits severe manifold shifts where samples from different sessions are highly entangled, leading to a poorly aligned regression line $\hat{y}_i = 0.11 y_i + 76.78$. FS-Aug and MAGR alleviate this overlap but still suffer from residual shifts, with regression lines remaining misaligned (e.g., MAGR: $\hat{y}_i = 0.73 y_i + 23.90$ (see \cref{fig:tsne-n})). In contrast, our method explicitly separates samples across sessions (see \cref{fig:tsne-j,fig:tsne-k,fig:tsne-l})  and achieves a much closer regression fit $\hat{y}_i = 0.90 y_i + 7.63$ (see \cref{fig:tsne-p}), approaching the ideal line $\hat{y}_i = y_i$. These results indicate that our approach effectively mitigates feature shifts, preserving both feature space consistency and correlation accuracy across tasks.  

\begin{figure}[t]
    \centering
    \includegraphics[width=\linewidth]{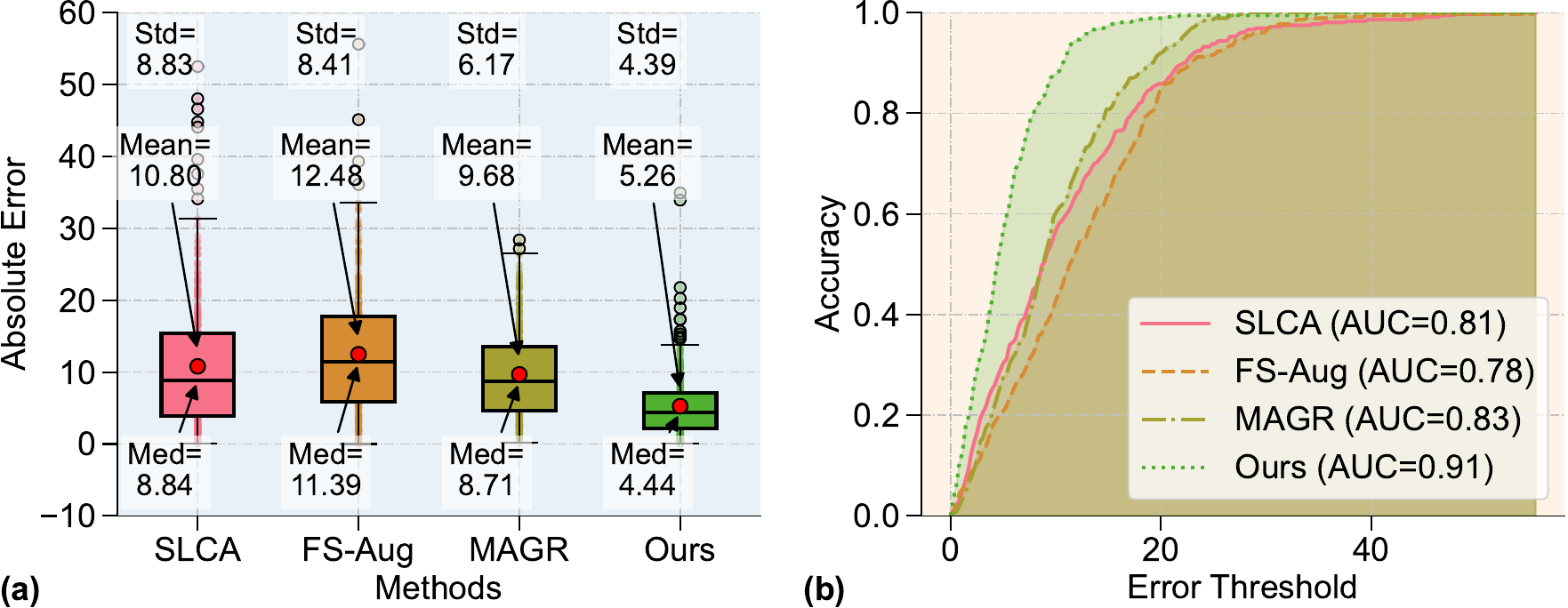}
    \caption{
        Error analysis on MTL-AQA. 
        \subref{fig:error_statistic-a}: Boxplots of absolute errors with mean, median, and standard deviation. 
        \subref{fig:error_statistic-b}: Cumulative error accuracy curves with area under the curve (AUC).
    }
    \label{fig:error_statistic}
    \phantomsubcaption\label{fig:error_statistic-a}
    \phantomsubcaption\label{fig:error_statistic-b}
    \vspace{-0.25cm}
\end{figure}

\myPara{Error Analysis.}  
\cref{fig:error_statistic} compares error distributions and cumulative accuracy among baselines and MAGR++.  
SLCA, FS-Aug, and MAGR achieve mean absolute errors of $10.80 \pm 8.83$, $12.48 \pm 8.41$, and $9.68 \pm 6.17$, respectively.  
In contrast, MAGR++ significantly reduces the error to $5.26 \pm 4.39$, indicating both lower bias and variance.  
The cumulative accuracy curves further emphasize this gain: MAGR++ reaches an AUC of $0.91$, surpassing SLCA ($0.81$), FS-Aug ($0.78$), and MAGR ($0.83$) by margins of $0.10$, $0.13$, and $0.08$, respectively.  
These results confirm the effectiveness of shift-aware rectification in improving both accuracy and stability for CAQA.  
 
\section{Discussion and Conclusion} \label{sec:conclusion}

In this work, we formalize and systematically extend continual AQA as a fine-grained continual regression problem. To support this setting, we construct four comprehensive benchmarks with tailored evaluation protocols and strong baselines. Through an empirical study, we find that sufficiently expressive backbone adaptation is important for bridging the gap between upstream action recognition and downstream fine-grained quality assessment, while our theoretical analysis further reveals that aggressive full-parameter adaptation can amplify forgetting and replay-feature drift during continual learning.
To address these challenges, we propose MAGR++, which combines layer-adaptive fine-tuning for controlled representation plasticity, residual manifold projection for rectifying stale replay features, and score-aware graph regularization for preserving cross-session quality structure. Experiments show that MAGR++ consistently achieves strong performance across diverse offline and online CAQA benchmarks.
More broadly, this study highlights two challenges that extend beyond CAQA: controlling representation plasticity under expressive continual adaptation, and maintaining the validity of replayed features as the representation space evolves. The proposed layer-adaptive tuning and residual feature rectification provide concrete mechanisms for addressing these challenges, and may also benefit other continual fine-grained regression problems with evolving representations. 
Despite these promising results, several avenues remain open. First, our current formulation focuses on CL of AQA tasks. Extending MAGR++ to more general continual fine-grained prediction scenarios could further validate its applicability. Second, integrating multi-modal inputs (e.g., audio and text) may improve robustness and effectiveness. Third, improving the efficiency of online adaptation could further facilitate real-time deployment. We view these directions as natural extensions toward advancing continual AQA and related fine-grained prediction problems.

\bibliographystyle{ieeetr}

\begin{thebibliography}{10}

\bibitem{gedamu2024self}
K.~Gedamu, Y.~Ji, Y.~Yang, J.~Shao, and H.~T. Shen, ``Self-supervised subaction
  parsing network for semi-supervised action quality assessment,'' {\em IEEE
  Transactions on Image Processing}, vol.~33, pp.~6057--6070, 2024.

\bibitem{majeedi2024rica}
A.~Majeedi, V.~R. Gajjala, S.~S. S.~N. GNVV, and Y.~Li, ``Rica$^2$:
  Rubric-informed, calibrated assessment of actions,'' in {\em European
  Conference on Computer Vision}, vol.~15121, pp.~143--161, 2024.

\bibitem{dong2024interpretable}
X.~Dong, X.~Liu, W.~Li, A.~Adeyemi-Ejeye, and A.~Gilbert, ``Interpretable
  long-term action quality assessment,'' {\em arXiv preprint arXiv:2408.11687},
  2024.

\bibitem{liu2025adaptive}
J.~Liu, H.~Wang, W.~Zhou, K.~Stawarz, P.~Corcoran, Y.~Chen, and H.~Liu,
  ``Adaptive spatiotemporal graph transformer network for action quality
  assessment,'' {\em IEEE Transactions on Circuits and Systems for Video
  Technology}, vol.~35, no.~7, pp.~6628--6639, 2025.

\bibitem{xu2025dancefix}
H.~Xu, X.~Ke, H.~Wu, R.~Xu, Y.~Li, P.~Xu, and W.~Guo, ``Dancefix: An
  exploration in group dance neatness assessment through fixing abnormal
  challenges of human pose,'' in {\em Association for the Advancement of
  Artificial Intelligence}, vol.~39, pp.~8869--8877, 2025.

\bibitem{zeng2024multimodal}
L.-A. Zeng and W.-S. Zheng, ``Multimodal action quality assessment,'' {\em IEEE
  Transactions on Image Processing}, vol.~33, pp.~1600--1613, 2024.

\bibitem{ji2023localization}
Y.~Ji, L.~Ye, H.~Huang, L.~Mao, Y.~Zhou, and L.~Gao, ``Localization-assisted
  uncertainty score disentanglement network for action quality assessment,'' in
  {\em ACM International Conference on Multimedia}, pp.~8590--8597, 2023.

\bibitem{zhang2023logo}
S.~Zhang, W.~Dai, S.~Wang, X.~Shen, J.~Lu, J.~Zhou, and Y.~Tang, ``Logo: A
  long-form video dataset for group action quality assessment,'' in {\em
  IEEE/CVF Conference on Computer Vision and Pattern Recognition},
  pp.~2405--2414, 2023.

\bibitem{zhou2023video}
K.~Zhou, R.~Cai, Y.~Ma, Q.~Tan, X.~Wang, J.~Li, H.~P. Shum, F.~W. Li, S.~Jin,
  and X.~Liang, ``A video-based augmented reality system for human-in-the-loop
  muscle strength assessment of juvenile dermatomyositis,'' {\em IEEE
  Transactions on Visualization and Computer Graphics}, vol.~29, no.~5,
  pp.~2456--2466, 2023.

\bibitem{parmar2021piano}
P.~Parmar, J.~Reddy, and B.~Morris, ``Piano skills assessment,'' in {\em IEEE
  International Workshop on Multimedia Signal Processing}, pp.~1--5, IEEE,
  2021.

\bibitem{li2022surgical}
Z.~Li, L.~Gu, W.~Wang, R.~Nakamura, and Y.~Sato, ``Surgical skill assessment
  via video semantic aggregation,'' in {\em International Conference on Medical
  Image Computing and Computer-Assisted Intervention}, vol.~13437,
  pp.~410--420, Springer, 2022.

\bibitem{carreira2017quo}
J.~Carreira and A.~Zisserman, ``Quo vadis, action recognition? a new model and
  the kinetics dataset,'' in {\em IEEE/CVF Conference on Computer Vision and
  Pattern Recognition}, pp.~6299--6308, 2017.

\bibitem{kay2017kinetics}
W.~Kay, J.~Carreira, K.~Simonyan, B.~Zhang, C.~Hillier, S.~Vijayanarasimhan,
  F.~Viola, T.~Green, T.~Back, P.~Natsev, {\em et~al.}, ``The kinetics human
  action video dataset,'' {\em arXiv preprint arXiv:1705.06950}, 2017.

\bibitem{dadashzadeh2024pecop}
A.~Dadashzadeh, S.~Duan, A.~Whone, and M.~Mirmehdi, ``Pecop: Parameter
  efficient continual pretraining for action quality assessment,'' in {\em IEEE
  Winter Conference on Applications of Computer Vision}, pp.~42--52, 2024.

\bibitem{zhou2024cofinal}
K.~Zhou, J.~Li, R.~Cai, L.~Wang, X.~Zhang, and X.~Liang, ``Cofinal: Enhancing
  action quality assessment with coarse-to-fine instruction alignment,'' in
  {\em International Joint Conference on Artificial Intelligence},
  pp.~1771--1779, 2024.

\bibitem{zhou2025phi}
K.~Zhou, H.~P. Shum, F.~W. Li, X.~Zhang, and X.~Liang, ``Phi: Bridging domain
  shift in long-term action quality assessment via progressive hierarchical
  instruction,'' {\em IEEE Transactions on Image Processing}, vol.~34,
  pp.~3718--3732, 2025.

\bibitem{wang2024comprehensive}
L.~Wang, X.~Zhang, H.~Su, and J.~Zhu, ``A comprehensive survey of continual
  learning: Theory, method and application,'' {\em IEEE Transactions on Pattern
  Analysis and Machine Intelligence}, vol.~46, no.~8, pp.~5362--5383, 2024.

\bibitem{chi2022metafscil}
Z.~Chi, L.~Gu, H.~Liu, Y.~Wang, Y.~Yu, and J.~Tang, ``Metafscil: A
  meta-learning approach for few-shot class incremental learning,'' in {\em
  IEEE/CVF Conference on Computer Vision and Pattern Recognition},
  pp.~14166--14175, 2022.

\bibitem{wang2023hierarchical}
L.~Wang, J.~Xie, X.~Zhang, M.~Huang, H.~Su, and J.~Zhu, ``Hierarchical
  decomposition of prompt-based continual learning: Rethinking obscured
  sub-optimality,'' {\em Advances in Neural Information Processing Systems},
  vol.~36, pp.~69054--69076, 2023.

\bibitem{wang2021afec}
L.~Wang, M.~Zhang, Z.~Jia, Q.~Li, C.~Bao, K.~Ma, J.~Zhu, and Y.~Zhong, ``Afec:
  Active forgetting of negative transfer in continual learning,'' {\em Advances
  in Neural Information Processing Systems}, vol.~34, pp.~22379--22391, 2021.

\bibitem{zhou2025adaptive}
K.~Zhou, Z.~Hao, L.~Wang, and X.~Liang, ``Adaptive score alignment learning for
  continual perceptual quality assessment of 360-degree videos in virtual
  reality,'' {\em IEEE Transactions on Visualization and Computer Graphics},
  vol.~31, no.~5, pp.~2880--2890, 2025.

\bibitem{xin2024parameter}
Y.~Xin {\em et~al.}, ``Parameter-efficient fine-tuning for pre-trained vision
  models: A survey and benchmark,'' {\em International Journal of Computer
  Vision}, vol.~134, Art.~no.~304, 2026.

\bibitem{zhou2026comprehensive}
K.~Zhou, R.~Cai, L.~Wang, H.~P.~H. Shum, and X.~Liang, ``A comprehensive survey
  of action quality assessment: Method and benchmark,'' {\em Pattern
  Recognition}, vol.~179, p.~113933, 2026.

\bibitem{zhou2024magr}
K.~Zhou, L.~Wang, X.~Zhang, H.~P. Shum, F.~W. Li, J.~Li, and X.~Liang, ``Magr:
  Manifold-aligned graph regularization for continual action quality
  assessment,'' in {\em European Conference on Computer Vision}, vol.~15069,
  pp.~375--392, 2024.

\bibitem{bai2022action}
Y.~Bai, D.~Zhou, S.~Zhang, J.~Wang, E.~Ding, Y.~Guan, Y.~Long, and J.~Wang,
  ``Action quality assessment with temporal parsing transformer,'' in {\em
  European Conference on Computer Vision}, vol.~13664, pp.~422--438, Springer,
  2022.

\bibitem{pan2025basket}
Y.~Pan, C.~Zhang, and G.~Bertasius, ``Basket: A large-scale video dataset for
  fine-grained skill estimation,'' in {\em IEEE/CVF Conference on Computer
  Vision and Pattern Recognition}, pp.~28952--28962, 2025.

\bibitem{xu2025mcmoe}
H.~Xu, H.~Wu, X.~Ke, J.~Wu, R.~Xu, and J.~Xu, ``Mcmoe: Completing missing
  modalities with mixture of experts for incomplete multimodal action quality
  assessment,'' in {\em Proceedings of the AAAI Conference on Artificial
  Intelligence}, vol.~40, no.~13, pp.~11241--11249, 2026.

\bibitem{zia2016automated}
A.~Zia, Y.~Sharma, V.~Bettadapura, E.~L. Sarin, T.~Ploetz, M.~A. Clements, and
  I.~Essa, ``Automated video-based assessment of surgical skills for training
  and evaluation in medical schools,'' {\em International Journal of Computer
  Assisted Radiology and Surgery}, vol.~11, pp.~1623--1636, 2016.

\bibitem{pan2021adaptive}
J.-H. Pan, J.~Gao, and W.-S. Zheng, ``Adaptive action assessment,'' {\em IEEE
  Transactions on Pattern Analysis and Machine Intelligence}, vol.~44, no.~12,
  pp.~8779--8795, 2022.

\bibitem{yu2021group}
X.~Yu, Y.~Rao, W.~Zhao, J.~Lu, and J.~Zhou, ``Group-aware contrastive
  regression for action quality assessment,'' in {\em IEEE/CVF International
  Conference on Computer Vision}, pp.~7919--7928, 2021.

\bibitem{parmar2019and}
P.~Parmar and B.~T. Morris, ``What and how well you performed? a multitask
  learning approach to action quality assessment,'' in {\em IEEE/CVF Conference
  on Computer Vision and Pattern Recognition}, pp.~304--313, 2019.

\bibitem{zhou2023hierarchical}
K.~Zhou, Y.~Ma, H.~P. Shum, and X.~Liang, ``Hierarchical graph convolutional
  networks for action quality assessment,'' {\em IEEE Transactions on Circuits
  and Systems for Video Technology}, vol.~33, no.~12, pp.~7749--7763, 2023.

\bibitem{ke2024two}
X.~Ke, H.~Xu, X.~Lin, and W.~Guo, ``Two-path target-aware contrastive
  regression for action quality assessment,'' {\em Information Sciences},
  vol.~664, p.~120347, 2024.

\bibitem{doughty2019pros}
H.~Doughty, W.~Mayol-Cuevas, and D.~Damen, ``The pros and cons: Rank-aware
  temporal attention for skill determination in long videos,'' in {\em IEEE/CVF
  Conference on Computer Vision and Pattern Recognition}, pp.~7862--7871, 2019.

\bibitem{doughty2018s}
H.~Doughty, D.~Damen, and W.~Mayol-Cuevas, ``Who's better? who's best? pairwise
  deep ranking for skill determination,'' in {\em IEEE/CVF Conference on
  Computer Vision and Pattern Recognition}, pp.~6057--6066, 2018.

\bibitem{dong2026uila}
X.~Dong, X.~Liu, W.~Li, A.~Adeyemi-Ejeye, and A.~Gilbert, ``Uil-aqa:
  Uncertainty-aware clip-level interpretable action quality assessment,'' {\em
  International Journal of Computer Vision}, vol.~134, Art.~no.~24, 2026.

\bibitem{xu2024fineparser}
J.~Xu, S.~Yin, G.~Zhao, Z.~Wang, and Y.~Peng, ``Fineparser: A fine-grained
  spatio-temporal action parser for human-centric action quality assessment,''
  in {\em IEEE/CVF Conference on Computer Vision and Pattern Recognition},
  pp.~14628--14637, 2024.

\bibitem{xu2025human}
J.~Xu, S.~Yin, and Y.~Peng, ``Human-centric fine-grained action quality
  assessment,'' {\em IEEE Transactions on Pattern Analysis and Machine
  Intelligence}, vol.~47, no.~8, pp.~6242--6255, 2025.

\bibitem{wang2021tsa}
S.~Wang, D.~Yang, P.~Zhai, C.~Chen, and L.~Zhang, ``Tsa-net: Tube
  self-attention network for action quality assessment,'' in {\em ACM
  International Conference on Multimedia}, pp.~4902--4910, 2021.

\bibitem{qi2025action}
M.~Qi, H.~Ye, J.~Peng, and H.~Ma, ``Action quality assessment via hierarchical
  pose-guided multi-stage contrastive regression,'' {\em IEEE Transactions on
  Image Processing}, vol.~34, pp.~6461--6474, 2025.

\bibitem{pan2019action}
J.-H. Pan, J.~Gao, and W.-S. Zheng, ``Action assessment by joint relation
  graphs,'' in {\em IEEE/CVF International Conference on Computer Vision},
  pp.~6331--6340, 2019.

\bibitem{xu2025quality}
H.~Xu, H.~Wu, X.~Ke, Y.~Li, R.~Xu, and W.~Guo, ``Quality-guided vision-language
  learning for long-term action quality assessment,'' {\em IEEE Transactions on
  Multimedia}, vol.~27, pp.~7326--7339, 2025.

\bibitem{xu2024vision}
H.~Xu, X.~Ke, Y.~Li, R.~Xu, H.~Wu, X.~Lin, and W.~Guo, ``Vision-language action
  knowledge learning for semantic-aware action quality assessment,'' in {\em
  European Conference on Computer Vision}, vol.~15100, pp.~423--440, 2024.

\bibitem{xu2025language}
H.~Xu, X.~Ke, H.~Wu, R.~Xu, Y.~Li, and W.~Guo, ``Language-guided audio-visual
  learning for long-term sports assessment,'' in {\em IEEE/CVF Conference on
  Computer Vision and Pattern Recognition}, pp.~23967--23977, 2025.

\bibitem{gedamu2023fine}
K.~Gedamu, Y.~Ji, Y.~Yang, J.~Shao, and H.~T. Shen, ``Fine-grained
  spatio-temporal parsing network for action quality assessment,'' {\em IEEE
  Transactions on Image Processing}, vol.~32, pp.~6386--6400, 2023.

\bibitem{huang2024egoexolearn}
Y.~Huang, G.~Chen, J.~Xu, M.~Zhang, L.~Yang, B.~Pei, H.~Zhang, L.~Dong,
  Y.~Wang, L.~Wang, {\em et~al.}, ``Egoexolearn: A dataset for bridging
  asynchronous ego-and exo-centric view of procedural activities in real
  world,'' in {\em IEEE/CVF Conference on Computer Vision and Pattern
  Recognition}, pp.~22072--22086, 2024.

\bibitem{li2024continual}
Y.-M. Li, L.-A. Zeng, J.-K. Meng, and W.-S. Zheng, ``Continual action
  assessment via task-consistent score-discriminative feature distribution
  modeling,'' {\em IEEE Transactions on Circuits and Systems for Video
  Technology}, vol.~34, no.~10, pp.~9112--9124, 2024.

\bibitem{wang2023incorporating}
L.~Wang, X.~Zhang, Q.~Li, M.~Zhang, H.~Su, J.~Zhu, and Y.~Zhong,
  ``Incorporating neuro-inspired adaptability for continual learning in
  artificial intelligence,'' {\em Nature Machine Intelligence}, vol.~5, no.~12,
  pp.~1356--1368, 2023.

\bibitem{zenke2017continual}
F.~Zenke, B.~Poole, and S.~Ganguli, ``Continual learning through synaptic
  intelligence,'' in {\em International Conference on Machine Learning},
  pp.~3987--3995, PMLR, 2017.

\bibitem{james2017ewc}
J.~Kirkpatrick, R.~Pascanu, N.~Rabinowitz, J.~Veness, G.~Desjardins, A.~A.
  Rusu, K.~Milan, J.~Quan, T.~Ramalho, A.~Grabska-Barwinska, D.~Hassabis,
  C.~Clopath, D.~Kumaran, and R.~Hadsell, ``Overcoming catastrophic forgetting
  in neural networks,'' {\em Proceedings of the National Academy of Sciences},
  vol.~114, no.~13, pp.~3521--3526, 2017.

\bibitem{li2017learning}
Z.~Li and D.~Hoiem, ``Learning without forgetting,'' {\em IEEE Transactions on
  Pattern Analysis and Machine Intelligence}, vol.~40, no.~12, pp.~2935--2947,
  2018.

\bibitem{riemer2019learning}
M.~Riemer, I.~Cases, R.~Ajemian, M.~Liu, I.~Rish, Y.~Tu, and G.~Tesauro,
  ``Learning to learn without forgetting by maximizing transfer and minimizing
  interference,'' in {\em International Conference on Learning
  Representations}, 2019.

\bibitem{buzzega2020dark}
P.~Buzzega, M.~Boschini, A.~Porrello, D.~Abati, and S.~Calderara, ``Dark
  experience for general continual learning: A strong, simple baseline,'' {\em
  Advances in Neural Information Processing Systems}, vol.~33,
  pp.~15920--15930, 2020.

\bibitem{tao2020few}
X.~Tao, X.~Hong, X.~Chang, S.~Dong, X.~Wei, and Y.~Gong, ``Few-shot
  class-incremental learning,'' in {\em IEEE/CVF Conference on Computer Vision
  and Pattern Recognition}, pp.~12183--12192, 2020.

\bibitem{kukleva2021generalized}
A.~Kukleva, H.~Kuehne, and B.~Schiele, ``Generalized and incremental few-shot
  learning by explicit learning and calibration without forgetting,'' in {\em
  IEEE/CVF International Conference on Computer Vision}, pp.~9020--9029, 2021.

\bibitem{eskandar2025star}
M.~Eskandar, T.~Imtiaz, D.~Hill, Z.~Wang, and J.~Dy, ``Star: Stability-inducing
  weight perturbation for continual learning,'' in {\em International
  Conference on Learning Representations}, 2025.

\bibitem{zhang2023slca}
G.~Zhang, L.~Wang, G.~Kang, L.~Chen, and Y.~Wei, ``Slca: Slow learner with
  classifier alignment for continual learning on a pre-trained model,'' in {\em
  IEEE/CVF International Conference on Computer Vision}, pp.~19148--19158,
  2023.

\bibitem{yang2023neural}
Y.~Yang, H.~Yuan, X.~Li, Z.~Lin, P.~Torr, and D.~Tao, ``Neural collapse
  inspired feature-classifier alignment for few-shot class incremental
  learning,'' in {\em International Conference on Learning Representations},
  2023.

\bibitem{wang2026rethinking}
Z.~Wang, W.~Yu, L.~Yang, and S.~Lin, ``Rethinking continual learning with
  progressive neural collapse,'' in {\em International Conference on Learning
  Representations}, 2026.

\bibitem{parmar2019action}
P.~Parmar and B.~Morris, ``Action quality assessment across multiple actions,''
  in {\em IEEE Winter Conference on Applications of Computer Vision},
  pp.~1468--1476, 2019.

\bibitem{xu2022finediving}
J.~Xu, Y.~Rao, X.~Yu, G.~Chen, J.~Zhou, and J.~Lu, ``Finediving: A fine-grained
  dataset for procedure-aware action quality assessment,'' in {\em IEEE/CVF
  Conference on Computer Vision and Pattern Recognition}, pp.~2949--2958, 2022.

\bibitem{he2021towards}
J.~He, C.~Zhou, X.~Ma, T.~Berg-Kirkpatrick, and G.~Neubig, ``Towards a unified
  view of parameter-efficient transfer learning,'' in {\em International
  Conference on Learning Representations}, 2022.

\bibitem{datta2025deep}
J.~Datta, R.~Rabbi, P.~Saha, A.~N. Zereen, M.~Abdullah-Al-Wadud, and J.~Uddin,
  ``Deep representation learning using layer-wise vicreg losses: J. datta et
  al.,'' {\em Scientific Reports}, vol.~15, no.~1, p.~27049, 2025.

\bibitem{gomez2024exemplar}
A.~Gomez-Villa, D.~Goswami, K.~Wang, A.~D. Bagdanov, B.~Twardowski, and J.~Van
  De~Weijer, ``Exemplar-free continual representation learning via learnable
  drift compensation,'' in {\em European Conference on Computer Vision},
  vol.~15065, pp.~473--490, Springer, 2024.

\bibitem{bao2023pcfgaze}
Y.~Bao and F.~Lu, ``Pcfgaze: Physics-consistent feature for appearance-based
  gaze estimation,'' {\em arXiv preprint arXiv:2309.02165}, 2023.

\bibitem{parmar2017learning}
P.~Parmar and B.~Tran~Morris, ``Learning to score olympic events,'' in {\em
  IEEE Conference on Computer Vision and Pattern Recognition Workshops},
  pp.~20--28, 2017.

\bibitem{zhang2023few}
J.~Zhang, L.~Liu, O.~Silven, M.~Pietik{\"a}inen, and D.~Hu, ``Few-shot
  class-incremental learning: A survey,'' {\em arXiv preprint
  arXiv:2308.06764}, 2023.

\bibitem{xu2025comprehensive}
S.~Xu, P.~Chen, Y.~Liu, M.~Wang, S.~Wang, H.~Yan, and S.~Kwong, ``Comprehensive
  action quality assessment through multi-branch modeling,'' {\em IEEE
  Transactions on Multimedia}, vol.~27, pp.~8776--8789, 2025.

\bibitem{tang2020uncertainty}
Y.~Tang, Z.~Ni, J.~Zhou, D.~Zhang, J.~Lu, Y.~Wu, and J.~Zhou,
  ``Uncertainty-aware score distribution learning for action quality
  assessment,'' in {\em IEEE/CVF Conference on Computer Vision and Pattern
  Recognition}, pp.~9839--9848, 2020.

\bibitem{zhang2024auto}
B.~Zhang, J.~Chen, Y.~Xu, H.~Zhang, X.~Yang, and X.~Geng, ``Auto-encoding score
  distribution regression for action quality assessment,'' {\em Neural
  Computing and Applications}, vol.~36, no.~2, pp.~929--942, 2024.

\bibitem{deng2021flattening}
D.~Deng, G.~Chen, J.~Hao, Q.~Wang, and P.-A. Heng, ``Flattening sharpness for
  dynamic gradient projection memory benefits continual learning,'' {\em
  Advances in Neural Information Processing Systems}, vol.~34,
  pp.~18710--18721, 2021.

\end{thebibliography}

\begin{IEEEbiography}[{\includegraphics[width=1in,height=1.1in,clip,keepaspectratio]{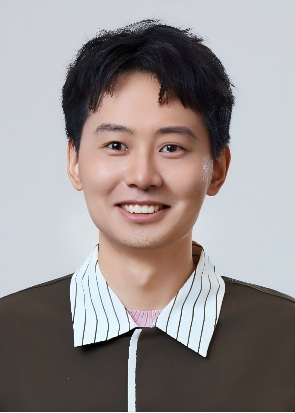}}]{Kanglei Zhou}
    is currently a Postdoctoral Researcher with the Department of Psychological and Cognitive Sciences, Tsinghua University, Beijing, China. He received his Ph.D. degree in Computer Science and Engineering from Beihang University in 2025 and his B.E. degree from Henan Normal University in 2020. In 2024, he was a Visiting Student at Durham University. His research interests include human motion analysis and continual learning.
\end{IEEEbiography}

\vspace{-3em}
\begin{IEEEbiography}[{\includegraphics[width=1in,height=1.1in,clip,keepaspectratio]{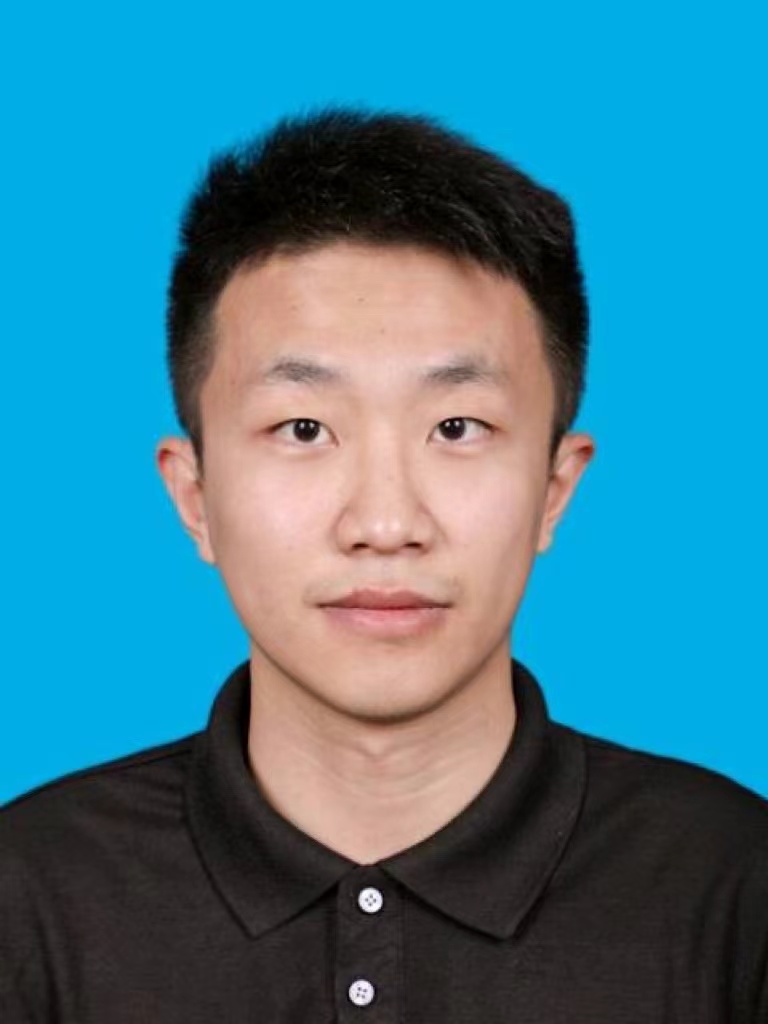}}]{Qingyi Pan}
is currently pursuing his Ph.D. degree at the Department of Statistics and Data  Science at Tsinghua University. He received his M.S. degree from the Department of Computer Science and Technology at Tsinghua University in 2022. His research interest focuses on continual learning, interpretability, and uncertainty quantification for multivariate time series forecasting.
\end{IEEEbiography}
\vspace{-3em}
\begin{IEEEbiography}[{\includegraphics[width=1in,height=1.1in,clip,keepaspectratio]{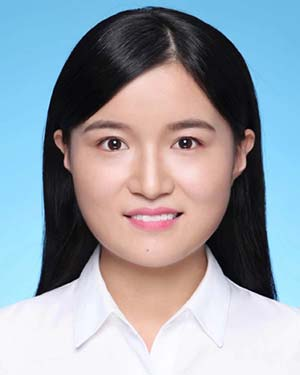}}]{Xingxing Zhang} received the BE and PhD degrees from the Institute of Information Science, Beijing Jiaotong University. She was also a visiting student with the Department of Computer Science, University of Rochester. She was a postdoc with the Department of Computer Science and Technology, Tsinghua University. Her research interests include continual learning and few-shot learning. She received the Excellent PhD Thesis Award from the Chinese Institute of Electronics.
\end{IEEEbiography}
\vspace{-3em}
\begin{IEEEbiography}[{\includegraphics[width=1in,height=1.1in,clip,keepaspectratio]{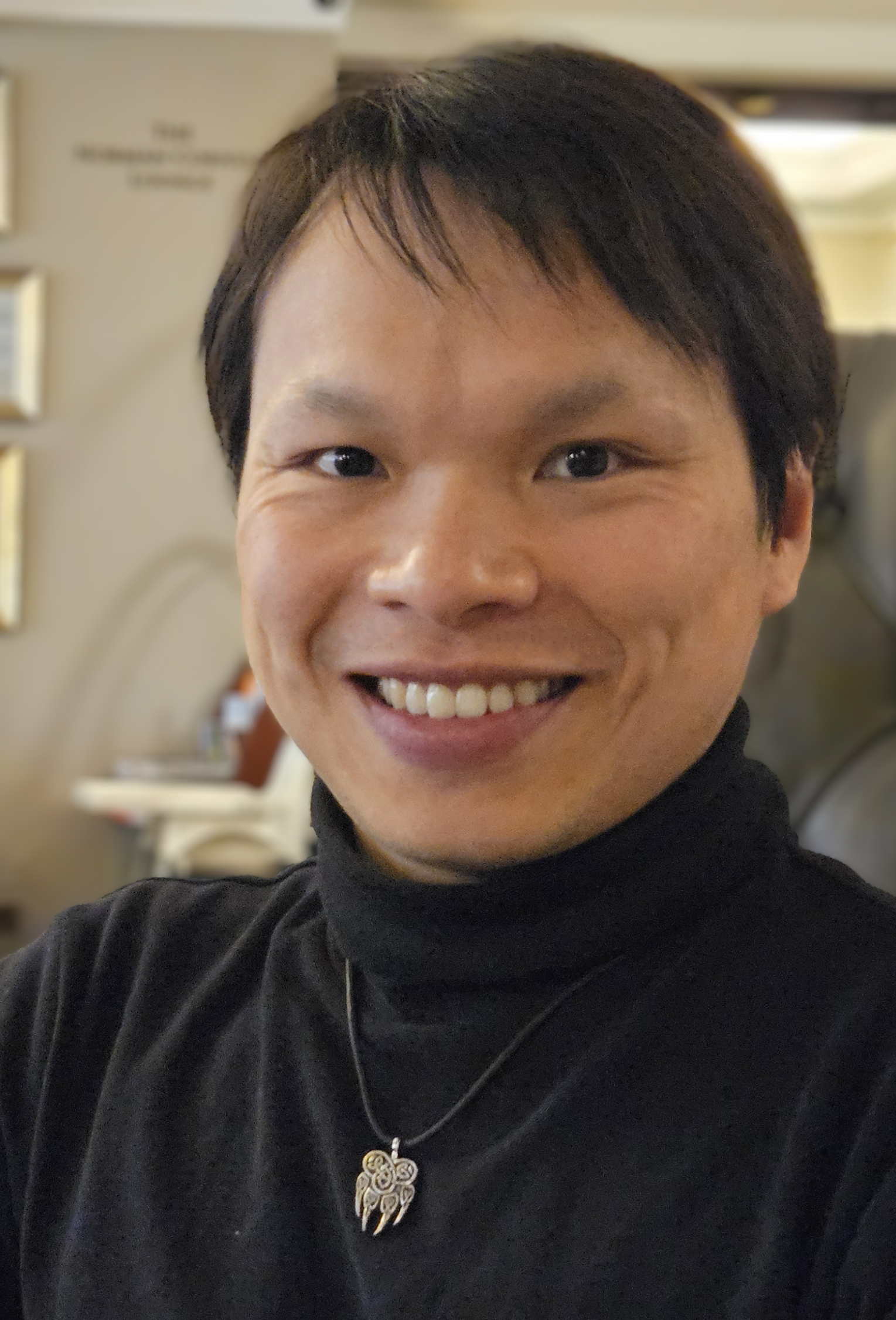}}]{Hubert P. H. Shum}
(Senior Member, IEEE) is a Professor of Visual Computing and the Director of Research for the Department of Computer Science at Durham University, specialising in modelling spatio-temporal information with responsible AI. He is also a Co-Founder and the Co-Director of Durham University Space Research Centre. He received his PhD degree from the University of Edinburgh. He chaired conferences such as Pacific Graphics, BMVC and SCA. He has authored over 200 research publications in the fields of Computer Vision, Computer Graphics and AI in Healthcare.
\end{IEEEbiography}

\vspace{-3em}
\begin{IEEEbiography}[{\includegraphics[width=1in,height=1.1in,clip,keepaspectratio]{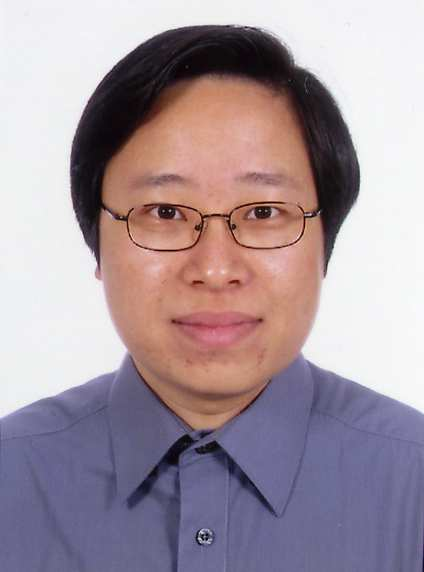}}]{Frederick W. B. Li} received a B.A. and an M.Phil. degree from Hong Kong Polytechnic University, and a Ph.D. degree from the City University of Hong Kong. He is currently an Associate Professor at Durham University, researching computer graphics, deep learning, collaborative virtual environments, and educational technologies. He is an Associate Editor of Frontiers in Education and an Editorial Board Member of Virtual Reality \& Intelligent Hardware. He chaired conferences such as ISVC and ICWL.
\end{IEEEbiography}

\vspace{-3em}
\begin{IEEEbiography}[{\includegraphics[width=1in,height=1.1in,clip,keepaspectratio]{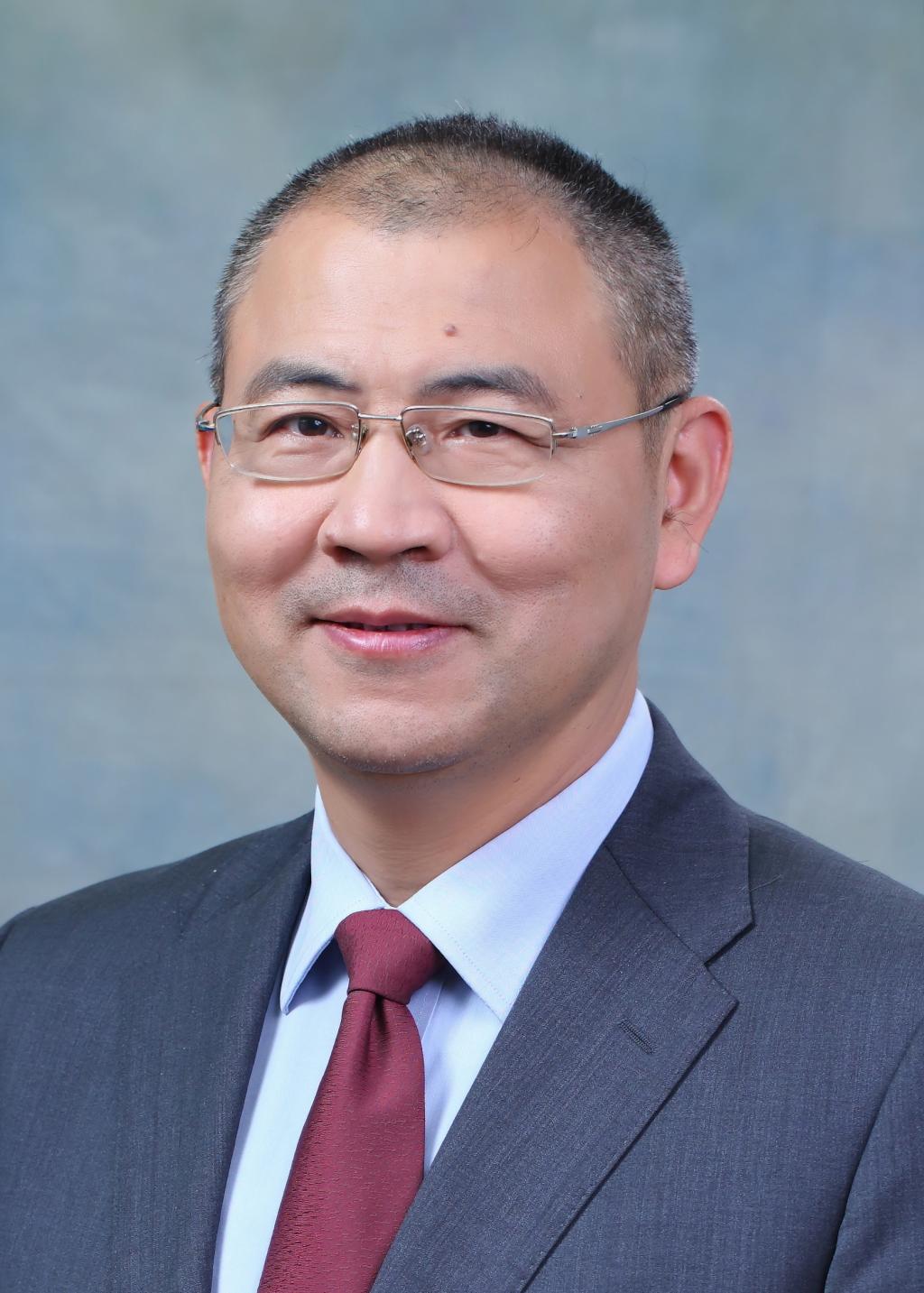}}]{Xiaohui Liang} (Member, IEEE) received his Ph.D. degree in computer science and engineering from Beihang University, China. He is currently a Professor, working in the School of Computer Science and Engineering at Beihang University. His main research interests
    include computer graphics and animation, visualization, and virtual reality.
\end{IEEEbiography}

\vspace{-3em}
\begin{IEEEbiography}[{\includegraphics[width=1in,height=1.1in,clip,keepaspectratio]{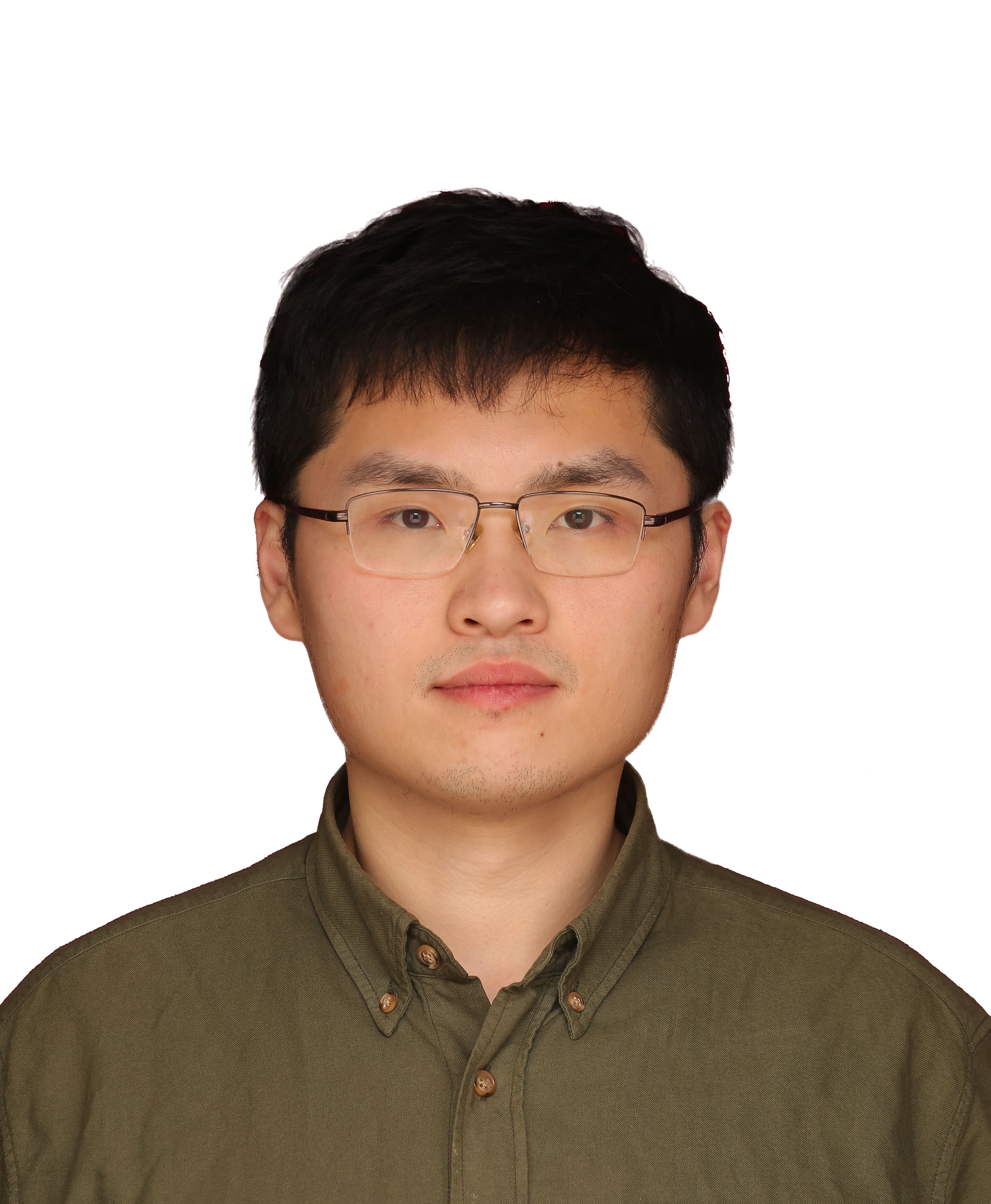}}]{Liyuan Wang} (Member, IEEE)
    is an Assistant Professor with the Department of Psychological and Cognitive Sciences, Tsinghua University, Beijing, China. He received the B.S. and Ph.D. degrees from Tsinghua University, where he also conducted his postdoctoral research. His work on continual learning has been published at major conferences and in leading journals in related fields, such as Nature Machine Intelligence, Nature Communications, TPAMI, NeurIPS, ICLR, ICML, etc.
\end{IEEEbiography} \clearpage

\appendices
\renewcommand{\thesection}{Appendix \Alph{section}}
\renewcommand{\thefigure}{A\arabic{figure}}
\renewcommand{\thetable}{A\arabic{table}}
\renewcommand{\theequation}{A\arabic{equation}}

\onecolumn

\crefalias{section}{appendix}

\section{Proof of Theorem~\ref{thm:peft-vs-fpft}}
\label{sec:peft-vs-fpft_proof}

\myPara{Notation Alignment.}
All symbols follow the definitions in \cref{sec:theoretical_foundation}.
In particular, $\bm{\theta}_{\text{up}}$, 
$\bm{\theta}_{\text{down}}^\star$, 
$\mathbf{J}_{\text{up}}(x)$, and 
$\mathbf{\Sigma}_0$ 
are defined in \cref{thm:peft-vs-fpft}.
For brevity, we denote 
$\bm{v}:=\bm{\theta}-\bm{\theta}_{\text{up}}$.
 
\begin{proof}
Let $\bm{v}:=\bm{\theta}-\bm{\theta}_{\text{up}}$ and $\mathcal{S}:=\mathrm{range}(\mathbf{U})$. 
Define the linearized risk
\begin{equation}
\label{eq:lin-risk}
\widetilde R_{\text{down}}(\bm{v})
:=\mathbb{E}_{(x,y)\sim\mathcal{D}_{\text{down}}}\,
\ell\!\big(\phi_{\bm{\theta}_{\text{up}}}(x)+\mathbf{J}_{\text{up}}(x)^\top \bm{v},\,y\big),
\end{equation}
and use the $\|\bm{w}\|_{\mathbf{\Sigma}_0}^2=\bm{w}^\top\mathbf{\Sigma}_0 \bm{w}$-norm, i.e., $\bm{w}^\top\mathbf{\Sigma}_0 \bm{w}$
where $\mathbf{\Sigma}_0$ is defined in
\cref{thm:peft-vs-fpft}.

By Taylor’s theorem, for $\bm{v}^\star:=\arg\min_{\bm{v}}\widetilde R_{\text{down}}(\bm{v})$ and any $\bm{v}$ in the NTK neighborhood, one has
\begin{equation}
\label{eq:taylor}
\widetilde R_{\text{down}}(\bm{v})
\;\ge\;\widetilde R_{\text{down}}(\bm{v}^\star)
+\tfrac12(\bm{v}-\bm{v}^\star)^\top \mathbf{H}(\bm{v})(\bm{v}-\bm{v}^\star),
\end{equation}
where
\begin{equation}
\label{eq:hessian}
\mathbf{H}(\bm{v})=\mathbb{E}_{(x,y)\sim\mathcal{D}_{\text{down}}}\!
\big[\ell''(\phi_{\bm{\theta}_{\text{up}}}(x)+\mathbf{J}_{\text{up}}(x)^\top \bar{\bm{v}},y)\,
\mathbf{J}_{\text{up}}(x)\mathbf{J}_{\text{up}}(x)^\top\big]\succeq \mu\,\mathbf{\Sigma}_0
\end{equation}
for some $\bar{\bm{v}}$ on the segment $[\bm{v}^\star,\bm{v}]$. 
This implies
\begin{equation}
\label{eq:qg}
\widetilde R_{\text{down}}(\bm{v})-\widetilde R_{\text{down}}(\bm{v}^\star)
\;\ge\;\tfrac{\mu}{2}\,\|\bm{v}-\bm{v}^\star\|_{\mathbf{\Sigma}_0}^2.
\end{equation}

Since PEFT restricts $\bm{v}$ to $\mathcal{S}$, let $\bm{v}_{\mathcal{S}}:=\arg\min_{\bm{v}\in\mathcal{S}}\widetilde R_{\text{down}}(\bm{v})$. 
By \cref{eq:qg}, it follows that
\begin{equation}
\label{eq:vs-gap}
\widetilde R_{\text{down}}(\bm{v}_{\mathcal{S}})-\widetilde R_{\text{down}}(\bm{v}^\star)
\;\ge\;\tfrac{\mu}{2}\,\inf_{\bm{s}\in\mathcal{S}}\|\bm{s}-\bm{v}^\star\|_{\mathbf{\Sigma}_0}^2
=\tfrac{\mu}{2}\,\big\|\bm{v}^\star-\Pi_{\mathcal{S}}^{(\mathbf{\Sigma}_0)}(\bm{v}^\star)\big\|_{\mathbf{\Sigma}_0}^2,
\end{equation}
where $\Pi_{\mathcal{S}}^{(\mathbf{\Sigma}_0)}$ is the $\mathbf{\Sigma}_0$-orthogonal projection onto $\mathcal{S}$. 
Substituting $\bm{v}^\star=\bm{\theta}_{\text{down}}^\star-\bm{\theta}_{\text{up}}$ gives
\begin{equation}
\label{eq:orth-decomp}
\big\|\bm{v}^\star-\Pi_{\mathcal{S}}^{(\mathbf{\Sigma}_0)}(\bm{v}^\star)\big\|_{\mathbf{\Sigma}_0}
=\big\|\Pi_{\mathcal{S}^\perp}^{(\mathbf{\Sigma}_0)}(\bm{\theta}_{\text{down}}^\star-\bm{\theta}_{\text{up}})\big\|_{\mathbf{\Sigma}_0}.
\end{equation}

Finally, by assumption \textbf{(A3)}, for $\|\bm{v}_{\mathcal{S}}\|,\|\bm{v}^\star\|\le\rho$,
\begin{equation}
\label{eq:final-step}
R_{\text{down}}(\bm{\theta}_{\text{up}}+\bm{v}_{\mathcal{S}})
- R_{\text{down}}(\bm{\theta}_{\text{down}}^\star)
\;\ge\;\widetilde R_{\text{down}}(\bm{v}_{\mathcal{S}})-\widetilde R_{\text{down}}(\bm{v}^\star)
- \tfrac{L_\varepsilon}{2}\big(\|\bm{v}_{\mathcal{S}}\|^2+\|\bm{v}^\star\|^2\big).
\end{equation}
Combining \cref{eq:vs-gap,eq:orth-decomp,eq:final-step} yields
\begin{equation}
\label{eq:proof-final}
R_{\text{down}}(\bm{\theta}_{\text{up}}+\bm{v}_{\mathcal{S}})
- R_{\text{down}}(\bm{\theta}_{\text{down}}^\star)
\;\ge\;\tfrac{\mu}{2}\,\big\|\Pi_{\mathcal{S}^\perp}^{(\mathbf{\Sigma}_0)}(\bm{\theta}_{\text{down}}^\star-\bm{\theta}_{\text{up}})\big\|_{\mathbf{\Sigma}_0}^2
- C_\varepsilon,
\end{equation}
where $C_\varepsilon:=\tfrac{L_\varepsilon}{2}(\|\bm{v}_{\mathcal{S}}\|^2+\|\bm{v}^\star\|^2)$. 
This establishes \cref{eq:excess-risk-peft-sigma}.
\end{proof}

\begin{algorithm}[!h]
\small
\renewcommand{\AlCapFnt}{\small}
\renewcommand{\AlCapNameFnt}{\small}

\caption{\small Training procedure of MAGR++ for CAQA}
\label{alg:magrpp}
\SetKwInOut{Input}{Input}\SetKwInOut{Output}{Output}
\Input{Sequential training sets $\{\mathcal{D}^t_{\text{train}}\}_{t=1}^{T}$, memory size $M$, backbone $f$, regressor $g$, projector $p$, threshold $\epsilon$}
\Output{Trained $(f^T,g^T,p^T)$}
\SetKwFunction{OUS}{OUS}
\SetKwFunction{LayerFPFT}{LayerSelection}
\SetKwFunction{ProjLearn}{TrainProjector}
\SetKwFunction{Rectify}{RectifyFeatures}
\SetKwFunction{IILoss}{IIJGRLoss}
\SetKwFunction{Refresh}{RefreshMemory}

\BlankLine
\textbf{Init:} Pretrain $f$ (e.g., I3D) and init $g,p$; $\mathcal{M}\leftarrow\varnothing$\;
\For{$t \leftarrow 1$ \KwTo $T$}{
  
  Copy previous backbone $f^{t-1}$ and freeze $f^{t-1}$ \tcp*{copy previous backbone}
  $l_{\text{opt}} \leftarrow$ \LayerFPFT(\,$f^{t-1},\ f^t, \mathcal{D}^1_{\text{train}},\ \epsilon$\,)\tcp*{see \cref{sec:layer_adaptive}, invoke \cref{alg:layer_selection}}

  \tcp{Mini-batch training loop}
  \While{not converged}{
    Sample current batch $\mathcal{B}^t = \{(\mathbf{x}_i^t,y_i^t)\}_{i=1}^{b_2} \subset \mathcal{D}^t_{\text{train}}$\;
    \tcp{Forward through current and previous backbones}
    $\bm{h}_i^{t,l} \leftarrow f^{t,l}(\mathbf{x}_i^t)$,\quad
    $\bm{z}_i^{t,l} \leftarrow f^{t-1,l}(\mathbf{x}_i^t)$ for all layers $l$\;

    $\hat{y}_i^t \leftarrow g(\bm{h}_i^{t,L})$;\quad
    $\mathcal{L}_{\text{D}} \leftarrow \frac{1}{b_2}\sum_i (\hat{y}_i^t - y_i^t)^2$\tcp*{Current-task loss (Eq.~\eqref{eq:caqa})}

    \tcp{Training phase 1: layer-adaptive FPFT  (Eq.~\eqref{eq:la})}
    $\mathcal{L}_{\text{tune}} \leftarrow \frac{1}{b_2}\sum_i \sum_{l<l_{\text{opt}}} \|\bm{h}_i^{t,l} - \bm{z}_i^{t,l}\|_2^2$\;

    \tcp{Training phase 2: projector learning (Eqs.~\eqref{eq_mp_learning}, \eqref{eq:proj})}
    $\bm{z}_i^{t} \leftarrow \bm{z}_i^{t,L}$; \quad $\bm{\hat{h}}_i^{t} \leftarrow \bm{z}_i^{t} + p(\bm{z}_i^{t})$;\quad
    $\mathcal{L}_{\text{proj}} \leftarrow \frac{1}{b_2}\sum_i \|\bm{h}_i^{t,L} - \bm{\hat{h}}_i^{t}\|_2^2$\;

    Sample old feature batch $\tilde{\mathcal{B}}=\{(\tilde{\bm{h}}_j,y_j)\}_{j=1}^{b_1}\subset \mathcal{M}$\;
    \ForEach{$(\tilde{\bm{h}}_j,y_j)\in \tilde{\mathcal{B}}$}{
      $\tilde{\bm{h}}_j \leftarrow \tilde{\bm{h}}_j + p(\tilde{\bm{h}}_j)$ \tcp*{deviated feature translation}
    }

    \tcp{Training phase 3: build joint batch and compute regularizer (Eq.~\eqref{eq:iij})}
    $\mathbf{H} \leftarrow [\tilde{\bm{h}}_{1:b_1},\ \bm{h}^{t,L}_{1:b_2}]$;\quad
    $\bm{y} \leftarrow [y_{1:b_1},\ y^t_{1:b_2}]$\;
    $\mathcal{L}_{\text{reg}} \leftarrow$ \IILoss(\,$\mathbf{H},\bm{y}$\,)\tcp*{angular distances + matrix partitions, see \cref{eq:iij}}

    $\hat{y}_j \leftarrow g(\tilde{\bm{h}}_j)$;\quad
    $\mathcal{L}_{\text{M}} \leftarrow \frac{1}{b_1}\sum_j (\hat{y}_j - y_j)^2$\tcp*{replay loss (Eq.~\eqref{eq:caqa})}

    Update $\{f,g,p\}$ by backprop on $\mathcal{L}$ (optimizer, LR schedule, etc.)\;
  }

  \tcp{End-of-session memory maintenance}
  \ForEach{$(\tilde{\bm{h}},y)\in \mathcal{M}$}{ 
    $\tilde{\bm{h}} \leftarrow \tilde{\bm{h}} + p(\tilde{\bm{h}})$ \tcp*{refresh old features via converged projector}
  }
  $\mathcal{P}^t \leftarrow \OUS(\mathcal{D}^t_{\text{train}}, f, g, M)$ \tcp*{select new prototypes, invoke \cref{alg:ous}}
  Update memory as $\mathcal{M} \leftarrow \mathcal{M} \cup \mathcal{P}^t$ and keep at most $M$ items\;
}
\end{algorithm}

\section{Proof of Theorem~\ref{thm:fpft-replay}}
\label{sec:proof-fpft-replay}

\myPara{Notation Alignment.}
All symbols follow the definitions in
\cref{sec:theoretical_foundation},
in particular the parameter sequence
$\{\bm{\theta}^t\}$ and update magnitude
$\Delta_t$.
 
\begin{proof}
Let the per-session update be defined as
\begin{equation}
\label{eq:update-def}
\bm{\Delta}_t := \bm{\theta}^{t}-\bm{\theta}^{t-1}, 
\qquad 
\Delta_t := \|\bm{\Delta}_t\|.
\end{equation}
Define the per-task risk and its empirical counterpart as
\begin{equation}
\label{eq:task-risk}
R_k(\bm{\theta})
:=\mathbb{E}_{(x,y)\sim\mathcal{D}_k}\,\ell\!\big(\phi_{\bm{\theta}}(x),y\big),
\qquad
\hat R_k(\bm{\theta})
:=\mathbb{E}_{(x,y)\sim\text{mem}(k)}\,\ell\!\big(\phi_{\bm{\theta}}(x),y\big).
\end{equation}
At session $t$, define the stale and ideal replay objectives:
\begin{equation}
\label{eq:stale-ideal}
\begin{aligned}
\mathcal{L}_t^{\text{stale}}(\bm{\theta})
&:=\mathbb{E}_{(x,y)\sim \mathcal{M}_{t-1}}\,
\ell\!\big(g_{\bm{\theta}_g}(f_{\bm{\theta}_f^{t-1}}(x)),y\big),\\
\mathcal{L}_t^{\text{ideal}}(\bm{\theta})
&:=\mathbb{E}_{(x,y)\sim \mathcal{M}_{t-1}}\,
\ell\!\big(g_{\bm{\theta}_g}(f_{\bm{\theta}_f}(x)),y\big).
\end{aligned}
\end{equation}

By Assumptions~\textbf{(B1)} and~\textbf{(B2)}, evaluating at $\bm{\theta}=\bm{\theta}^t$ yields
\begin{equation}
\label{eq:l-diff}
\begin{aligned}
\Big|\mathcal{L}_t^{\text{stale}}(\bm{\theta}^t)-\mathcal{L}_t^{\text{ideal}}(\bm{\theta}^t)\Big|
&\le 
\mathbb{E}\,\big\|g_{\bm{\theta}_g^{t}}\!\big(f_{\bm{\theta}_f^{t-1}}(x)\big)
- g_{\bm{\theta}_g^{t}}\!\big(f_{\bm{\theta}_f^{t}}(x)\big)\big\| \\
&\le L_g\,\mathbb{E}\,\big\|f_{\bm{\theta}_f^{t-1}}(x)-f_{\bm{\theta}_f^{t}}(x)\big\|.
\end{aligned}
\end{equation}
Using Assumption~\textbf{(B3)} with $\|\bm{\theta}_f^{t}-\bm{\theta}_f^{t-1}\|=\Delta_t$, we have
\begin{equation}
\label{eq:drift}
\Big|\mathcal{L}_t^{\text{stale}}(\bm{\theta}^t)-\mathcal{L}_t^{\text{ideal}}(\bm{\theta}^t)\Big|
\;\le\; L_g\,L_f\,\Delta_t.
\end{equation}
If there exists a projector $\mathbf{P}_t$ such that
\begin{equation}
\label{eq:app-proj}
\mathbb{E}\,\big\|f_{\bm{\theta}_f^{t}}(x)-\mathbf{P}_t f_{\bm{\theta}_f^{t-1}}(x)\big\|\le \varepsilon_t,
\end{equation}
the bound refines to
\begin{equation}
\label{eq:drift-proj}
\Big|\mathcal{L}_t^{\text{stale}}(\bm{\theta}^t)-\mathcal{L}_t^{\text{ideal}}(\bm{\theta}^t)\Big|
\;\le\; L_g\,\varepsilon_t.
\end{equation}

Let $\mathcal{H}_{t-1}$ denote the model class realizable near $\bm{\theta}^{t-1}$, and define
\begin{equation}
\label{eq:hclass}
\mathcal{H}_t := \mathcal{H}_{t-1} \cup 
\big\{\phi_{\bm{\theta}}:\|\bm{\theta}-\bm{\theta}^{t-1}\|\le \Delta_t\big\}.
\end{equation}
By Assumption~\textbf{(B4)}, enlarging the parameter ball by $\Delta_t$ increases localized complexity (e.g., Rademacher or covering bounds) by at most $C_0 L_\phi \Delta_t$, giving
\begin{equation}
\label{eq:gen}
\mathbb{E}\big[R_k(\bm{\theta}^t)-\hat R_k(\bm{\theta}^t)\big]
\;\le\;
\mathbb{E}\big[R_k(\bm{\theta}^{t-1})-\hat R_k(\bm{\theta}^{t-1})\big]
\;+\; C\,L_\phi\,\Delta_t,
\end{equation}
for some constant $C>0$.

Define forgetting as
\begin{equation}
\label{eq:forget-def}
\psi_t(k) := R_k(\bm{\theta}^t) - R_k(\bm{\theta}^{t-1}).
\end{equation}
Adding and subtracting empirical risks gives
\begin{equation}
\label{eq:three-terms}
\psi_t(k)
=\underbrace{R_k(\bm{\theta}^t)-\hat R_k(\bm{\theta}^t)}_{\text{(I)}}
+\underbrace{\hat R_k(\bm{\theta}^t)-\hat R_k(\bm{\theta}^{t-1})}_{\text{(II)}}
+\underbrace{\hat R_k(\bm{\theta}^{t-1})-R_k(\bm{\theta}^{t-1})}_{\text{(III)}}.
\end{equation}
Taking expectations and applying \cref{eq:gen} to (I) and (III) gives
\begin{equation}
\label{eq:gen-contrib}
\mathbb{E}[\text{(I)}+\text{(III)}]\;\lesssim\; C\,L_\phi\,\Delta_t.
\end{equation}
For (II), the empirical change is bounded by the stale–ideal gap plus the session-$t$ optimization suboptimality:
\begin{equation}
\label{eq:emp-diff}
\mathbb{E}\big[\hat R_k(\bm{\theta}^t)-\hat R_k(\bm{\theta}^{t-1})\big]
\;\lesssim\;
\Big|\mathcal{L}_t^{\text{stale}}(\bm{\theta}^t)-\mathcal{L}_t^{\text{ideal}}(\bm{\theta}^t)\Big|
\;+\; E_\text{opt}.
\end{equation}
Combining \cref{eq:drift,eq:gen-contrib,eq:emp-diff} yields
\begin{equation}
\label{eq:forgetting-final}
\mathbb{E}[\psi_t(k)]
\;\lesssim\;
L_g L_f\,\Delta_t \;+\; C\,L_\phi\,\Delta_t \;+\; E_\text{opt},
\end{equation}
matching the bound in \cref{eq:forgetting-bound}.  
In the projector case, replace $L_g L_f \Delta_t$ by $L_g \varepsilon_t$ from \cref{eq:drift-proj}, completing the proof.
\end{proof}

\section{Training Procedure} \label{sec:training}
At each session $t$, MAGR++ jointly optimizes the backbone, regressor, and projector under the composite objective in \cref{eq:magrpp-obj}. The training begins with Ordered Uniform Sampling (OUS) to store representative features from session $t{--}1$ in the memory bank $\mathcal{M}$, ensuring efficient replay coverage as required by the CAQA loss in \cref{eq:caqa}. The backbone $f^t$ is then adapted to current data $\mathcal{D}^t_{\text{train}}$ via layer-adaptive FPFT, where shallow layers below $l_{\text{opt}}$ are constrained by the feature-matching loss in \cref{eq:la}, while deeper layers are fully fine-tuned to capture evolving quality cues. To handle manifold shift, the Manifold Projector (MP) is trained using the projection loss in \cref{eq:proj}, aligning $f^{t-1}$ and $f^t$ representations with only current-session inputs, and subsequently applied to translate old features from $\mathcal{M}$ into the updated space for replay. In parallel, the IIJ-GR minimizes the regularization loss in \cref{eq:iij}, enforcing both intra- and inter-session consistency between feature geometry and quality scores. Finally, the regressor $g^t$ is optimized on a mixture of rectified old features and current-session features, and the memory bank is refreshed by updating old features through MP and adding new prototypes. This coordinated pipeline ensures that MAGR++ balances adaptation and stability across sessions, while mitigating forgetting through replay.
The details of the training procedure are shown in \cref{alg:magrpp}.

\begin{figure*}[!h]
\centering
\begin{minipage}[t]{0.48\linewidth}
\begin{algorithm}[H]
\small
\renewcommand{\AlCapFnt}{\small}
\renewcommand{\AlCapNameFnt}{\small}
\caption{\small \texttt{LayerSelection}: Layer Selection}
\label{alg:layer_selection}
\KwIn{Base-session set $\mathcal{D}^0$, backbone $f$, threshold $\epsilon$}
\KwOut{Optimal boundary $l_{\text{opt}}$}

\For{$l \leftarrow 1$ \KwTo $L$}{
  $\mathbf{Z}^l_{\text{fix}} \leftarrow f_{\text{fix}}^l(\mathcal{D}^0)$; \quad
  $\mathbf{Z}^l_{\text{tune}} \leftarrow f_{\text{tune}}^l(\mathcal{D}^0)$\;

  $r^l \leftarrow 
  \mathcal{C}(\mathbf{Z}^l_{\text{tune}}) / \mathcal{C}(\mathbf{Z}^l_{\text{fix}})$\;

  \If{$r^l > 1 + \epsilon$}{
     $l_{\text{opt}} \leftarrow l$\;
  }
}
\textbf{return}~{$l_{\text{opt}}$}
\end{algorithm}
\end{minipage}
\hfill
\begin{minipage}[t]{0.48\linewidth}
\begin{algorithm}[H]
\small
\renewcommand{\AlCapFnt}{\small}
\renewcommand{\AlCapNameFnt}{\small}
\caption{\small  \texttt{OUS}: Ordered Uniform Sampling}
\label{alg:ous}
\KwIn{Training set $\mathcal{D}^t$, memory size $M$, scorer $g$}
\KwOut{Prototype set $\mathcal{P}^t$}

Compute scores $y_i = g^t(f^t(\mathbf{x}_i))$ for all $\mathbf{x}_i \in \mathcal{D}^t$\;
Sort $\mathcal{D}^t$ by $y_i$ in ascending order\;
Divide sorted samples into $M$ intervals uniformly across score range\;
Select one representative sample from each interval\;
\KwRet $\mathcal{P}^t$
\end{algorithm}
\end{minipage}
\end{figure*}

\begin{table}
    \centering
    \MStabcaption{tab:vault}{Comparison results on UNLV-Vault and JIGSAWS. The primary metric is $\rho_{\text{avg}}$, with Sequential Fine-Tuning (SFT) as the Lower Bound (LB). Best results are in bold, excluding UB and LB. $\uparrow$/$\downarrow$ indicate higher/lower is better.}{}
\begin{minipage}{0.50\linewidth}
\centering
\subcaption{UNLV-Vault (offline)}
    \vspace{-1.5mm}
    \resizebox{\linewidth}{!}{
    \begin{tabular}{l r c S[table-format=+1.4, input-symbols={\textbf}]  c S[table-format=+1.4, input-symbols={\textbf}] c}
        \toprule
        \textbf{Method} & \textbf{Publisher} & \textbf{Memory} & \multicolumn{1}{c}{$\rho_{\text{avg}}$ ($\uparrow$)} & $\rho_{\text{aft}}$ ($\downarrow$) & \multicolumn{1}{c}{$\rho_{\text{fwt}}$ ($\uparrow$)} & RL2 ($\downarrow$) \\
        \midrule
        \rowcolor{rowgrayblue}JT (UB) & -- & None & 0.7514 & -- & {--} & 0.0223 \\
        \rowcolor{rowgrayblue}SFT (LB) & -- & None & 0.5168 & 0.1887 & 0.3445 & 0.0764 \\
        \rowcolor{rowgraygreen}SI \cite{zenke2017continual} & ICML'17 & None & 0.5165 & 0.2287 & 0.2839 & 0.0748 \\
        \rowcolor{rowgraygreen}EWC \cite{james2017ewc} & PNAS'17 & None & 0.5173 & 0.2250 & 0.2371 & 0.0739 \\
        \rowcolor{rowgraygreen}LwF \cite{li2017learning} & TPAMI'17 & None & 0.6659 & 0.1371 & 0.4318 & 0.0346 \\
        \rowcolor{rowgrayyellow}MER \cite{riemer2019learning} & ICLR'19 & Raw Data & 0.5883 & 0.1458 & 0.4053 & 0.0568 \\
        \rowcolor{rowgrayyellow}DER++ \cite{buzzega2020dark} & NeurIPS'20 & Raw Data & 0.5905 & 0.3693 & 0.2150 & 0.0615 \\
        \rowcolor{rowgrayyellow}TOPIC \cite{tao2020few} & CVPR'20 & Raw Data & 0.5429 & 0.1712 & 0.3183 & 0.0674 \\
        \rowcolor{rowgrayyellow}GEM \cite{kukleva2021generalized} & ICCV'21 & Raw Data & 0.5339 & 0.1854 & 0.0608 & 0.0701 \\
        \rowcolor{rowgrayyellow}STAR \cite{eskandar2025star} & ICLR'25 & Raw Data & 0.6305 & \bf 0.0472 & -0.0406 & 0.0408 \\
        \rowcolor{rowgraypink}Feature MER & -- & Feature & 0.4342 & 0.1856 & 0.4578 & 0.0892 \\
        \rowcolor{rowgraypink}SLCA \cite{zhang2023slca} & ICCV'23 & Feature & 0.4919 & 0.1221 & 0.3972 & 0.0785 \\
        \rowcolor{rowgraypink}NC-FSCIL \cite{yang2023neural} & ICLR'23 & Feature & 0.5747 & 0.2664 & \bf 0.4863 & 0.0581 \\
        \rowcolor{rowgraypink}FS-Aug \cite{li2024continual} & TCSVT'24 & Feature & 0.5146 & 0.2113 & 0.4275 & 0.0728 \\
        \rowcolor{rowgraypink}MAGR \cite{zhou2024magr} & ECCV'24 & Feature & 0.6526 & 0.0687 & 0.2853 & 0.0362 \\
        \rowcolor{rowgraypink}ProNC \cite{wang2026rethinking} & ICLR'26 & Feature & 0.6474 & 0.1901 & -0.0233 & 0.0317 \\
        \rowcolor{rowgraypink}MAGR++ (Ours) & -- & Feature & \bf 0.7012 & 0.2425 & 0.2534 & \bf 0.0278 \\
        \bottomrule
    \end{tabular}
    }
\end{minipage}\hfill \begin{minipage}{0.50\linewidth}
\centering
\subcaption{JIGSAWS (offline)}
    \vspace{-1.5mm}
    \resizebox{\linewidth}{!}{
    \begin{tabular}{l r c S[table-format=+1.4, input-symbols={\textbf}]  c S[table-format=+1.4, input-symbols={\textbf}] c}
        \toprule
        \textbf{Method} & \textbf{Publisher} & \textbf{Memory} & \multicolumn{1}{c}{$\rho_{\text{avg}}$ ($\uparrow$)} & $\rho_{\text{aft}}$ ($\downarrow$) & \multicolumn{1}{c}{$\rho_{\text{fwt}}$ ($\uparrow$)} & RL2 ($\downarrow$) \\
        \midrule
        \rowcolor{rowgrayblue}JT (UB) & -- & None & 0.8452 & -- & {--} & 0.0409 \\
        \rowcolor{rowgrayblue}SFT (LB) & -- & None & 0.7694 & 0.0583 & 0.5826 & 0.0432 \\
        \rowcolor{rowgraygreen}SI \cite{zenke2017continual} & ICML'17 & None & 0.7838 & 0.0223 & 0.5908 & 0.0458 \\
        \rowcolor{rowgraygreen}EWC \cite{james2017ewc} & PNAS'17 & None & 0.7640 & 0.0930 & 0.2662 & 0.0462 \\
        \rowcolor{rowgraygreen}LwF \cite{li2017learning} & TPAMI'17 & None & 0.7611 & 0.0404 & 0.5416 & 0.0439 \\
        \rowcolor{rowgrayyellow}MER \cite{riemer2019learning} & ICLR'19 & Raw Data & 0.7977 & 0.0620 & 0.6483 & 0.0357 \\
        \rowcolor{rowgrayyellow}DER++ \cite{buzzega2020dark} & NeurIPS'20 & Raw Data & 0.5935 & 0.1175 & 0.2298 & 0.1254 \\
        \rowcolor{rowgrayyellow}TOPIC \cite{tao2020few} & CVPR'20 & Raw Data & 0.6880 & 0.1506 & 0.4595 & 0.0541 \\
        \rowcolor{rowgrayyellow}GEM \cite{kukleva2021generalized} & ICCV'21 & Raw Data & 0.7815 & 0.0384 & 0.5662 & 0.0418 \\
        \rowcolor{rowgrayyellow}STAR \cite{eskandar2025star} & ICLR'25 & Raw Data & 0.7778 & 0.1997 & 0.6155 & \bf 0.0322  \\
        \rowcolor{rowgraypink}Feature MER & -- & Feature & 0.7776 & 0.0895 & 0.5908 & 0.0452 \\
        \rowcolor{rowgraypink}SLCA \cite{zhang2023slca} & ICCV'23 & Feature & 0.7994 & 0.2316 & 0.6319 & 0.0350 \\
        \rowcolor{rowgraypink}NC-FSCIL \cite{yang2023neural} & ICLR'23 & Feature & 0.7431 & 0.0960 & 0.6483 & 0.0445 \\
        \rowcolor{rowgraypink}FS-Aug \cite{li2024continual} & TCSVT'24 & Feature & 0.7270 & 0.0452 & 0.6647 & 0.0648 \\
        \rowcolor{rowgraypink}MAGR \cite{zhou2024magr} & ECCV'24 & Feature & 0.7999 & 0.0197 &  0.7385 & 0.0349 \\
        \rowcolor{rowgraypink}ProNC \cite{wang2026rethinking} & ICLR'26 & Feature & 0.8012 & 0.0583 & 0.6811 & 0.0401 \\
        \rowcolor{rowgraypink}MAGR++ (Ours) & -- & Feature & \bf 0.8238 &  \bf 0.0192 & \bf 0.7549 & 0.0358 \\
        \bottomrule
    \end{tabular}
    }
\end{minipage}

\begin{minipage}{0.50\linewidth}
    \centering
    \vspace{2mm}
    \subcaption{UNLV-Vault (online)}
    \vspace{-1.5mm}
    \resizebox{\linewidth}{!}{
    \begin{tabular}{l r c S[table-format=+1.4, input-symbols={\textbf}] c S[table-format=+1.4, input-symbols={\textbf}] c}
    \toprule
    \textbf{Method} & \textbf{Publisher} & \textbf{Memory} & \multicolumn{1}{c}{$\rho_{\text{avg}}$ ($\uparrow$)} & $\rho_{\text{aft}}$ ($\downarrow$) & \multicolumn{1}{c}{$\rho_{\text{fwt}}$ ($\uparrow$)} & RL2 ($\downarrow$) \\
    \midrule
    \rowcolor{rowgrayblue}SFT (LB) & -- & None & 0.2139 & 0.0684 & 0.5507 & 0.1891 \\
    \rowcolor{rowgraygreen}SI \cite{zenke2017continual} & ICML'17 & None & -0.2904 & 0.0897 & 0.2912 & 0.4528 \\
    \rowcolor{rowgraygreen}EWC \cite{james2017ewc} & PNAS'17 & None & 0.0585 & 0.0385 & 0.2288 & 0.3186 \\
    \rowcolor{rowgraygreen}LwF \cite{li2017learning} & TPAMI'17 & None & -0.1075 & 0.2311 & 0.1540 & 0.3874 \\
    \rowcolor{rowgrayyellow}MER \cite{riemer2019learning} & ICLR'19 & Raw Data & 0.0441 & 0.2533 & 0.2597 & 0.3325 \\
    \rowcolor{rowgrayyellow}DER++ \cite{buzzega2020dark} & NeurIPS'20 & Raw Data & -0.1701 & 0.1642 & 0.2853 & 0.4127 \\
    \rowcolor{rowgrayyellow}TOPIC \cite{tao2020few} & CVPR'20 & Raw Data & 0.0590 & 0.1013 & 0.3340 & 0.3098 \\
    \rowcolor{rowgrayyellow}GEM \cite{kukleva2021generalized} & ICCV'21 & Raw Data & 0.0391 & 0.1013 & 0.3340 & 0.3216 \\
    \rowcolor{rowgrayyellow}STAR \cite{eskandar2025star} & ICLR'25 & Raw Data & -0.2879 & 0.1400 & -0.1179 & 0.4413 \\
    \rowcolor{rowgraypink}Feature MER & -- & Feature & 0.3571 & 0.1444 & -0.0213 & 0.2364 \\
    \rowcolor{rowgraypink}SLCA \cite{zhang2023slca} & ICCV'23 & Feature & 0.0962 & 0.1242 & 0.2670 & 0.2875 \\
    \rowcolor{rowgraypink}NC-FSCIL \cite{yang2023neural} & ICLR'23 & Feature & 0.4971 & 0.0291 & -0.0463 & 0.1214 \\
    \rowcolor{rowgraypink}FS-Aug \cite{li2024continual} & TCSVT'24 & Feature & 0.1998 & 0.1350 & 0.1497 & 0.2486 \\
    \rowcolor{rowgraypink}MAGR \cite{zhou2024magr} & ECCV'24 & Feature & 0.1986 & 0.1201 & -0.1483 & 0.2597 \\
    \rowcolor{rowgraypink}ProNC \cite{wang2026rethinking} & ICLR'26 & Feature & -0.2894 & 0.3150 & 0.0725 & 0.4876 \\
    \rowcolor{rowgraypink}MAGR++ (Ours) & -- & Feature & \bf 0.5806 & \bf 0.0000 & \bf 0.7057 & \textbf{0.0821} \\
    \bottomrule
    \end{tabular}
    }
\end{minipage}\hfill \begin{minipage}{0.50\linewidth}
    \centering
    \vspace{2mm}
    \subcaption{JIGSAWS (online)}
    \vspace{-1.5mm}
    \resizebox{\linewidth}{!}{
    \begin{tabular}{l r c S[table-format=+1.4, input-symbols={\textbf}] c S[table-format=+1.4, input-symbols={\textbf}] c}
    \toprule
            \textbf{Method} & \textbf{Publisher} & \textbf{Memory} & \multicolumn{1}{c}{$\rho_{\text{avg}}$ ($\uparrow$)} & $\rho_{\text{aft}}$ ($\downarrow$) & \multicolumn{1}{c}{$\rho_{\text{fwt}}$ ($\uparrow$)} & RL2 ($\downarrow$) \\
    \midrule
\rowcolor{rowgrayblue}SFT (LB) & -- & None & 0.0003 & 0.5085 & -0.1313 & 0.4289 \\
    \rowcolor{rowgraygreen}SI \cite{zenke2017continual} & ICML'17 & None & -0.2904 & 0.0897 & 0.2912 & 0.1840 \\
    \rowcolor{rowgraygreen}EWC \cite{james2017ewc} & PNAS'17 & None & 0.1170 & 0.5048 & 0.2872 & 0.1681 \\
    \rowcolor{rowgraygreen}LwF \cite{li2017learning} & TPAMI'17 & None & 0.3054 & 0.4234 & 0.1805 & 0.0896 \\
    \rowcolor{rowgrayyellow}MER \cite{riemer2019learning} & ICLR'19 & Raw Data & -0.0270 & 0.3956 & 0.1313 & 0.2276  \\
    \rowcolor{rowgrayyellow}DER++ \cite{buzzega2020dark} & NeurIPS'20 & Raw Data & 0.1192 & 0.3282 & 0.0985 & 0.1388 \\
    \rowcolor{rowgrayyellow}TOPIC \cite{tao2020few} & CVPR'20 & Raw Data & 0.3865 & \bf 0.0739 & 0.2298 & 0.0898 \\
    \rowcolor{rowgrayyellow}GEM \cite{kukleva2021generalized} & ICCV'21 & Raw Data & 0.2194 & 0.4185 & 0.1805 & 0.1484  \\
    \rowcolor{rowgrayyellow}STAR \cite{eskandar2025star} & ICLR'25 & Raw Data &  0.3620 & 0.2413 & -0.0821 & 0.0897 \\
    \rowcolor{rowgraypink}Feature MER & -- & Feature & 0.3447 & 0.1307 & 0.1067  & 0.1135   \\
    \rowcolor{rowgraypink}SLCA \cite{zhang2023slca} & ICCV'23 & Feature & 0.3316 & 0.2496 & 0.1231 & 0.1630 \\
    \rowcolor{rowgraypink}NC-FSCIL \cite{yang2023neural} & ICLR'23 & Feature & 0.2462 & 0.5285 & \bf 0.5416 & 0.1075 \\
    \rowcolor{rowgraypink}FS-Aug \cite{li2024continual} & TCSVT'24 & Feature & 0.0906 & 0.2400 & -0.1067  & 0.2355 \\
    \rowcolor{rowgraypink}MAGR \cite{zhou2024magr} & ECCV'24 & Feature & 0.3445 & 0.1313 & 0.4103  & 0.1630 \\
    \rowcolor{rowgraypink}ProNC \cite{wang2026rethinking} & ICLR'26 & Feature & 0.3593 & 0.4559 & 0.3282 & 0.0877 \\
    \rowcolor{rowgraypink}MAGR++ (Ours) & -- & Feature & \bf 0.4693 & 0.3774 & 0.4103 & \bf 0.0737 \\
    \bottomrule
    \end{tabular}
    }
\end{minipage}
\end{table}  
\begin{figure}
    \centering
    \includegraphics[width=\linewidth,clip,trim=50 278 50 286]{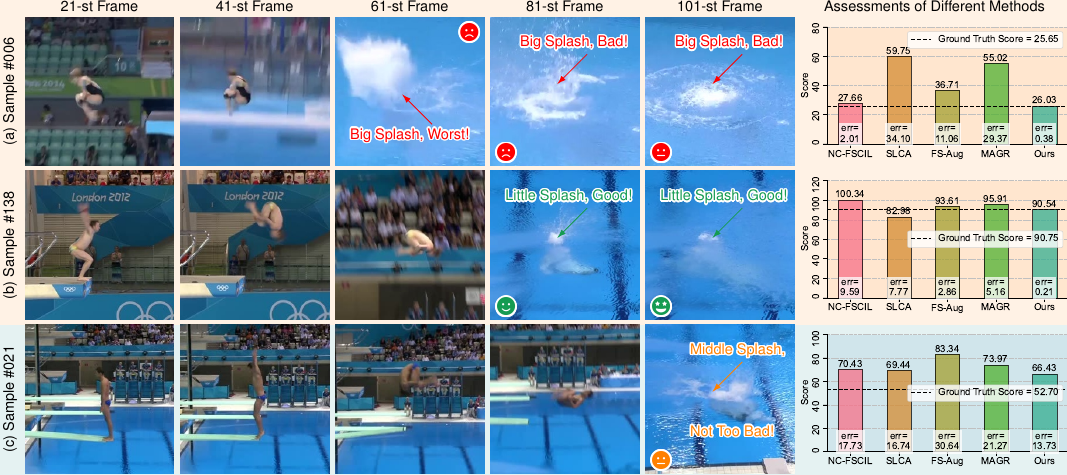}
    \caption{
    Representative samples from MTL-AQA covering high-, mid-, and low-score cases. 
    The first five columns show sampled frames, and the last column reports assessment results with errors. 
    \subref{fig:case_study-a} and \subref{fig:case_study-b} show successful cases, 
    while \subref{fig:case_study-c} depicts a failure case.
    }
    \label{fig:case_study}
    \phantomsubcaption\label{fig:case_study-a}
    \phantomsubcaption\label{fig:case_study-b}
    \phantomsubcaption\label{fig:case_study-c}
\end{figure}

\section{Additional Experiments} \label{sec:add_exp}
\subsection{Datasets} \label{sec:add_setting}

\myPara{UNLV-Vault.}
To further verify the generalization of our method beyond diving, we include the UNLV-Vault dataset in our evaluations. 
UNLV-Vault contains 176 gymnastics vault videos, which is also included as the ``vault'' class in the AQA-7 benchmark~\cite{yu2021group}. 
Each video sequence is sampled to 103 frames, covering the complete vault motion from run-up to landing.
Expert judges annotate each sample under the standard vault scoring system. 
In our experiments, we follow the same split as in prior works (120 for training and 56 for testing). 
Since vault actions differ substantially from diving, with shorter durations, more abrupt motions, and distinct visual cues, this dataset serves as a complementary testbed to assess whether our method can maintain performance under domain shift. 
The results show that MAGR++ retains strong performance on UNLV-Vault, supporting its generalization across different action domains.

\myPara{JIGSAWS.}
To evaluate cross-domain generalization beyond sports, we include the JIGSAWS surgical skill assessment dataset \cite{zhou2023hierarchical}.
JIGSAWS contains 103 annotated video clips covering three surgical tasks: Suturing (39), Needle Passing (28), and Knot Tying (36).
Each trial is evaluated by multiple human annotators using several sub-scores (e.g., motion flow and final outcome quality), which are aggregated into an overall skill score.
Following the provided annotations, we treat each trial as a regression sample and average the multiple human ratings to obtain the final target score.
The resulting scores lie roughly within a 0–5 range depending on annotators.
For continual evaluation, we follow the same class-incremental protocol as in our previous settings, where the three surgical tasks are treated as three sequential sessions.
This setup enables a controlled cross-domain CAQA benchmark that differs significantly from sports-based datasets in both motion characteristics and scoring criteria.

\begin{table}[!h]
\centering
\setlength{\tabcolsep}{3pt}
\caption{Additional ablation results on MTL-AQA. Reported percentages denote relative changes compared to ID~1.}
\label{tab:ablation_extra}
\begin{tabular}{p{0.2cm} p{3.8cm} p{1.5cm} p{1.6cm} p{1.6cm}}
\toprule
\textbf{ID} & \textbf{Setting} & $\rho_{\text{avg}}$ ($\uparrow$) & $\rho_{\text{aft}}$ ($\downarrow$) & $\rho_{\text{fwt}}$ ($\uparrow$) \\
\midrule
\rowcolor{rowgraypink} 1 & MAGR++ (Ours) & 0.9205 & 0.0103 & 0.1274 \\
\rowcolor{rowgraygreen} 2 & ~~MP w/o Residual Link & 0.8933$^{-3\%}$ & 0.0389$^{+278\%}$ & 0.0693$^{-46\%}$  \\
\rowcolor{rowgrayyellow} 3 & ~~\cref{eq:iij} w/ KL Loss & 0.9173$^{-0.3\%}$ & 0.0155$^{+51\%}$ & 0.1029$^{-19\%}$ \\
\bottomrule
\end{tabular}
\end{table}

\subsection{Baseline Implementation Details} \label{sec:base_details}

For a fair comparison, we distinguish benchmark-level settings from method-specific hyper-parameters. Benchmark-level settings, including the backbone, input features, train/test splits, session order, memory budget, replay mini-batch size, batch size, and online/offline evaluation protocol, are fixed before method evaluation and kept identical for all methods. These settings are determined by the CAQA benchmark protocol rather than by the performance of MAGR++.
For method-specific hyper-parameters, we follow the original papers or official implementations whenever available, and adapt them to the CAQA setting within comparable tuning ranges. For replay-based methods, identical memory capacities and replay constraints are used in both offline and online settings. For online CL, all methods follow the same streaming protocol, where each sample is observed only once, i.e., one epoch, under the same update schedule.
Detailed hyper-parameter configurations and implementation settings are summarized in \cref{tab:baseline_impl} and further provided in the supplementary material. TOPIC is re-implemented following the original paper with $\lambda_1=0.5$ and $\lambda_2=0.55$, while Feature MER is adapted from the original MER implementation to support feature-level replay in CAQA.

\begin{table}[t] 
\centering 
\MStabcaption{tab:baseline_impl}{Implementation details of compared methods. Benchmark-level settings, such as memory size, replay mini-batch, batch size, and online/offline training epochs, are fixed by the CAQA protocol and shared across methods. Method-specific hyper-parameters, such as regularization weights and learning rates when applicable, follow the original papers or official implementations.} 
\resizebox{0.9\linewidth}{!}{ \begin{tabular}{lcccccccc} \toprule Method & Optimizer & LR & Epochs & Memory & Replay Ratio & Replay Mini-Batch & Batch Size & Source \\ \midrule EWC & Adam & $1\mathrm{e}{-4}$ & 1/50 & -- & -- & -- & 5 & \href{https://github.com/aimagelab/mammoth/blob/master/models/ewc.py}{Repo} \\ SI & Adam & $1\mathrm{e}{-4}$ & 1/50 & -- & -- & -- & 5 & \href{https://github.com/aimagelab/mammoth/blob/master/models/si.py}{Repo} \\ LwF & Adam & $1\mathrm{e}{-4}$ & 1/50 & -- & -- & -- & 5 & \href{https://github.com/aimagelab/mammoth/blob/master/models/lwf.py}{Repo} \\ MER & Adam & $1\mathrm{e}{-4}$ & 1/50 & 200 & 1:1 & 3 & 5 & \href{https://github.com/aimagelab/mammoth/blob/master/models/mer.py}{Repo} \\ DER++ & Adam & $1\mathrm{e}{-4}$ & 1/50 & 200 & 1:1 & 3 & 5 & \href{https://github.com/aimagelab/mammoth/blob/master/models/derpp.py}{Repo} \\ TOPIC & Adam & $1\mathrm{e}{-4}$ & 1/50 & 200 & 1:1 & 3 & 5 & Re-implemented from the paper ($\lambda_1=0.5$,$\lambda_2=0.55$) \\ GEM & Adam & $1\mathrm{e}{-4}$ & 1/50 & 200 & 1:1 & 3 & 5 & \href{https://github.com/aimagelab/mammoth/blob/master/models/gem.py}{Repo} \\ STAR & Adam & $1\mathrm{e}{-4}$ & 1/50 & 200 & 1:1 & 3 & 5 & \href{https://github.com/Gnomy17/STAR_CL}{Repo} \\ Feature MER & Adam & $1\mathrm{e}{-4}$ & 1/50 & 200 & 1:1 & 3 & 5 & Adapted from MER \\ SLCA & Adam & $1\mathrm{e}{-4}$ & 1/50 & 200 & 1:1 & 3 & 5 & \href{https://github.com/aimagelab/mammoth/blob/master/models/slca.py}{Repo} \\ NC-FSCIL & Adam & $1\mathrm{e}{-4}$ & 1/50 & 200 & 1:1 & 3 & 5 & \href{https://github.com/NeuralCollapseApplications/FSCIL}{Repo} \\ FS-Aug & Adam & $1\mathrm{e}{-4}$ & 1/50 & 200 & 1:1 & 3 & 5 & \href{https://github.com/iSEE-Laboratory/Continual-AQA}{Repo} \\ MAGR & Adam & $1\mathrm{e}{-4}$ & 1/50 & 200 & 1:1 & 3 & 5 & \href{https://github.com/ZhouKanglei/MAGR_CAQA}{Repo} \\ ProNC & Adam & $1\mathrm{e}{-4}$ & 1/50 & 200 & 1:1 & 3 & 5 & \href{https://github.com/Continue-Edge-AI-Lab/ProNC}{Repo} \\ MAGR++ & Adam & $1\mathrm{e}{-4}$ & 1/50 & 200 & 1:1 & 3 & 5 & Ours \\ 
\bottomrule 
\end{tabular}} 
\end{table} 
\subsection{Results and Analysis}

\begin{figure}
    \centering
    \sf
    \begin{minipage}{0.9\linewidth}
        \begin{overpic}[width=0.55\linewidth,height=0.18\linewidth,clip,trim=190 190 200 172]{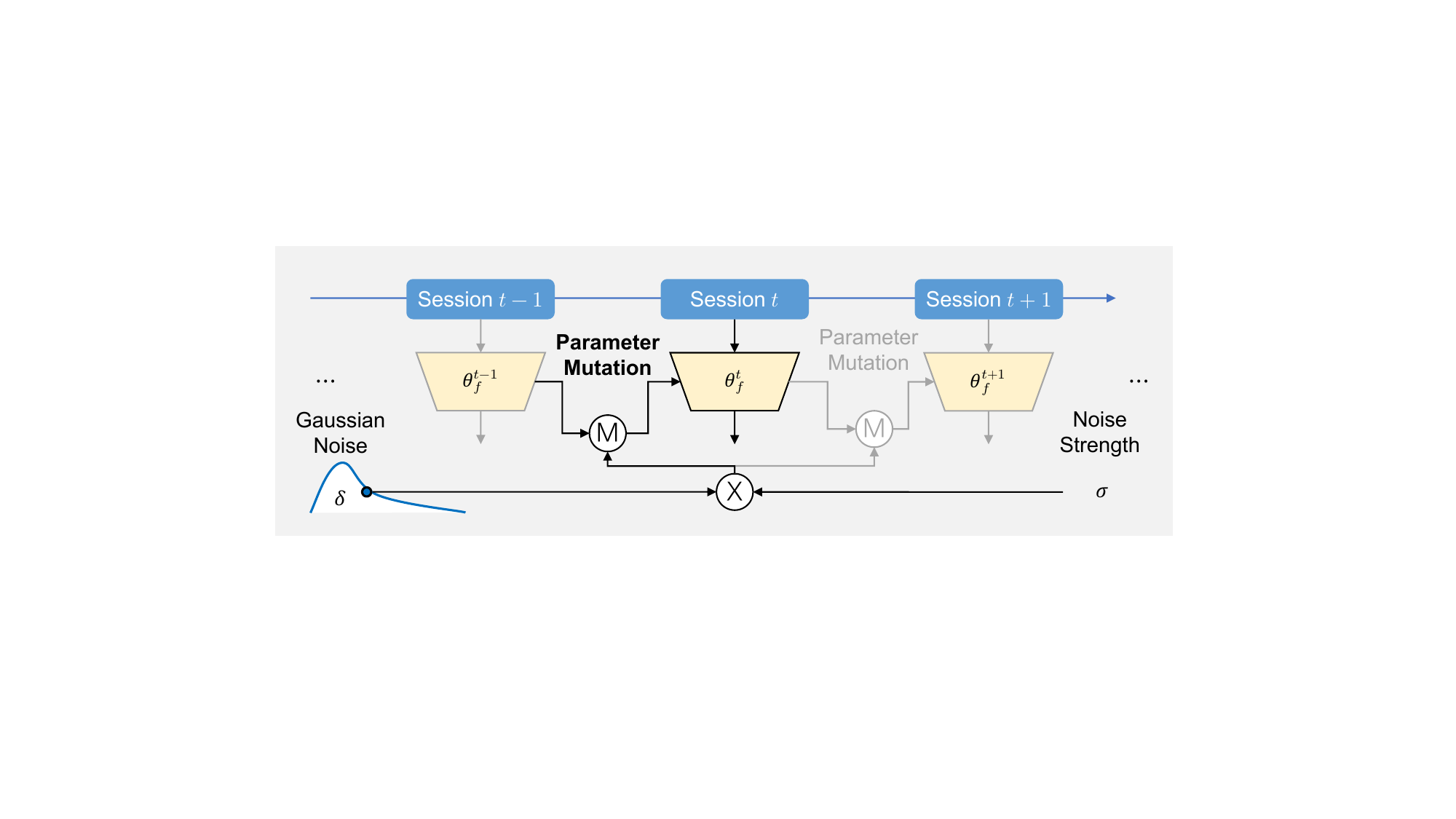}\put(102,1){\scriptsize\textbf{(b)}}
            \put(96,1){\scriptsize\textbf{(a)}}
        \end{overpic}\begin{overpic}[width=0.36\linewidth,height=0.18\linewidth]{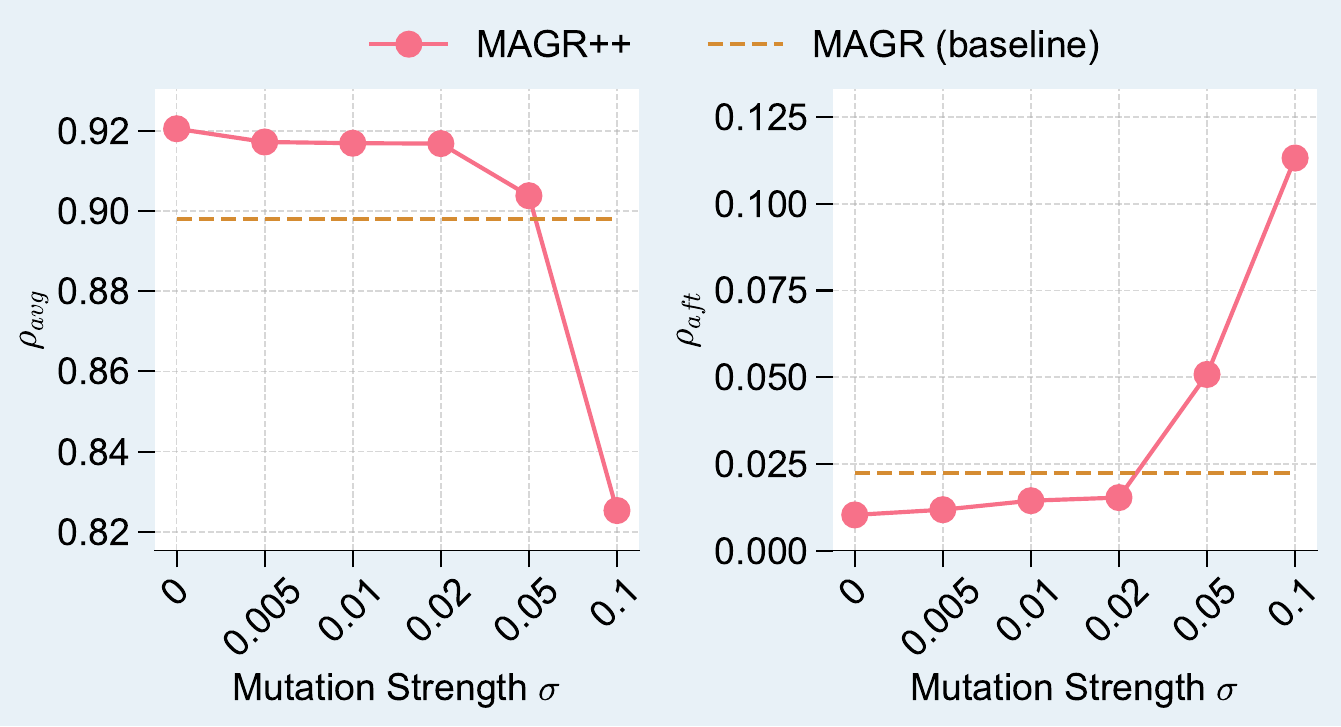}
        \end{overpic}
    \end{minipage}
    \MSfigcaption{fig:mutation}{
    Robustness under parameter mutation.
    (a) Illustration of the mutation process, where Gaussian noise is injected into the backbone parameters at the beginning of each session to simulate violations of the model continuity assumption.
    (b) Performance under increasing mutation strength $\sigma$.
    MAGR++ maintains a stable average correlation under mild and moderate mutations, and degrades more gracefully than prior methods, demonstrating strong robustness beyond the theoretical assumptions.
    }{
        \phantomsubcaption\label{fig:mutation-a}
        \phantomsubcaption\label{fig:mutation-b}
    }
    \vspace{-0.2cm}
\end{figure}

\myPara{Robustness under Parameter Mutation.}
To examine robustness when the continuity assumption is violated, we simulate parameter mutation by injecting Gaussian perturbations into the backbone parameters at the beginning of each session. As shown in \cref{fig:mutation-a}, the parameters are updated as
\begin{equation}
\bm{\theta}_f^{t} \leftarrow 
\bm{\theta}_f^{t-1} + 
\sigma \cdot \frac{\|\bm{\theta}_f^{t-1}\|_2}{\sqrt{d}}\,\bm{\delta},
\quad
\bm{\delta} \sim \mathcal{N}(\mathbf{0}, \mathbf{I}),
\label{eq:mutation}
\end{equation}
where $d$ is the number of perturbed parameters and $\sigma$ controls the mutation strength. This normalized formulation ensures that the perturbation magnitude is proportional to the parameter norm, providing a scale-invariant and controlled simulation of abrupt parameter changes.
This setup directly relaxes the model continuity assumption (B4) in \cref{thm:fpft-replay}. 
While \cref{eq:forgetting-bound} assumes bounded prediction drift under smooth parameter updates, the perturbation in \cref{eq:mutation} explicitly introduces discontinuities into the parameter trajectory. 
Therefore, this experiment provides a practical stress test of the forgetting-control mechanism of MAGR++ under conditions that deviate from the theoretical assumptions.
As shown in \cref{fig:mutation-b}, MAGR++ maintains stable performance under mild to moderate mutations. When $\sigma$ increases from 0 to 0.02, the average correlation only slightly decreases from 0.9205 to 0.9168, indicating strong tolerance to parameter perturbations. Even under stronger mutation ($\sigma=0.05$), MAGR++ still achieves a competitive $\rho_{\text{avg}}$ of 0.9038, outperforming both FS-Aug (0.8060) and the original MAGR (0.8979). Only under extreme perturbation ($\sigma=0.10$) does performance degrade significantly, which is expected when the continuity assumption is heavily violated.
These results demonstrate that MAGR++ effectively controls forgetting even when the theoretical continuity condition is relaxed, confirming the practical robustness of the proposed design.

\myPara{Generalization Beyond Diving.}  
To further verify the cross-domain robustness of our approach, we evaluate MAGR++ on the UNLV-Vault dataset, which differs significantly from diving in terms of motion dynamics and temporal structure. 
As summarized in \cref{tab:vault}, MAGR++ achieves the best overall performance in both offline and online settings. 
In the offline case, it attains $\rho_{\text{avg}}=0.7012$, outperforming the strongest baseline MAGR~\cite{zhou2024magr} by $+0.0486$, NC-FSCIL~\cite{yang2023neural} by $+0.1265$, and SLCA~\cite{zhang2023slca} by $+0.2093$. 
In the online setting, MAGR++ reaches $\rho_{\text{avg}}=0.5806$, yielding a substantial improvement of $+0.0835$ over NC-FSCIL (0.4971) and $+0.3810$ over FS-Aug (0.1998). 
It also maintains zero forgetting ($\rho_{\text{aft}}=0$) and the highest forward transfer ($\rho_{\text{fwt}}=0.7057$), indicating strong adaptability without sacrificing stability.  
These results demonstrate that the proposed layer-adaptive fine-tuning and two-step rectification effectively preserve feature–score alignment even when transferred to unseen domains with distinct motion and visual characteristics.

\myPara{Generalization to Surgical Skill Assessment.}  
We further evaluate MAGR++ on the JIGSAWS dataset, which represents a substantially different domain from sports AQA, featuring fine-grained hand–tool interactions and highly constrained surgical motions. 
As shown in \cref{tab:vault}, MAGR++ achieves the best performance in both offline and online settings. 
In the offline case, it reaches $\rho_{\text{avg}}=0.8238$, outperforming the strongest baseline ProNC~\cite{wang2026rethinking} by $+0.0226$, MAGR~\cite{zhou2024magr} by $+0.0239$, and SLCA~\cite{zhang2023slca} by $+0.0244$, while also achieving the lowest forgetting ($\rho_{\text{aft}}=0.0192$) and the highest forward transfer ($\rho_{\text{fwt}}=0.7549$). 
In the online setting, MAGR++ attains $\rho_{\text{avg}}=0.4693$, exceeding the strongest baseline STAR~\cite{eskandar2025star} by $+0.1073$ and ProNC by $+0.1100$. 
These results indicate that MAGR++ maintains stable performance and effective knowledge transfer even under fine-grained surgical motion distributions, demonstrating strong cross-domain generalization beyond sports scenarios.
 
\myPara{Ablation Study.}
As shown in \cref{tab:ablation_extra}, removing the residual link in MP leads to a notable decline in performance, 
with the average correlation dropping by about 3\%, the average forgetting increasing nearly threefold, 
and the forward transfer reduced by almost half. 
This observation highlights that the residual connection is essential for capturing substantial feature variations and maintaining stability across continual updates. 
In contrast, substituting the MSE term in \cref{eq:iij} with a KL divergence yields only marginal changes, 
demonstrating that our method remains robust regardless of the specific feature-matching loss. 
Overall, these results confirm the stability and robustness of MAGR++, showing that its performance is largely insensitive to minor design variations, yet heavily reliant on the residual link for effective adaptation.

\myPara{Hyper-Parameter Sensitivity Analysis.}
We analyze the sensitivity of MAGR++ to the loss weights 
$\lambda_{\text{tune}}$, $\lambda_{\text{proj}}$, and $\lambda_{\text{reg}}$
by varying one parameter at a time while fixing the others to 1.
As shown in \cref{fig:hyperparam}, performance varies smoothly over a wide range of values
(0.1, 0.5, 1.0, 2.0, 5.0) for all three parameters, with no abrupt degradation observed.
Overall, the best performance is typically achieved when the weights are close to 1.0,
while using non-unit values leads to comparable performance trends.
This behavior is expected because all loss terms are explicitly normalized to comparable scales.
Setting the weights around 1 therefore maintains a balanced contribution from backbone regularization
($\mathcal{L}_{\text{tune}}$), manifold projection ($\mathcal{L}_{\text{proj}}$), and graph regularization
($\mathcal{L}_{\text{reg}}$).
When a weight deviates significantly from this range, optimization may overemphasize a single component,
slightly disturbing the stability–plasticity balance.
Nevertheless, the smooth performance variation around this region indicates that MAGR++ does not rely on
precise hyperparameter tuning, and the default choice of unit weights provides a robust and effective operating point.
We therefore set all loss weights to 1 in the main experiments,
both for robustness and to ensure fair and consistent comparison with prior methods
and our conference version~\cite{zhou2024magr}.

\begin{figure}
    \centering
    \includegraphics[width=0.75\linewidth]{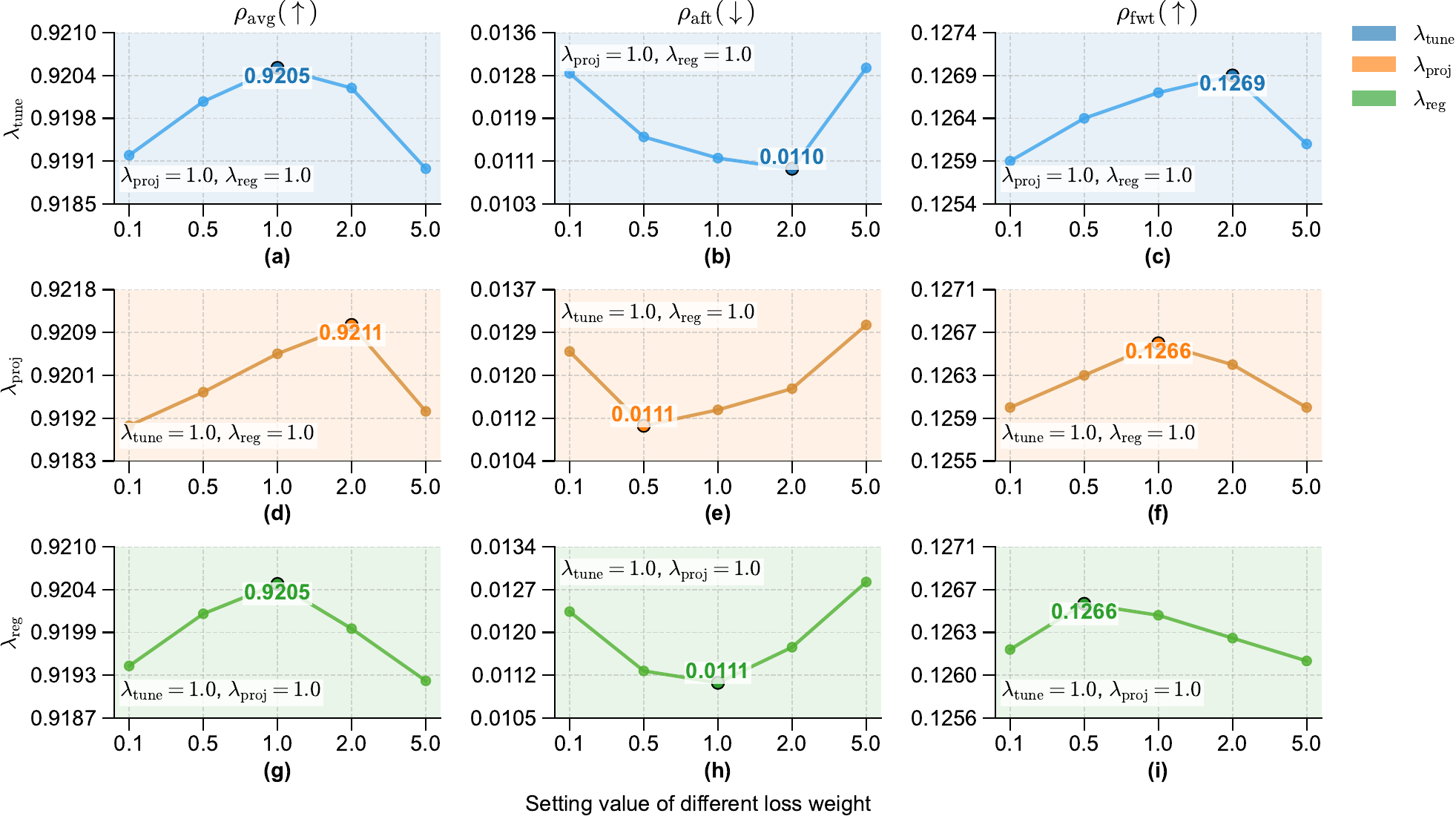}
    \MSfigcaption{fig:hyperparam}{Sensitivity analysis of MAGR++ with respect to the loss weights 
$\lambda_{\text{tune}}$, $\lambda_{\text{proj}}$, and $\lambda_{\text{reg}}$.
Each parameter is varied independently while fixing the others to 1.}{}

\end{figure} 
\myPara{Case Study.}  
\cref{fig:case_study} illustrates representative samples from the MTL-AQA dataset, covering low-, high-, and mid-score diving scenarios, with predictions compared across SLCA~\cite{zhang2023slca}, NC-FSCIL~\cite{yang2023neural}, FS-Aug~\cite{li2024continual}, MAGR~\cite{zhou2024magr}, and our approach. In the low-score case (Sample~\#006), the dive produces a large splash indicating poor execution, where the ground-truth score is 25.65, and our method predicts 26.03 with only 0.38 error, while SLCA (59.75, error 34.10) and FS-Aug (36.71, error 11.06) perform much worse. In the high-score case (Sample~\#138), the nearly splash-free entry yields a ground truth of 90.75, and our method achieves 90.54 with 0.21 error, whereas NC-FSCIL (100.34, error 9.59) and SLCA (82.58, error 7.77) deviate substantially. Even in the more challenging mid-score case (Sample~\#021), where the ground truth is 52.70, our method outputs 66.43, closer to the target than NC-FSCIL (70.43) and SLCA (69.44). These examples highlight the superior reliability of our method across both extreme and intermediate performance levels. At the same time, they reveal common error modes: occlusions (e.g., body–water overlap) may lead to misaligned features, while abrupt motion changes can induce projection failures, which explains the remaining discrepancies. Analyzing such cases not only clarifies why errors occur but also highlights promising directions for future research, such as developing representations that are intrinsically robust to occlusion and designing projection mechanisms that explicitly account for dynamic motion patterns.

\section{Additional Discussion and Future Work}
While MAGR++ demonstrates strong performance and generalization across diverse CAQA benchmarks, several open challenges remain. 
First, although the proposed layer-adaptive fine-tuning effectively balances stability and adaptability, it relies on clustering-based abstraction estimation. Future work could explore more efficient or theoretically grounded criteria, such as information-theoretic or gradient-based layer importance measures. 
Second, MP is currently implemented as a simple MLP, which assumes local smoothness in feature transitions. Incorporating spatiotemporal attention or motion-conditioned projection could better handle complex distribution shifts caused by abrupt dynamics or occlusions.  
Third, while our two-step rectification preserves feature–score alignment, it primarily focuses on visual modality. Extending this framework to multi-modal settings (e.g., integrating pose or textual feedback) would further enhance interpretability and robustness.  
Finally, although our evaluations cover multiple datasets and both offline and online CAQA settings, future studies could examine real-time deployment and memory-limited environments to further assess scalability and practicality.  
Overall, we envision MAGR++ as a foundation for building trustworthy, adaptive AQA systems capable of CL under realistic scenarios.

\end{document}